\documentclass[pdflatex,sn-nature]{sn-jnl}

\usepackage{graphicx}%
\usepackage{multirow}%
\usepackage{amsmath,amssymb,amsfonts}%
\usepackage{amsthm}%
\usepackage{mathrsfs}%
\usepackage[title]{appendix}%
\usepackage{xcolor}%
\usepackage{textcomp}%
\usepackage{manyfoot}%
\usepackage{booktabs}%
\usepackage{algorithm}%
\usepackage{algorithmicx}%
\usepackage{algpseudocode}%
\usepackage{listings}%
\usepackage{lineno}

\renewcommand{\appendixname}{SI}
\newcommand{\appendixsectionformat}{\renewcommand{\thesection}{\arabic{section}}}
\newcommand{\appendixsubsectionformat}{\renewcommand{\thesubsection}{SI \arabic{section}.\Alph{subsection}}}
\newcommand{\appendixfigureformat}{\renewcommand{\thefigure}{S\arabic{figure}}\renewcommand{\theHfigure}{S\arabic{figure}}}
\newcommand{\appendixtableformat}{\renewcommand{\thetable}{S\arabic{table}}\renewcommand{\theHtable}{S\arabic{table}}}
\newcommand{\appendixequationformat}{\renewcommand{\theequation}{S\arabic{equation}}\renewcommand{\theHequation}{S\arabic{equation}}}

\usepackage{hyperref}
\usepackage{bookmark}
\usepackage{url}

\usepackage{nicefrac} 
\usepackage{adjustbox}
\usepackage{dsfont}
\usepackage{makecell}
\usepackage{microtype}
\usepackage{tabularx}
\usepackage{xspace}
\usepackage{lmodern}
\usepackage{anyfontsize}
\usepackage[T1]{fontenc}
\usepackage{multicol}

\begin{document}

\title[Article Title]{Revealing emergent human-like conceptual representations from language prediction\footnote{Accepted manuscript. Final version published in \textit{Proceedings of the National Academy of Sciences (PNAS)}: \url{https://www.pnas.org/doi/10.1073/pnas.2512514122}.}}

\author[1,2]{\fnm{Ningyu} \sur{Xu}} 

\author[1,3,5]{\fnm{Qi} \sur{Zhang}} 

\author[4]{\fnm{Chao} \sur{Du}} 

\author[4]{\fnm{Qiang} \sur{Luo}} 

\author*[1,5]{\fnm{Xipeng} \sur{Qiu}}\email{xpqiu@fudan.edu.cn}

\author*[1,5]{\fnm{Xuanjing} \sur{Huang}}\email{xjhuang@fudan.edu.cn}

\author*[2,4,6]{\fnm{Menghan} \sur{Zhang}}\email{mhzhang@fudan.edu.cn}

\affil[1]{\orgdiv{College of Computer Science and Artificial Intelligence}, \orgname{Fudan University}, \orgaddress{\state{Shanghai}, \country{China}}}

\affil[2]{\orgdiv{Institute of Modern Languages and Linguistics}, \orgname{Fudan University},  \orgaddress{\state{Shanghai}, \country{China}}}

\affil[3]{\orgdiv{Shanghai Key Laboratory of Intelligent Information Processing}, \orgaddress{\state{Shanghai}, \country{China}}}

\affil[4]{\orgdiv{Research Institute of Intelligent Complex Systems}, \orgname{Fudan University}, \orgaddress{\state{Shanghai}, \country{China}}}

\affil[5]{\orgdiv{Institute of Trustworthy Embodied Artificial Intelligence, Fudan University}, \orgaddress{\state{Shanghai}, \country{China}}}

\affil[6]{\orgdiv{State Key Laboratory of Genetics and Development of Complex Phenotypes}, \orgname{Fudan University}, \orgaddress{\state{Shanghai}, \country{China}}}

\abstract{
People acquire concepts through rich physical and social experiences and use them to understand and navigate the world. In contrast, large language models (LLMs), trained solely through next-token prediction on text, exhibit strikingly human-like behaviors. Are these models developing concepts akin to those of humans? If so, how are such concepts represented, organized, and related to behavior? Here, we address these questions by investigating the representations formed by LLMs during an in-context concept inference task. We found that LLMs can flexibly derive concepts from linguistic descriptions in relation to contextual cues about other concepts. The derived representations converge toward a shared, context-independent structure, and alignment with this structure reliably predicts model performance across various understanding and reasoning tasks. Moreover, the convergent representations effectively capture human behavioral judgments and closely align with neural activity patterns in the human brain, providing evidence for biological plausibility. Together, these findings establish that structured, human-like conceptual representations can emerge purely from language prediction without real-world grounding, highlighting the role of conceptual structure in understanding intelligent behavior. More broadly, our work suggests that LLMs offer a tangible window into the nature of human concepts and lays the groundwork for advancing alignment between artificial and human intelligence.
}

\maketitle

\bmhead{Significance Statement}

Large language models (LLMs) show intriguing human-like behaviors despite being trained solely via language prediction. Are these models developing human-like concepts central to human understanding? Here, we demonstrate that LLMs can flexibly derive concepts from linguistic descriptions across varying contexts. The derived representations converge toward a shared, context-independent structure, which effectively predicts human behavioral judgments and aligns closely with neural activity patterns in the human brain. Our findings reveal that critical aspects of human concepts are learnable purely from language prediction. Rather than relying on real-world grounding, LLMs organize concepts through meaningful interrelationships preserved across contexts, providing a tractable window into human conceptual representation and organization.

\clearpage

\section{Introduction}
\label{sec1}

Humans are able to construct mental models of the world and use them to understand and navigate their environment~\citep{johnson-laird_mental_1986, lake_building_2017}. Central to this ability is the capacity to form broad concepts that constitute the building blocks of these models~\citep{murphy_big_2004}. These concepts, often regarded as mental representations, capture regularities in the world while abstracting away extraneous details, enabling flexible generalization to novel situations~\citep{mitchell_abstraction_2021}. For example, the concept of \textsc{sun} can be instantly formed and deployed across diverse contexts: observing it rise or set in the sky, yearning for its warmth on a chilly winter day, or encountering someone who exudes positivity. Its role in the solar system can be analogized to the nucleus in an atom, enriching learning and understanding. The nature of concepts has long been a focus of inquiry across philosophy, cognitive science, neuroscience, and linguistics~\citep{aristotle_1980, ross_plato_1966, shepard_toward_1987, mcclelland_letting_2010, tenenbaum_how_2011, mitchell_predicting_2008, jackendoff_foundations_2002}. These investigations have uncovered diverse properties that concepts need to satisfy, often framed by the long-standing divide between symbolism and connectionism. Symbolism emphasizes discrete, explicit symbols with structured and compositional properties, enabling abstract reasoning and recombination of ideas~\citep{fodor_language_1975, fodor_connectionism_1988}. In contrast, connectionism conceptualizes concepts as distributed, emergent patterns across networks, prioritizing continuity and gradedness, which excel in handling noisy inputs and learning from experience~\citep{rumelhart_parallel_1986, mcclelland_parallel_2003}. Although there is growing consensus on the need to integrate the strengths of both paradigms to account for the complexity and richness of human concepts~\citep{margolis_conceptual_2015, smolensky_neurocompositional_2022}, reconciling the competing demands remains a significant challenge.

Recent advances in large language models (LLMs) within artificial intelligence (AI) have exhibited human-like behavior across various cognitive and linguistic tasks, from language generation~\citep{brown_language_2020, jones_does_2024} to decision-making~\citep{binz_using_2023} and reasoning~\citep{webb_emergent_2023, hagendorff_human-like_2023, lampinen_language_2024}. It has sparked intense interest and debate about whether these models are approaching human-like cognitive capacities based on concepts~\citep{frank_openly_2023, mitchell_debate_2023, lake_word_2023, shiffrin_probing_2023, mahowald_dissociating_2024, ananthaswamy_how_2024}. Some argue that LLMs, trained solely on text data for next-token prediction, operate only on statistical associations between linguistic forms and lack concept-based understanding grounded in physical and social situations~\citep{bender_climbing_2020, bender_danger_2021, mahowald_dissociating_2024}. This argument is evidenced by their inconsistent performance, which often reveals vulnerabilities such as non-human-like errors and over-sensitivity to minor contextual variations~\citep{binz_using_2023, lewis_using_2024, dentella_testing_2024}. Conversely, others contend that the performance of a system alone is insufficient to characterize its underlying competence~\citep{firestone_performance_2020}, and the extent to which concepts should be grounded remains open~\citep{pavlick_symbols_2023, kim_shared_2021}. Instead, language may provide essential cues for people's use of concepts and enable LLMs to approximate fundamental aspects of human cognition, where meaning arises from the relationships among concepts~\citep{piantadosi_meaning_2022, piantadosi_why_2024}. Despite the conflicting views, there is broad consensus on the central role of concepts in understanding both human and machine cognition. The core questions driving the debate are whether LLMs possess concepts that embody the key properties of human concepts, and how such concepts are represented, organized, and linked to behavior. 

\begin{figure}[t!]
\centering
\includegraphics[width=\textwidth]{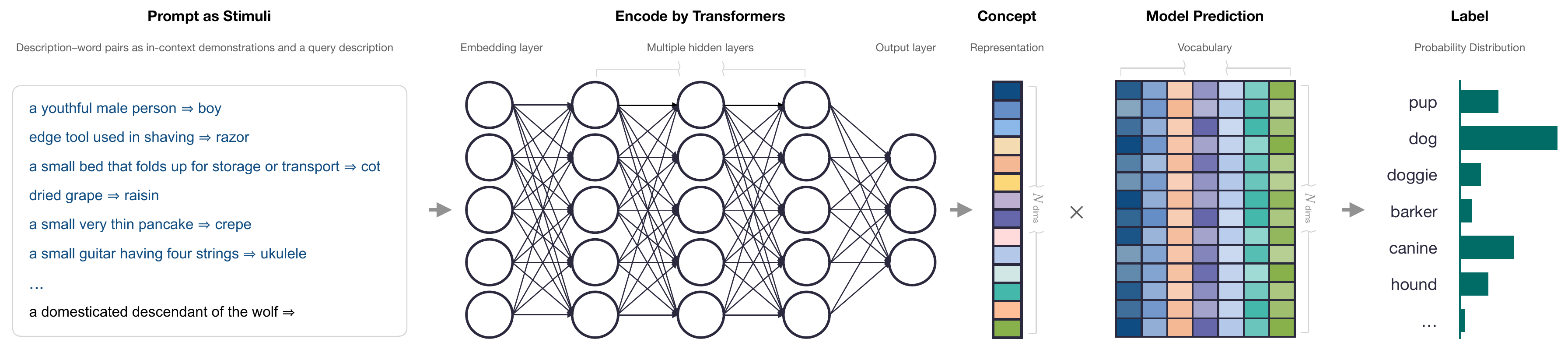}
\caption{Illustration of the reverse dictionary task as a conceptual probe. A Transformer-based LLM is presented with $N$ description--word pairs as demonstrations in context, followed by a query description. The model is then prompted to encode the query description into a conceptual representation and predict the term that best matches the described concept.}\label{fig:concept}
\end{figure}

To address these questions, we investigated conceptual representations emerging from language prediction in LLMs. Our approach unfolded in three stages. First, we examined LLMs' capacities to form and organize conceptual representations, focusing on the definitional and structured properties of human concepts. We guided LLMs to derive concepts from descriptions with a few contextual demonstrations (Fig.~\ref{fig:concept}). The models' outputs offered a behavioral lens to evaluate the derived representations. We then explored the organization of these representations by analyzing how their relational structure varied across contexts, and further examined how this structure related to model performance on a range of language understanding and reasoning tasks. Second, we assessed the extent to which the LLM-derived representational structure aligns with psychological measures and captures rich, context-dependent human knowledge. These properties were tested by using computations over the representations to predict human behavioral judgments. Finally, we examined the biological plausibility of these representations by mapping them to neural activity patterns in the human brain. Our experiments spanned thousands of naturalistic object concepts~\citep{hebart_things-data_2023} and beyond, providing a comprehensive analysis of LLMs' conceptual representations. The findings reveal that language prediction alone can give rise to human-like conceptual representations in LLMs. These representations are structured in a way that integrates critical aspects of human concepts, combining the definitional and structural focus of symbolism with the continuity and graded nature of connectionist models. Alignment with this structure reliably predicts model performance across diverse tasks. Collectively, this work underscores the profound connection between language and human conceptual structure, and provides a simple, generalizable framework for deriving conceptual representations from LLMs as a reusable resource, opening new pathways for exploring the nature of human concepts. 

\section{Results}
\label{sec:results}

\subsection{Reverse dictionary as a conceptual probe}

We reframed the reverse dictionary task as a domain-general probe of LLMs' capacity for concept inference. People can identify and think about concepts from definitions or descriptions, even when these are incomplete or imprecise~\citep{murphy_big_2004}. The reverse dictionary task provides a natural simulation of this process. Formally, given a description of a concept $c$, a system encodes it into a latent representation conditioned on context, $\mathbf{h}_{c} = \textrm{encode} \left(\textrm{description} \mid \textrm{context}\right)$, and generates a corresponding term: $\textrm{predict}\left(\mathbf{h}_{c}\right)$. Rather than relying on a single, potentially ambiguous word (e.g., ``bat''), this process requires abstraction beyond surface forms, unifying diverse linguistic expressions of the same underlying concept (e.g., ``a round, glowing object that hangs in the night sky'' and ``Earth's natural satellite'' both correspond to the concept \textsc{moon}). The generated term (e.g., ``moon'') provides a tangible lens on whether the inferred representation aligns with the intended concept. This paradigm thus offers a targeted probe into a system's capacity to contextually form consistent and shareable representations, capturing two key properties of human concepts: definitions and language generation.

To guide LLMs through the process, we took advantage of their in-context learning ability and presented them with a small number of demonstrations ($N$) in a reverse-dictionary format (``$[\textrm{Description}] \Rightarrow [\textrm{Word}]$''), followed by a query description (Fig.~\ref{fig:concept}). We then compared model-generated completion to the intended term of the query concept, thereby evaluating how well the model-derived conceptual representations aligned with human understanding. Importantly, the demonstrations provided minimal and controllable context, and the query description served as an input stimulus for concept inference. We can analyze how models construct conceptual representations in relation to contextual cues from other concepts by varying the number of demonstrations and the description--word pairings within them.

\subsection{Deriving concepts from definitional descriptions}
\label{sec:results-concepts}

\begin{figure}[t!]
\centering
\includegraphics[width=\textwidth]{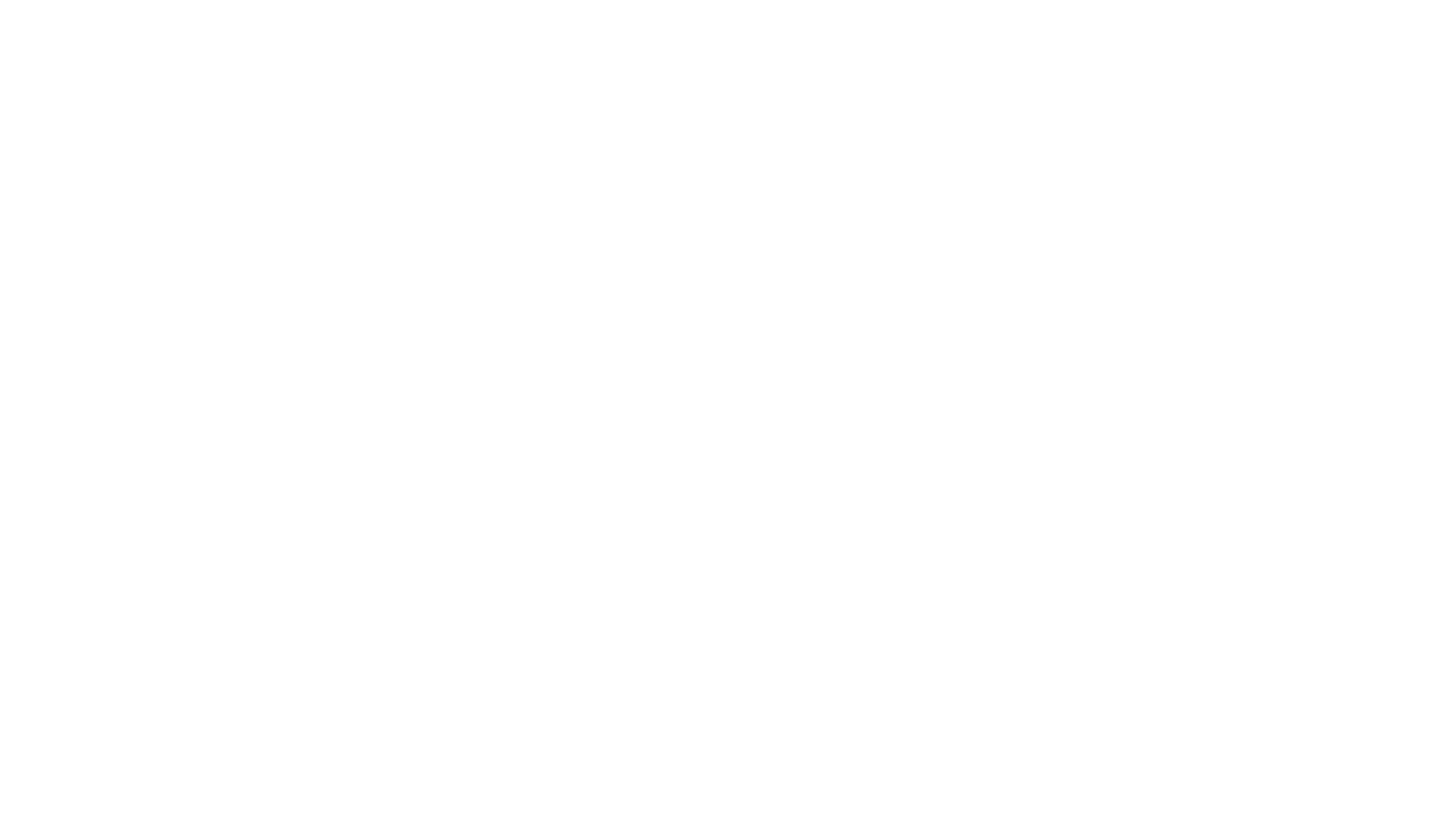}
\caption{Performance of LLaMA3-70B on the reverse dictionary task measured through exact match accuracy. Black lines show performance when the model is given $N$ correct demonstrations sampled from the training set and evaluated on an independent test set. Blue and red lines show performance when one misleading demonstration---a description paired with a proxy label (capital letter, random string, or random word, as specified in the figure titles)---is added to $N - 1$ correct demonstrations of other concepts. The model is queried with the description identical to the misleading one. The blue line shows the frequency with which the model reproduces the proxy label, while the red line indicates how often it generates the correct word for the query concept given contextual information from other concepts. Shaded areas denote $95\%$ bootstrapped confidence intervals, calculated from 10,000 resamples over five independent runs.}\label{fig:relation}
\end{figure}

We investigated whether LLMs can construct concepts from definitional descriptions through the reverse dictionary task. Leveraging data from the THINGS database~\citep{hebart_things_2019}, we prompted LLMs with randomly selected description--word pair demonstrations and examined their capacity to predict appropriate terms for the query descriptions (\textit{Materials and Methods}). Fig.~\ref{fig:relation} shows the performance of LLaMA3-70B, a state-of-the-art open-source LLM, averaged across five independent runs at varying numbers of demonstrations. The exact match accuracy improved progressively as the number of demonstrations increased from 1 to 24, rising from $79.51\%$ ($\pm 0.30\%$) to $89.45\%$ ($\pm 0.30\%$). Gains were marginal beyond this threshold. These results indicate that, with a few demonstrations as context, LLMs can effectively combine words into coherent representations and reliably infer the corresponding concepts, despite varying contextual cues.

As a follow-up, we conducted a counterfactual analysis to investigate how LLMs accomplish this: Do they merely function as lookup tables, rigidly mapping descriptions to words, or do they contextually infer concepts based on their interrelationships? As shown in Fig.~\ref{fig:relation}, when presented with only a single demonstration identical to the query description, the model replicated the context, even though the description was paired with a proxy symbol instead of the correct term. As additional correct demonstrations of other concepts were introduced, the model gradually shifted from replication to predicting the proper word for the query concept. This behavior persisted across various proxy symbols, suggesting that contextual cues about other concepts successfully guided the model in disregarding misleading information. When the correct demonstration for the query concept was provided, model performance slightly declined as more demonstrations were added, eventually dropping below the average in the standard generalization setting. This decline indicated that subtle conflicts among demonstrations could arise, with the influence of other concepts overshadowing that of the identical query concept. Collectively, these findings underscore the intricate interrelationships among concepts and their pivotal role in shaping model inference.

Furthermore, we extended our analyses to a broader range of descriptions and concepts to assess the generality of our findings derived from the THINGS database. Using data from WordNet, our extended analysis demonstrated the strong adaptability of LLMs across concepts differing in concreteness and word classes (Fig.~\ref{fig:probe-si}\textit{A}). Consistent results were also observed across various descriptions of the same concept (Fig.~\ref{fig:probe-si}\textit{B}). Introducing varying degrees of word order permutations to the query descriptions revealed that LLM predictions were sensitive to linguistic structure degradation when combining words to form concepts (Fig.~\ref{fig:probe-si}\textit{C}). These findings highlight the model's effectiveness in concept inference and its ability to capture at least some of the compositional structure in natural language. Beyond LLaMA3-70B, we tested 66 additional open-source LLMs with different model architectures, scales and training data. The results revealed trends similar to those observed with LLaMA3-70B. They also indicated that larger models generally perform better and more effectively leverage contextual cues for concept inference (\textit{SI Appendix}, section~\ref{si:results-generality}, Fig.~\ref{fig:probe-si} \textit{D}--\textit{F}).

\subsection{Uncovering a context-independent conceptual structure}
\label{sec:results-converge}

\begin{figure}[t!]
\centering
\includegraphics[width=.95\textwidth]{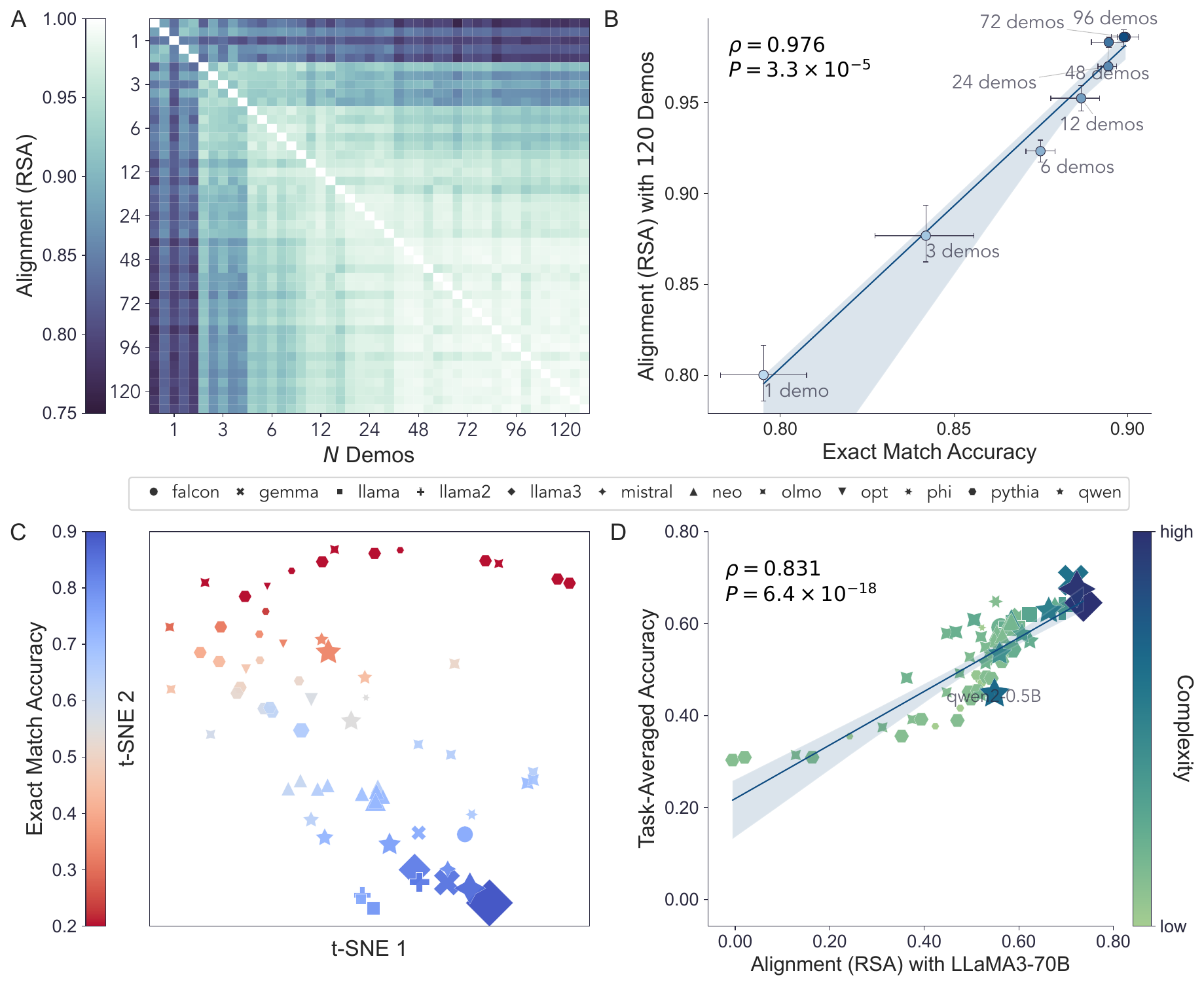}
\caption{LLMs converge toward a similar representational structure of concepts. (\textit{A}) Pairwise alignment (RSA) of LLaMA3-70B conceptual representations across different contextual demonstrations. Axes indicate the number of demonstrations, and each cell shows the alignment from a single run. (\textit{B}) LLM performance on the reverse dictionary task reflects alignment with the representations formed with 120 demonstrations. Each point corresponds to a representation space formed with $N$ demonstrations, with the x-axis indicating exact match accuracy and the y-axis denoting alignment with the 120-demonstration space. Error bars indicate $95\%$ confidence intervals from 10,000 bootstrap resamples across five runs. (\textit{C}) A t-SNE visualization of conceptual representations formed by different LLMs with 24 demonstrations. Distances are calculated as $1 - \textrm{alignment}$, averaged over five runs. Each point corresponds to an LLM, plotted in proportion to its complexity and color-coded by exact match accuracy on the reverse dictionary task. Better-performing models (blue) exhibit more aligned representations. (\textit{D}) Alignment with LLaMA3-70B conceptual representations predicts model performance across downstream tasks. The x-axis shows the alignment with LLaMA3-70B representations, and the y-axis indicates overall performance, computed as accuracy averaged across a range of language understanding and reasoning tasks. Each point represents a different model and is color-coded by model complexity. Linear fits are shown as straight lines, with shaded areas representing $95\%$ confidence intervals derived from 10,000 bootstrap resamples. Overall, models with higher complexity align more closely with the LLaMA3-70B. An exception is Qwen2-0.5B, which shows relatively high complexity but low alignment, possibly due to constraints in model scale or training data quality.}\label{fig:converge}
\end{figure}

To uncover how concepts are represented and organized within LLMs, we looked into the representation spaces they constructed for concept inference and analyzed the interrelationships among the conceptual representations. We characterized the representation spaces formed under different contexts by their relational structure, captured through a (dis)similarity matrix, and measured pairwise alignment by correlating these matrices~\citep{SHEPARD19701, kriegeskorte_representational_2008} (\textit{Materials and Methods}). For concepts in the THINGS database, the alignment between representation spaces gradually increased as the number of demonstrations rose from 1 to 24, with diminishing gains beyond this threshold (Fig.~\ref{fig:converge}\textit{A}). The alignment with the space formed by 120 demonstrations increased from $0.800$ ($\pm 0.017$) with one demonstration to $0.970$ ($\pm 0.003$) after 24 demonstrations. This alignment demonstrated a strong correlation with the model's exact match accuracy on concept inference ($\rho = 0.976$, $P < 0.0001$, $95\%$ CI: $0.730$--$1.000$) (Fig.~\ref{fig:converge}\textit{B}). These results suggest that LLMs are able to construct a coherent and context-independent relational structure, which is reflected in their concept inference capacity. Additional metrics corroborated this conclusion (\textit{SI Appendix}, section~\ref{si:results-converge}), highlighting that intricate relationships among concepts are consistently preserved within LLMs' representation spaces across varying contexts. This invariance supports the generalization of knowledge encoded in these relationships.

To examine whether different LLMs trained for language prediction develop similar conceptual representations, we compared the representation spaces they formed based on 24 demonstrations. Using t-SNE~\citep{maaten_visualizing_2008}, we visualized pairwise alignments between models and observed that LLMs with over $70\%$ exact match accuracy on concept inference clustered closely, whereas those with accuracy below $50\%$ exhibited greater dispersion (Fig.~\ref{fig:converge}\textit{C}). This indicates that better-performing models share more similar relational structure of concepts, while those with weak performance diverge in distinct ways.

We further conducted a correlational analysis to assess how the relational structure relates to model behavior more broadly. Models whose structures aligned more closely with that of LLaMA3-70B achieved higher accuracy not only on the reverse dictionary task ($\rho = 0.870$, $P < 0.0001$, $95\%$ CI: $0.756$--$0.940$) but also across a diverse set of language understanding and reasoning tasks ($\rho = 0.831$, $P < 0.0001$, $95\%$ CI: $0.699$--$0.916$; Fig.~\ref{fig:converge}\textit{D}; see \textit{SI Appendix}, Fig.~\ref{fig:converge-tasks-si} for results on each task). Additionally, when quantifying model complexity based on scale and training data (\textit{Materials and Methods}), we found that higher-complexity LLMs tended to align better with LLaMA3-70B, though exceptions likely stemmed from constraints in model scale or training data quality (Fig.~\ref{fig:converge}\textit{D}). These findings suggest that, with sufficient model scale and extensive, high-quality training data, LLMs can converge toward a shared conceptual structure. This convergence is predictive of their performance on understanding and reasoning tasks, and arises naturally from the objective of language prediction, without requiring real-world reference.

\subsection{Predicting various facets of human concept usage}
\label{sec:results-human-behavior}

\begin{figure}[t!]
\centering
\includegraphics[width=\textwidth]{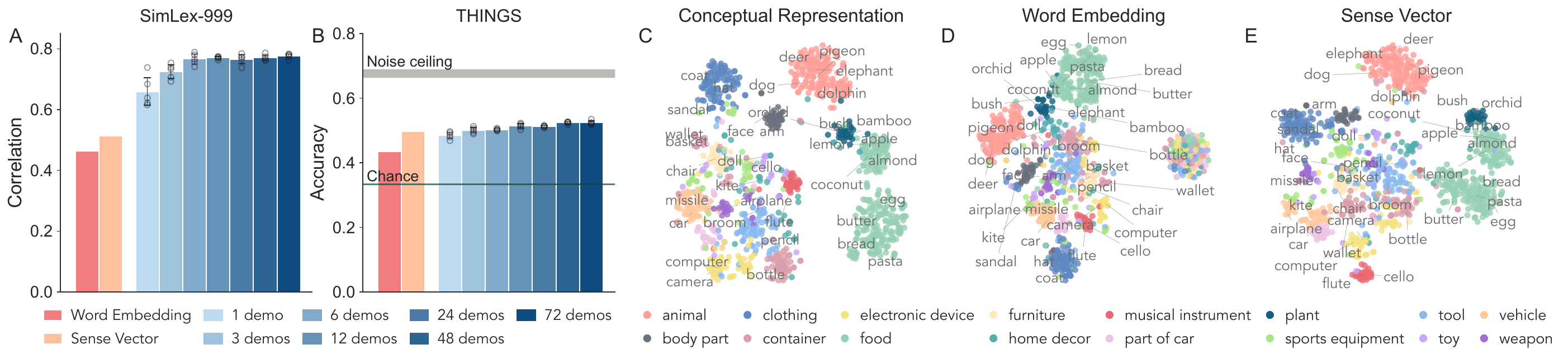}
\caption{Alignment between LLM-derived conceptual representations and psychological measures of similarity. (\textit{A}--\textit{B}) Performance of LLaMA3-70B conceptual representations compared with static word embeddings (FastText) and sense vectors (DeConf) in predicting human similarity judgments. (\textit{A}) Spearman's rank correlation with human similarity ratings for concept pairs from SimLex-999. (\textit{B}) Prediction accuracy for human triplet odd-one-out judgments in THINGS. The noise ceiling reflects the upper bound of performance based on inter-subject consistency. Error bars denote $95\%$ confidence intervals calculated from five independent runs. (\textit{C}--\textit{E}) t-SNE visualizations of LLM-derived conceptual representations (\textit{C}), static word embeddings (\textit{D}) and sense vectors (\textit{E}). Data points are color-coded by human-labeled categories from THINGS. Conceptual representations derived from LLaMA3-70B exhibit a clearer alignment with category structure than word embeddings or sense vectors.}\label{fig:similarity}
\end{figure}

Next, we investigated how well the LLM-derived conceptual representations align with different aspects of human concept usage. We tested this by using these representations to predict human behavioral data across three key psychological phenomena: similarity judgments, categorization, and gradient scales along various features. To contextualize performance, we compared against established embeddings as baselines, including FastText~\citep{grave-etal-2018-learning}, GloVe~\citep{pennington_glove_2014}, word2vec~\citep{mikolov_distributed_2013}, BERT~\citep{cassani2023meaning}, and DeConf sense vectors~\citep{pilehvar_de_2016} (\textit{Materials and Methods}).

For similarity judgments, we evaluated LLM-derived conceptual representations against human similarity ratings for concept pairs from SimLex-999~\citep{hill_simlex_2015}, which has proven challenging for traditional word embeddings. We also complemented this analysis using human triplet odd-one-out judgments from the THINGS-data collection~\citep{hebart_things-data_2023}, which relied on image-based rather than text-based stimuli and introduced contextual effects through the third concept. As shown in Fig.~\ref{fig:similarity}\textit{A}, similarity scores derived from LLM representations correlated strongly with human ratings in SimLex-999, with correlations improving as the number of contextual demonstrations increased ($\rho=0.776 \pm 0.007$ with 72 demonstrations across five runs, $P<0.0001$). These scores substantially exceeded those obtained with widely used embeddings, with FastText word embeddings achieving $\rho = 0.464$ ($P<0.0001$) and sense vectors $\rho = 0.513$ ($P<0.0001$). For THINGS triplets, we calculated pairwise similarities to identify the odd-one-out (\textit{Materials and Methods}). LLM prediction accuracy similarly improved with more contextual demonstrations, plateauing at $52.41\%$ ($\pm 0.31\%$) after 48 demonstrations (Fig.~\ref{fig:similarity}\textit{B}) and significantly exceeding FastText ($43.49\%$) and sense vectors ($49.71\%$). Across both datasets, LLM-derived representations also consistently outperformed other baselines (\textit{SI Appendix}, Fig.~\ref{fig:similarity-si}). These results indicate that, within proper context, the conceptual representations formed by LLMs effectively support the computation of similarities, a core property of human concepts. Nonetheless, performance remained below the noise ceiling estimated from inter-subject agreement ($67.67\% \pm 1.08\%$) for odd-one-out judgments. Error analyses revealed that LLMs struggled most when human choices relied primarily on perceptual properties related to color or texture (\textit{SI Appendix}, section~\ref{si:results-behavior-error}; Fig.~\ref{fig:rep-ana-limit}\textit{A}), suggesting a divergence from human judgments in these visually grounded dimensions.

\begin{figure}[t!]
\centering
\includegraphics[width=\textwidth]{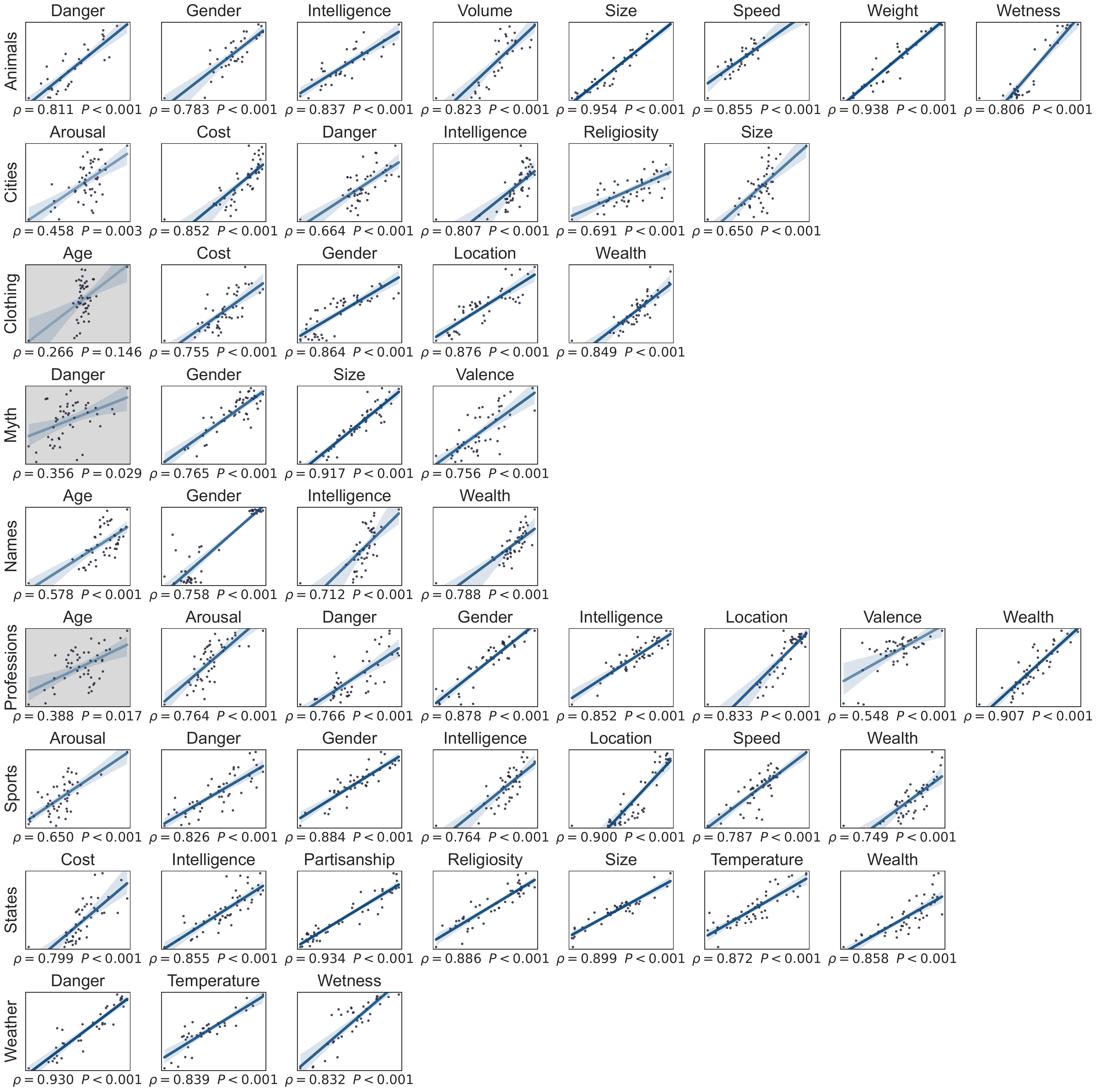}
\caption{Performance of conceptual representations derived from LLaMA3-70B in predicting context-dependent human ratings across 52 category--feature pairs. Scatter plots illustrate the relationship between predicted ratings from conceptual representations (x-axis) and the average human ratings (y-axis). Linear fits are shown as straight lines, with shaded regions representing $95\%$ confidence intervals derived from 10,000 bootstrap resamples. Category--feature pairs with statistically significant correlations (Spearman's rank correlation, FDR $P < 0.01$) are displayed against a white background.}\label{fig:sem-proj-rep-all}
\end{figure}

We then examined whether categories can be induced from the relative similarity between LLM-derived conceptual representations, using the human-labeled high-level categories from THINGS~\citep{stoinski_thingsplus_2024}. Applying a prototype-based categorization approach (\textit{Materials and Methods}), we observed that LLM-derived representations consistently achieved high accuracy, reaching $90.64\%$ ($\pm 0.28\%$) with only 24 demonstrations, significantly outperforming baseline embeddings (FastText $77.86\%$, sense vectors $85.13\%$). An alternative exemplar-based approach yielded comparable results (see \textit{SI Appendix}, section~\ref{si:results-categorization}; Fig.~\ref{fig:category} for full results, including those with an earlier version of the THINGS category set~\citep{hebart_things_2019} and all baseline comparisons), suggesting that LLM-derived representations support categorization under both views of concepts. A t-SNE visualization (Fig.~\ref{fig:similarity} \textit{C}--\textit{E}) showed that the LLM-derived conceptual representations formed clear clusters corresponding to high-level categories, such as animals and food, and also captured broader distinctions, separating natural and animate concepts from man-made and inanimate ones. Within these groupings, human body parts clustered more closely with man-made objects than other animals do, while processed food appeared closer to natural and animate objects. These patterns align with previous findings on human mental representations~\citep{hebart_revealing_2020} and are especially pronounced in LLM representations, underscoring their meaningful correspondence with human knowledge.

\begin{figure}[t!]
\centering
\includegraphics[width=.6\textwidth]{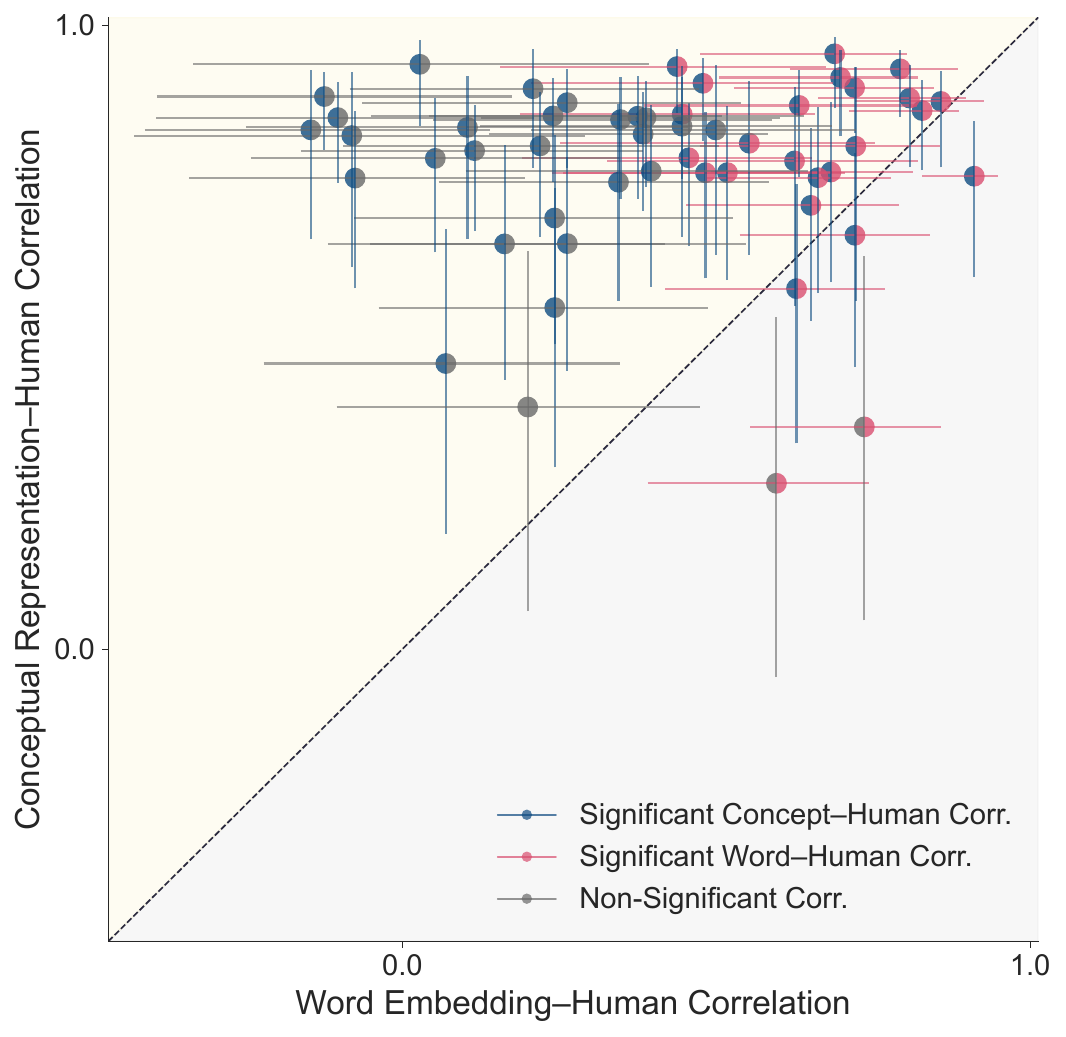}
\caption{Comparison of LLM-derived conceptual representations and static word embeddings in predicting context-dependent human ratings. The x-axis represents correlations with human ratings based on static word embeddings, while the y-axis shows correlations based on conceptual representations derived from LLaMA3-70B. Colored points indicate significant correlations (Spearman's rank correlation, FDR $P < 0.01$). Error bars denote $95\%$ confidence intervals estimated from 10,000 bootstrap resamples.}\label{fig:sem-proj-concept-vs-word}
\end{figure}

Finally, we investigated whether LLM-derived conceptual representations could capture gradient scales of concepts along various features. For example, on a scale of 1 to 5 relative to other animals, \textsc{cheetahs} might rank a 5 for speed but fall closer to 3 for size. Using LLM-derived representations for such ratings, we predicted human behavioral data across 52 category--feature pairs (e.g., animals rated for speed)~\citep{grand_semantic_2022} (\textit{Materials and Methods}). The results (Fig.~\ref{fig:sem-proj-rep-all}) revealed strong correlations with human ratings ($\rho > 0.5$, FDR $P < 0.001$) for 48 out of 52 category--feature pairs, with a median correlation of $0.817$ ($95\%$ CI: $0.777$--$0.851$) across all pairs. When accounting for the split-half reliability of human ratings (median $\rho = 0.954$), the median correlation reached $0.871$ ($95\%$ CI: $0.841$--$0.891$). Among the three category--feature pairs without statistically significant correlations (FDR $P > 0.01$), two exhibited marginal significance (FDR $P < 0.05$), while the weakest correlation appeared in the rating of clothing by age. As shown in Fig.~\ref{fig:sem-proj-concept-vs-word}, the LLM conceptual representations outperformed the strongest baseline embeddings~\citep{grand_semantic_2022} across most category--feature pairs. This advantage remained robust after excluding concepts with extreme feature values from the correlation analysis, confirming that the success of LLMs is not driven by outlier items (see \textit{SI Appendix}, section~\ref{si:results-sem-proj}; Figs.~\ref{fig:sem-proj-rep-rme}--\ref{fig:sem-proj-concept-vs-word-rme} for detailed results and full baseline comparisons). Across all three psychological phenomena, the LLM-derived conceptual representations advantageously handle intricate human knowledge spanning diverse object categories and features, highlighting their promise as computational tools for advancing our understanding of conceptual representation in the human mind.

\subsection{Mapping to activity patterns in the human brain}
\label{sec:results-brain}

\begin{figure}[htp]
\centering
\includegraphics[width=\textwidth]{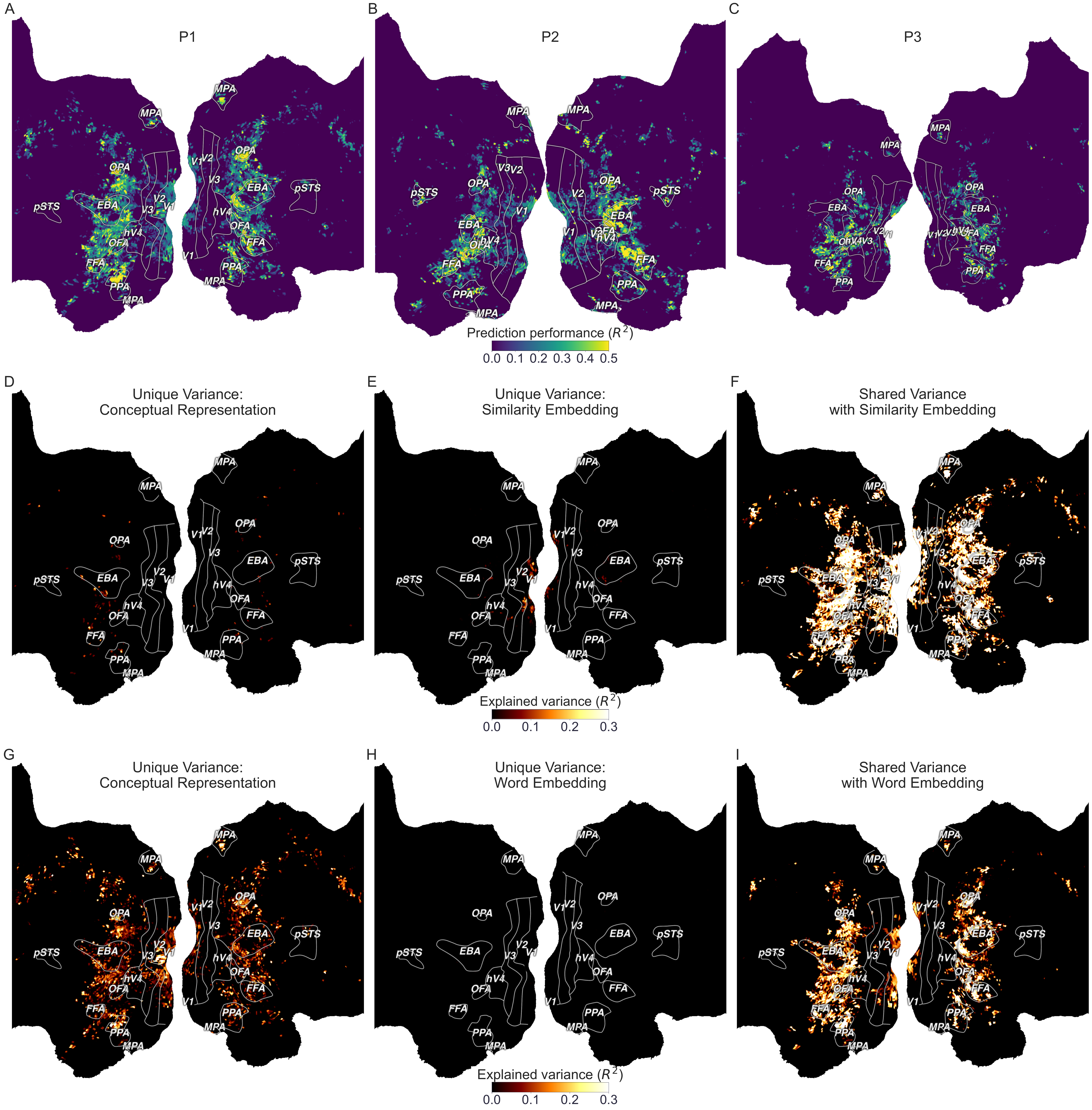}
\caption{Prediction performance of LLM-derived conceptual representation (LLaMA3-70B) in voxel-wise encoding and comparisons with baseline models. (\textit{A}--\textit{C}) Prediction performance of LLM-derived conceptual representation visualized on cortical maps for three individual participants. (\textit{D}--\textit{F}) Comparison between LLM-derived conceptual representation and a similarity embedding learned from human similarity judgments. (\textit{D}) Variance uniquely explained by the LLM-derived conceptual representation. (\textit{E}) Variance uniquely explained by the human-derived similarity embedding. (\textit{F}) Shared variance explained by both models. (\textit{G}--\textit{I}) Comparison between LLM-derived conceptual representation and a static word embedding (FastText). (\textit{G}) Variance uniquely explained by the LLM-derived conceptual representation. (\textit{H}) Variance uniquely explained by the static word embedding. (\textit{I}) Shared variance explained by both models. Color intensity indicates the proportion of explained variance, normalized relative to the noise ceiling and averaged over five runs. Only voxels with statistically significant prediction performance are shown (FDR $P < 0.01$).}\label{fig:mri-encoding}
\end{figure}

We further explored the biological plausibility of LLM-derived conceptual representations by mapping them onto the activity patterns in the human brain. Using fMRI data from the THINGS-data collection~\citep{hebart_things-data_2023}, we fitted a voxel-wise linear encoding model to predict neural responses evoked by viewing concept images, based on the corresponding conceptual representations derived from LLaMA3-70B (\textit{Materials and Methods}). Fig.~\ref{fig:mri-encoding} \textit{A}--\textit{C} shows the prediction performance maps, indicating voxels where the predicted activations best correlated with actual activations (FDR $P < 0.01$). The LLM-derived conceptual representations successfully predicted activity patterns across widely distributed brain regions, encompassing the visual cortex and beyond. In particular, category-selective regions were more strongly represented, including the lateral occipital complex (LOC), occipital face area (OFA), fusiform face area (FFA), parahippocampal place area (PPA), extrastriate body area (EBA) and medial place area (MPA). These patterns are consistent with prior work suggesting that abstract semantic information was primarily represented in the higher-level visual cortex~\citep{dicarlo_untangling_2007, schindler_visual_2016}. Meanwhile, significant prediction performance was observed in early visual regions, including V1, indicating that information processed in low-level visual areas is also relevant and can be effectively inferred by LLMs trained exclusively on language data. The consistent location of informative voxels across participants supports the generality of our findings (\textit{SI Appendix}, Fig.~\ref{fig:mri-encoding-corr}).

To better elucidate the alignment between LLM-derived conceptual representations and neural coding of concepts, we compared them with alternative embeddings including a similarity embedding optimized to account for human similarity judgments of the THINGS concepts~\citep{hebart_things-data_2023}, FastText word embeddings, and DeConf sense vectors. For each baseline, we combined it with the LLM-derived representations and applied variance partitioning to disentangle their unique and shared contributions (\textit{Materials and Methods}).

As shown in Fig.~\ref{fig:mri-encoding} \textit{D}--\textit{F}, LLM-derived conceptual representations and the human-derived similarity embedding shared a substantial portion of explained variance, particularly within higher-level visual areas, with some overlap in early visual regions. This indicates that both accounted well for neural responses associated with visual concepts. The variance uniquely explained by the LLM-derived representations was modest but measurable, primarily localized to high-level visual areas (Fig.~\ref{fig:mri-encoding}\textit{D}). By comparison, the human-derived similarity embedding uniquely accounted for more variance in early visual regions including V1 and V2 (Fig.~\ref{fig:mri-encoding}\textit{E}). These results suggest that LLM-derived representations captured key aspects of the brain's encoding of concepts, especially those related to higher-level and abstract content. Nevertheless, certain aspects of visual information that are prominent in early visual regions and relevant to human behavior may be underrepresented in models trained solely on language.

Compared with FastText word embeddings, LLM-derived conceptual representations alone accounted for considerably more variance across the visual system (Fig.~\ref{fig:mri-encoding}\textit{G}), whereas word embeddings contributed little unique variance (Fig.~\ref{fig:mri-encoding}\textit{H}). The variance shared between the two was also less extensive than that observed between LLM-derived representations and the human-derived similarity embedding (Fig.~\ref{fig:mri-encoding}\textit{I}). These results indicate that LLM-derived representations capture richer and more nuanced information than word embeddings, suggesting that neural encoding of concepts extends beyond lexical form. A similar pattern was observed for sense vectors (\textit{SI Appendix}, section~\ref{si:results-brain}; Fig.~\ref{fig:mri-encoding-varpart-all} \textit{G}--\textit{I}). Additionally, LLM representations derived with more demonstrations accounted for a small increase in unique variance, indicating that consistently structured representations more faithfully map onto brain activity patterns (\textit{SI Appendix}, section~\ref{si:results-brain}; Fig.~\ref{fig:mri-encoding-varpart-all} \textit{J}--\textit{L}). Together, these findings support the biological plausibility of LLM-derived conceptual representations and underscore their relevance for understanding how the brain abstracts and processes visual---and potentially other sensory---information. A complementary variance partitioning approach based on orthogonalization yielded qualitatively similar results (\textit{SI Appendix}, Figs.~\ref{fig:mri-encoding-varpart-ortho-all}--\ref{fig:mri-encoding-varpart-ortho-llm_1demo}), corroborating these conclusions.

\section{Discussion}

In this paper, we demonstrated that next-token prediction over language naturally gives rise to human-like conceptual representations and organization, even without real-world grounding. Our work builds upon the long-explored idea of vectors as conceptual representations~\citep{SHEPARD19701, shepard_toward_1987, mcclelland_parallel_2003, piantadosi_why_2024}, while previous work has predominantly focused on word embeddings~\citep{grand_semantic_2022, lake_word_2023}. We viewed concepts as latent representations used for word generation and guided LLMs to infer them from definitional descriptions. Our findings revealed that LLMs can adaptively derive concepts based on contextual demonstrations, reflecting the interrelationships among them. These representations converged toward a context-independent relational structure predictive of model performance across downstream tasks, suggesting that language prediction inherently fosters the development of a shared conceptual structure. This structure supports the generalization of knowledge by effectively capturing key properties of human concepts, such as similarity judgments, categorical distinctions and gradient scales along various features. Notably, the conceptual representations showed a strong alignment with neural coding patterns observed in the human brain, even in response to non-linguistic visual stimuli. These findings suggest that LLMs offer a tangible window into human conceptual representation and organization, underscoring the central role of conceptual structure in understanding AI systems and in advancing their alignment with human cognition.

Concepts are considered as mental representations that abstract away specific details and enable flexible generalization in novel situations~\citep{murphy_big_2004, margolis_conceptual_2015, mitchell_abstraction_2021}. Using the reverse dictionary task, we showed that LLMs can effectively derive such conceptual representations from definitional descriptions. While word embeddings have shown potential to capture certain properties required for conceptual representations~\citep{hill-etal-2016-learning-understand, grand_semantic_2022, pavlick_symbols_2023}, their capacity is constrained by the inherent context-sensitivity of words, which do not correspond to concepts in a straightforward way~\citep{murphy_big_2004, malt_where_2015}. Accordingly, word embeddings are either limited by their static nature---failing to account for context-dependent nuances---or their contextual variability, which precludes consistent mapping to distinct concepts. In contrast, the conceptual representations derived from informative descriptions bypass the ambiguity of words and can be consistently mapped to appropriate terms despite contextual variations (Fig.~\ref{fig:relation}). We argue that these representations are abstract, as their relational structures were consistently preserved across varying contexts (Fig.~\ref{fig:converge} \textit{A}--\textit{B}). This consistency indicates that the representations capture the underlying relationships among concepts while discarding surface-level details, a hallmark of abstraction. Such abstraction is essential for generalization, enabling the learned relationships to be flexibly adapted to novel situations. Similar abstract, though disentangled, representations have been observed in humans~\citep{courellis_abstract_2024}, monkeys~\citep{bernardi_geometry_2020}, rodents~\citep{boyle_tuned_2024}, and neural networks trained for multitasking~\citep{johnston_abstract_2023}. However, the focus on task-specific low-level features is insufficient to model the richness of broadly generalizable real-world concepts. Our results, spanning diverse LLM architectures (Fig.~\ref{fig:converge} \textit{C}--\textit{D}), reveal that abstract representations encompassing a wide array of real-world concepts can emerge solely from language prediction. This highlights a promising pathway toward modeling the complexity of human conceptual repertoire.  

Relationships among concepts have long been a cornerstone of cognitive theories~\citep{SHEPARD19701, shepard_toward_1987, fodor_connectionism_1988}. Here, we showed that language prediction naturally gives rise to interrelated concepts. This is demonstrated through the concept inference behavior of LLMs, which was shaped by contextual cues rather than occurring in isolation (Fig.~\ref{fig:relation}), and the consistently preserved relational structure within their representation spaces (Fig.~\ref{fig:converge} \textit{A}--\textit{B}). Comparisons with human data further revealed that these relationships aligned with psychological measures of similarity and encapsulated a wealth of human conceptual knowledge (Fig.~\ref{fig:similarity} and Fig.~\ref{fig:sem-proj-rep-all}). According to conceptual role semantics~\citep{greenberg_crs_2005, block_conceptual_1996}, the meaning of a concept is determined by its relationships to other concepts and its role in thinking, rather than reference to the real world. It has been claimed that LLMs may possess human-like meaning in this sense, with language serving as a valuable source for inferring how people use concepts in thought~\citep{piantadosi_meaning_2022}. However, evidence was needed to determine whether the objective of language prediction can lead to the discovery of the right conceptual roles. Our results support this claim, showing that the relationships among LLM-derived conceptual representations approximate human meaning. While these relationships could be further refined to match those of humans, particularly with respect to real-world grounding, compositionality and abstract reasoning~\citep{lake_word_2023, mcclelland_placing_2020}, our findings suggest that LLMs offer a prospective computational footing for implementing conceptual role semantics, paving the way for the development of more human-like, concept-based AI systems.

Moreover, we observed that LLMs converged on a shared relational structure of concepts that reliably predicted downstream performance (Fig.~\ref{fig:converge} \textit{C}--\textit{D}), aligning with the recently proposed ``Platonic Representation Hypothesis''~\citep{huh_position_2024}. This hypothesis posits that AI models, despite differing training objectives, data, and architectures, will converge on a universal representation of reality. Prior work has hinted at such convergence, showing alignment among models trained on different modalities~\citep{huh_position_2024} and between models and neural activity patterns in the human brain~\citep{wang_better_2023, schrimpf_neural_2021}. Our results reveal the representational structure of concepts emerging solely from language prediction and demonstrate substantial alignment between these representations and neural responses to visual stimuli (Fig.\ref{fig:mri-encoding}). This convergence lays the groundwork for concept-based alignment across modalities, between AI systems, and between AI and humans~\citep{rane_concept_2024}, and sets the stage for future research on the interaction between linguistic and perceptual systems.

Theories of concepts have identified various properties that conceptual representations must satisfy, highlighted by the symbolism--connectionism divide. Our results show that LLM-derived conceptual representations successfully reconcile the seemingly competing properties, integrating the definitions, relations and structures emphasized by the symbolic approaches with the graded and continuous nature of neural networks. These representations were structured in a way that their relationships supported straightforward computations of human-like similarities, categories (Fig.~\ref{fig:similarity}; \textit{SI Appendix}, section~\ref{si:results-categorization}) and gradient distinctions (Fig.~\ref{fig:sem-proj-rep-all}). Previous work has employed distributed word embeddings to approximate these aspects of human concept usage, revealing preliminary correspondence~\citep{hill_simlex_2015, hill-etal-2016-learning-understand, grand_semantic_2022}. However, these embeddings have exhibited inconsistent alignment with human similarity judgments across various datasets~\citep{lake_word_2023}. Our results align with previous findings, showing that word embeddings primarily reflect association or relatedness but fail to capture genuine similarity (Fig.~\ref{fig:similarity} \textit{A}--\textit{B}; \textit{SI Appendix}, section~\ref{si:results-similarity}). In contrast, LLM-derived conceptual representations demonstrated distinctive strengths. They performed well in modeling human similarity judgments, including those based on images, suggesting that they advantageously capture conceptual information extending beyond specific stimulus modalities. They also supported flexible, context-dependent reorganization to reflect gradient scales along different features, surpassing word embeddings in both breadth and depth. Importantly, these representations exhibited superior alignment with brain activity patterns, especially in high-level visual areas associated with abstract semantic information (Fig.~\ref{fig:mri-encoding} \textit{D}--\textit{I}), supporting the view that vector-based models offer a plausible framework for capturing the neural activations underlying cognitive processes~\citep{mcclelland_parallel_2003, mcclelland_letting_2010}. Collectively, these findings underscore the potential of LLM-derived conceptual representations to reconcile symbolic and connectionist approaches and to illuminate the representational basis of human concepts.

Despite this promise, our results also reveal notable divergences from human behavioral and neural patterns. LLM-derived conceptual representations fell below the noise ceiling in predicting human similarity judgments and accounted for less of the variance in low-level visual regions than embeddings derived from human similarity judgments. To probe this discrepancy, we regressed the human-derived similarity embedding onto the LLM-derived representations and found the most pronounced gaps in dimensions related to color, followed by texture and shape (\textit{SI Appendix}, section~\ref{si:results-brain-limitations}; Fig.~\ref{fig:rep-ana-limit}\textit{B}). These gaps closely mirrored the error profile of the LLM-derived representations in behavioral predictions and mapped onto neural activity in low-level visual areas (\textit{SI Appendix}, section~\ref{si:results-brain-limitations}; Figs.~\ref{fig:rep-ana-limit}, \ref{fig:mri-encoding-varpart-ortho-similarity}), highlighting a key limitation in capturing visually grounded perceptual properties. These findings parallel previous research showing that congenitally blind individuals acquire these properties primarily through inference and differ markedly from sighted people in their knowledge of color~\citep{kim_shared_2021, kim_knowledge_2019}. Such information, therefore, may be inefficient to learn from language alone, or even entirely absent.

Our work provides a foundational framework for exploring emergent conceptual representations from language prediction, positioning LLMs as powerful tools for advancing theories of concepts. While our approach to deriving conceptual representations from LLMs exhibits promising generalizability across datasets, tasks, and models, the human conceptual repertoire is far richer and more nuanced than what we have examined. Extending this approach to more diverse domains---such as abstract concepts, compositional structures, and cross-linguistic variation---and systematically investigating the factors that drive convergence and divergence between LLM and human representations~\citep{mahner_dimensions_2025} will be essential for fully characterizing model competence and limitations. Moreover, although the structure of the conceptual representations correlates strongly with downstream performance, LLMs operate at the token level and do not necessarily draw on these representations during typical text generation. This disconnect may underlie their observed limitations in certain aspects of concept usage, including compositionality~\citep{lake_word_2023} and reasoning~\citep{mitchell_abstraction_2021, binz_using_2023}.

Future work could aim to build better models of human cognition by incorporating more cognitively plausible incentives~\citep{irie_neural_2024} such as systematic generalization~\citep{lake_human-like_2023} and reasoning~\citep{akyurek_surprising_2024}, while explicitly guiding models to leverage conceptual representations, rather than linguistic forms, to generalize across tasks. Recent progress in steering LLMs to operate within their representation spaces has shown promise for enhancing language generation and reasoning~\citep{thelcmteam_large_2024, hao_training_2024}. Efforts in this direction could narrow the gaps between LLMs and human conceptual abilities and help elucidate their current limitations in compositionality~\citep{lake_word_2023}, reasoning~\citep{binz_using_2023, mitchell_abstraction_2021} and over-sensitivity to minor contextual shifts~\citep{binz_using_2023, mccoy_embers_2024}. Moreover, the LLM-derived conceptual representations could be enriched with information from diverse sources, like vision~\citep{vong_grounded_2024, mcclelland_placing_2020}, to better align with human cognition~\citep{rane_concept_2024} and foster human-machine collaboration~\citep{stolk_conceptual_2016, collins_building_2024}. Finally, incorporating brain data beyond the visual domain would offer a richer understanding of the neural underpinnings of conceptual representations in the human mind. Despite current limitations in models and data, the emergence of human-like conceptual representations within LLMs marks a critical step toward resolving enduring questions in the science of human concepts. This progress opens new avenues for bridging the gaps between human and machine intelligence, offering valuable insights for cognitive science and artificial intelligence.

\section{Materials and Methods}\label{methods}

\subsection{Large language models used in our experiments}
\label{sec:methods-llms}

This paper focuses on base models---LLMs pretrained solely for next-token prediction without additional fine-tuning or reinforcement learning. We used only open-source LLMs, as their hidden representations are necessary for our experiments. Primary analysis were conducted on LLaMA 3 models, including LLaMA3-70B and LLaMA3-8B, each pretrained on over 15 trillion tokens~\citep{dubey_llama_2024}. Another 65 Transformer-based decoder-only LLMs from various model series were also employed for experiments on the reverse dictionary task and representational convergence. Further details regarding model names, scales, training data and sources can be found in \textit{SI Appendix}, Tables~\ref{tab:llms-details}--\ref{tab:llms-details-checkpoints}. These models vary in architecture, scale, and pretraining data, enabling exploratory analyses of how these factors might shape the conceptual representations within LLMs.

\subsection{Details on deriving conceptual representations}
\label{sec:methods-rev-dict}

We used data from the THINGS database~\citep{hebart_things_2019} to probe LLMs' conceptual representations via the reverse dictionary task. The dataset includes 1,854 concrete and nameable object concepts, paired with their WordNet synset IDs, definitional descriptions, several linked images and human-labeled category membership. The concepts and images were selected to be representative of everyday objects commonly used in American English, providing a useful testbed for analyzing model representations. The dataset is available at \url{https://osf.io/jum2f/}. To assess the generality of our results, we extended our analysis to a broader range of descriptions and concepts. Specifically, (1) we tested LLMs' generalizability across a broader set of 21,402 concepts, spanning different word classes (nouns, verbs, adjectives and adverbs), age-of-acquisition~\citep{kuperman_aoa_2012}, and degrees of concreteness~\citep{brysbaert_concreteness_2014}, all intersecting with Wordnet~\citep{fellbaum_wordnet_1998}. (2) We evaluated LLMs' prediction consistency using three additional distinct definitional descriptions for each THINGS concept, which were generated by GPT-3.5 and manually checked for diversity (see \textit{SI Appendix}, section~\ref{si:methods-gpt-desc} for details). (3) We examined LLMs' sensitivity to linguistic structure by introducing varying degrees of word order permutations to the query descriptions. We shuffled $30\%$, $60\%$ and $100\%$ of the words in the query descriptions and reinserted them into the original text.

To guide LLMs in the reverse dictionary task~\citep{xu_tip_2024}, we selected a random $20\%$ subset of concepts as the training set (from THINGS for most analyses, and from WordNet for the 21,402-concept extension), with the remaining concepts used for testing. From the training set, $N$ description--word pairs were randomly chosen as demonstrations. Model performance was evaluated based on strict exact matches across five independent runs, each with a unique random sample of $N$ demonstrations. For each test concept, we prompted an LLM with a specific description followed by the arrow symbol ``$\Rightarrow$'' and truncated the output at the first newline character (``\texttt{\textbackslash n}''). We then assessed whether the resulting output matched the expected word or any listed synonyms in THINGS (or WordNet). We opted for greedy search as our decoding method for a straightforward and equitable comparison across models. For subsequent representational analyses, we extracted conceptual representations from the penultimate layer of LLMs at the arrow symbol ``$\Rightarrow$'', which directly yielded subsequent predictions and bridges phrasal and lexical terms within the models.

To probe the role of contextual information in concept inference, we examined how modifying the description--word pairings in context affected model predictions. For each target concept, we paired its description with a proxy symbol and combined it with $N - 1$ correct description--word pairs from other concepts. We then queried the LLM using the same description. Model performance was evaluated based on how often the LLM replicated the proxy symbol or generated the correct word for the target concept. We tested four types of proxy symbols: (1) a random uppercase English letter, excluding ``A'' and ``I'' to avoid potential semantic associations; (2) a randomly generated lowercase letter string, with length sampled from a shifted Poisson distribution ($\textrm{mean} = 6.94$, $\textrm{variance}$ = 5.80) to approximate typical English word lengths~\citep{rothschild_distribution_1986}; (3) a random word selected from the THINGS database, distinct from the target concept and absent from the current context; and (4) the correct word for the target concept, used as a reference.

\subsection{Structural analysis of representation spaces}
\label{sec:methods-rsa}

We used representational similarity analysis (RSA)~\citep{kriegeskorte_representational_2008} to measure the alignment between conceptual representations derived under different conditions (i.e., across contexts or models), which is non-parametric and has been widely adopted to measure topological alignment between representation spaces~\citep{kriegeskorte_peeling_2019}. Let $X \in \mathbb{R}^{m \times d_{1}}$ and $Y \in \mathbb{R}^{m \times d_{2}}$ denote two sets of conceptual representations for $m$ concepts with dimensionality of $d_{1}$ and $d_{2}$, respectively. Each space is characterized by its relational structure, represented by a (dis)similarity matrix $M \in \mathbb{R}^{m \times m}$, where the entry $M_{i,j}$ denotes the (dis)similarity between the representations of the $i^{\textrm{th}}$ and $j^{\textrm{th}}$ concepts in the corresponding space. The alignment can then be calculated as the Spearman's rank correlation between the upper (or lower) diagonal portion of the two matrices $M_{X}$ and $M_{Y}$, yielding values ranging from $-1$ to $1$. 

An appropriate similarity function is needed for the (dis)similarity matrix to characterize the relational structure of the representation space. While cosine similarity has been widely adopted since the advent of static word embeddings, it is sensitive to outliers and may be suboptimal for capturing non-linear relationships~\citep{zhelezniak-etal-2019-correlation}. We thus employed two metrics: cosine similarity and Spearman's rank correlation. Comparison with human behavioral data suggests that the cosine similarity captures semantic similarity while Spearman's rank correlation reflects both similarity and association (\textit{SI Appendix}, section~\ref{si:results-similarity}; Fig.~\ref{fig:similarity-si} \textit{A}--\textit{B}). We adopted Spearman correlation as our primary similarity function, with results from cosine similarity provided in \textit{SI Appendix}, section~\ref{si:results-converge}; Fig.~\ref{fig:converge-si} \textit{A}--\textit{B}. 

Additionally, we employed the parallelism score~\citep{bernardi_geometry_2020} to evaluate if the differences between conceptual representations are preserved across conditions, which have been thought to reflect relations in the representation space~\citep{mikolov_distributed_2013}. The results complemented the RSA findings (\textit{SI Appendix}, section~\ref{si:results-converge}; Fig.~\ref{fig:converge-si} \textit{C}--\textit{D}). 

For qualitative comparison across LLMs, we visualized these models using t-SNE with a perplexity of 30 and 1000 iterations, where pairwise distances were defined as $1 - \textrm{alignment}$.

\subsection{Evaluation of model performance on downstream tasks}

We evaluated LLMs' language understanding and reasoning capabilities using a range of widely adopted multiple-choice benchmark tasks. Models were tested in a zero-shot setting by ranking candidate answers according to their assigned probabilities, with performance measured by accuracy in selecting the correct option. Full details of the benchmarks, prompt templates, and evaluation protocols are provided in \textit{SI Appendix}, section~\ref{si:methods-eval-tasks}; Table~\ref{tab:general-templates}.

\subsection{Characterization of model complexity}
\label{sec:methods-model-complexity}

To characterize the complexity of the LLMs used in our experiments, we represented each by its number of parameters and amount of training data (i.e., the number of tokens used during training)---factors identified as crucial for determining model quality~\citep{kaplan_scaling_2020}. A principal component analysis (PCA) was then applied to these factors. The first principal component, accounting for $70.67\%$ of the total variance, was used to represent model complexity. For the Mistral models, where training data volume was not accessible, we estimated it based on comparable models released around the same time.

\subsection{Prediction of human similarity judgments}
\label{sec:methods-similarity}

To assess the alignment between LLM-derived representations and psychological measures of similarity, we first compared them with human ratings for concept pairs from SimLex-999~\citep{hill_simlex_2015}, available at \url{https://fh295.github.io/simlex.html}. SimLex-999 explicitly distinguishes semantic similarity from association or relatedness and contains human ratings for 999 concept pairs spanning 1,028 concepts across nouns, verbs and adjectives. We used description--word pairs from the WordNet data to provide contextual demonstrations to the LLMs, thereby deriving conceptual representations. Model performance was assessed through Spearman's rank correlation between the similarity scores derived from the LLM representations and the human ratings.

We further employed the odd-one-out similarity judgments from the THINGS-data collection~\citep{hebart_things-data_2023} to validate the effectiveness of LLM representations in handling the computation of similarities. The dataset consists of triplets sampled from the 1,854 concepts in THINGS, accessible at \url{https://osf.io/f5rn6/}. We presented LLMs with contextual demonstrations randomly sampled from THINGS to obtain their conceptual representations. For each triplet $\left(i, j, k\right)$, we computed pairwise similarities among the corresponding conceptual representations $\left(\mathbf{h}_{i}, \mathbf{h}_{j}, \mathbf{h}_{k}\right)$ and identified the item outside the most similar pair as the odd-one-out. Model predictions were compared with individual human judgments across the validation set (over 450,000 triplets).

\subsection{Categorization}
\label{sec:methods-category}

We used high-level human-labeled natural categories in THINGS~\citep{stoinski_thingsplus_2024}. We excluded subcategories of other categories, concepts belonging to multiple categories and categories with fewer than ten concepts~\citep{hebart_revealing_2020}. This filtering resulted in 23 out of the original 53 categories, comprising 962 concepts in total (\textit{SI Appendix}, Table~\ref{tab:categories}). Categorization was evaluated using two approaches corresponding to prototype and exemplar models. For the prototype model, category membership was determined with a leave-one-out nearest-centroid classifier. In each iteration, the centroid for each category was computed by averaging the representations of the remaining concepts, and the left-out concept was assigned to the closest centroid. For the exemplar model, categorization followed a nearest-neighbor decision rule~\citep{sorscher_neural_2022} (\textit{SI Appendix}, section~\ref{si:methods-exemplar}).

We combined multidimensional scaling (MDS) with t-SNE to visualize the representations in two dimensions, thereby preserving the global structure while better capturing local similarities. Representations were first reduced to 64 dimensions using MDS, with distances calculated as $1 - \textrm{similarity}$, and then embedded in 2D using t-SNE with a perplexity of 30 and 1000 iterations.

\subsection{Prediction of gradient scales along features}
\label{sec:methods-sem-proj}

We used human ratings from a dataset spanning 52 category--feature pairs~\citep{grand_semantic_2022} (\url{https://osf.io/5r2sz/}), in which participants rated a concept (e.g., \textsc{whale}) within a certain category (e.g., animal) along multiple feature dimensions (e.g., size and danger). The dataset covers nine categories, each paired with a subset of 17 features. 
For each category--feature pair, we provided the model with two demonstrations illustrating the extreme values of a target feature within a category. We then queried the model for the rating of each concept in the category to derive its representation. 
Similar to previous work~\citep{grand_semantic_2022}, we constructed a scale vector by subtracting the representation of the minimum extreme from the maximum (e.g., $\overrightarrow{\prime\prime\textrm{size}\prime\prime_{\textrm{animal}}} = \overrightarrow{\textrm{whale}_{\textrm{size}}} - \overrightarrow{\textrm{ant}_{\textrm{size}}}$), and compared all representations to this scale to obtain their relative feature ratings. Model performance was evaluated via Spearman's rank correlation between model-derived and human ratings. We estimated $95\%$ confidence intervals for each category--feature pair using 10,000 bootstrap samples, and corrected $P$-values for multiple comparisons across the 52 pairs using the false discovery rate (FDR) method.


\subsection{Word embeddings for comparison}
\label{sec:methods-word-embs}

For predicting human concept usage, we evaluated five widely used embeddings: (1) 300-dimensional FastText word embeddings trained on Common Crawl and Wikipedia~\citep{grave-etal-2018-learning} (\url{https://fasttext.cc/docs/en/crawl-vectors.html}); (2) 300-dimensional DeConf sense vectors~\citep{pilehvar_de_2016}, which integrate word2vec embeddings with WordNet knowledge (\url{https://pilehvar.github.io/deconf/}); (3) GloVe~\citep{pennington_glove_2014}; (4) word2vec~\citep{mikolov_distributed_2013}; and (5) BERT embeddings~\citep{cassani2023meaning}. Details and results for all baselines are provided in \textit{SI Appendix}. In the main text, we reported FastText as a broadly covering, representative word embedding, and DeConf for its strong performance and word-sense representation. For gradient-scale analyses, we reported the best-performing embedding (GloVe), identified in the original study~\citep{grand_semantic_2022} and confirmed here, which also provided full coverage. For each embedding, the similarity measure was selected to maximize alignment with human behavioral data (\textit{SI Appendix}, section~\ref{si:results-similarity}). 

Word embeddings were evaluated using the same procedures as LLM-derived conceptual representations, except for the gradient-scale analyses, where static embeddings cannot handle contextual demonstrations. Following previous work~\citep{grand_semantic_2022}, scale vectors were constructed from antonym pairs denoting opposite feature values (e.g., $\overrightarrow{\textrm{big}}$ vs. $\overrightarrow{\textrm{small}}$). Each item's embedding was projected onto the corresponding scale vector, and the resulting projections were correlated with human ratings.

\subsection{Encoding model of neural representations in the brain}
\label{sec:methods-encoding-brain}

We used fMRI data from the THINGS-data collection~\citep{hebart_things-data_2023}, available at \url{https://openneuro.org/datasets/ds004192/versions/1.0.5}, to explore whether LLM-derived conceptual representations align with neural responses to visually grounded concepts. The dataset encompasses brain imaging data from three participants exposed to 8,740 images of 720 concepts over 12 sessions. These concepts, sampled from the 1,854 in THINGS, overlap with those used in our analyses of human similarity judgments. We obtained neural representations for each concept by averaging responses across its associated images. LLM representations were derived using 24 contextual demonstrations, which our analyses showed were sufficient to yield coherent and consistently structured conceptual representations.

We trained linear encoding models to predict voxel-wise activity from LLM-derived representations. Specifically, the activation at each voxel was modeled as $y_{v,c} = \mathbf{h}_{c} \mathbf{w} + b + \epsilon$, where $y_{v,c}$ is the activation at voxel $v$ for concept $c$, $\mathbf{h}_c$ is the corresponding LLM representation, $\mathbf{w}$ is a vector of regression coefficients, $b$ is a constant, and $\epsilon$ is residual error. The parameters $\mathbf{w}$ and $b$ were estimated via ridge regression, with the regularization parameter selected via cross-validation within the training set. Model performance was evaluated using 20-fold cross-validation with non-overlapping concepts, measuring the correlation between predicted and observed neural responses across folds. Statistical significance was assessed by comparing predicted correlations to a null distribution of correlation values generated from 10,000 random permutations~\citep{contier_distributed_2024}, with FDR correction applied to voxel-wise $P$-values. We computed noise-ceiling-normalized $R^{2}$ for each voxel by dividing the original $R^{2}$ by the estimated noise ceiling. The noise ceilings were derived based on signal and noise variance, estimated from variability in neural responses to repeated presentations of the same concept~\citep{allen_massive_2022} (\textit{SI Appendix}, Fig.~\ref{fig:mri-nc}).

\subsection{Variance partitioning between conceptual and alternative representations}
\label{sec:methods-var-part}

We compared LLM-derived conceptual representations against three alternatives: (1) a 66-dimensional similarity embedding~\citep{hebart_things-data_2023} trained on 4.10 million human odd-one-out judgments of concepts in THINGS (\url{https://osf.io/f5rn6/}), which aligns well with human similarity judgments; (2) static FastText embeddings~\citep{grave-etal-2018-learning}, chosen for their full coverage of the THINGS concepts; and (3) DeConf sense vectors~\citep{pilehvar_de_2016}, selected for their best performance among baselines in predicting behavioral data for the THINGS concepts. These comparisons examined the explanatory power of LLM-derived representations relative to established word embeddings and human-derived similarity spaces.

For each baseline, we reduced the dimensionality of the LLM-derived conceptual representations to match that of the baseline using singular value decomposition (SVD), providing a conservative estimate of unique variance. We then combined the two and performed variance partitioning to assess their shared and unique contributions~\citep{deHeer_hierarchical_2017, wang_better_2023}. Specifically, we predicted fMRI responses using a joint model of the two representations (denoted $\mathcal{A}$ and $\mathcal{B}$) via ridge regression with 20-fold cross-validation, yielding the total variance explained $R^{2}_{\mathcal{A}\textrm{\&}\mathcal{B}}$. We then computed the variance explained by each model individually---$R^{2}_{\mathcal{A}}$ and $R^{2}_{\mathcal{B}}$---using the same cross-validation procedure. The unique variance explained by each model was calculated as: $R^{2}_{\mathcal{A}, \textrm{unique}} = R^{2}_{\mathcal{A}\textrm{\&}\mathcal{B}} - R^{2}_{\mathcal{B}}$, $R^{2}_{\mathcal{B}, \textrm{unique}} = R^{2}_{\mathcal{A}\textrm{\&}\mathcal{B}} - R^{2}_{\mathcal{A}}$. The shared variance was defined as: $R^{2}_{\textrm{shared}} = R^{2}_{\mathcal{A}} + R^{2}_{\mathcal{B}} - R^{2}_{\mathcal{A}\textrm{\&}\mathcal{B}}$. We restricted our analyses to voxels with a noise ceiling above $5\%$ and reported the noise-ceiling-normalized $R^{2}$. For reference, results from a complementary variance partitioning approach based on orthogonalization are provided in \textit{SI Appendix}, section~\ref{si:results-brain}; Figs.~\ref{fig:mri-encoding-varpart-ortho-all}--\ref{fig:mri-encoding-varpart-ortho-llm_1demo}.

\bigskip

\backmatter

\bmhead{Supplementary information} 

The supplementary information contains supplementary text, Figures~\ref{fig:probe-si} to~\ref{fig:mri-nc}, and Tables~\ref{tab:llms-details} to~\ref{tab:categories}.

\bmhead{Data, Materials, and Software Availability}

This work builds on previously published datasets and materials~\citep{hebart_things_2019, hebart_things-data_2023, fellbaum_wordnet_1998, hill_simlex_2015, grand_semantic_2022, grave-etal-2018-learning, mahner_dimensions_2025, pilehvar_de_2016, pennington_glove_2014, mikolov_distributed_2013, cassani2023meaning}. All code and supplementary data necessary to reproduce the analyses are available on GitHub (\url{https://github.com/ningyuxu/llm_concept}), and archived on Zenodo (\url{https://doi.org/10.5281/zenodo.17092983}).

\bibliography{sn-bibliography}

\newpage

\begin{appendices}
\appendixsectionformat   
\appendixsubsectionformat
\appendixfigureformat    
\appendixtableformat
\appendixequationformat

\section{Supplementary Results}

\subsection{Assessing the generality of LLMs' concept inference capacity}
\label{si:results-generality}

\begin{figure}[htp]
\centering
\includegraphics[width=.95\textwidth]{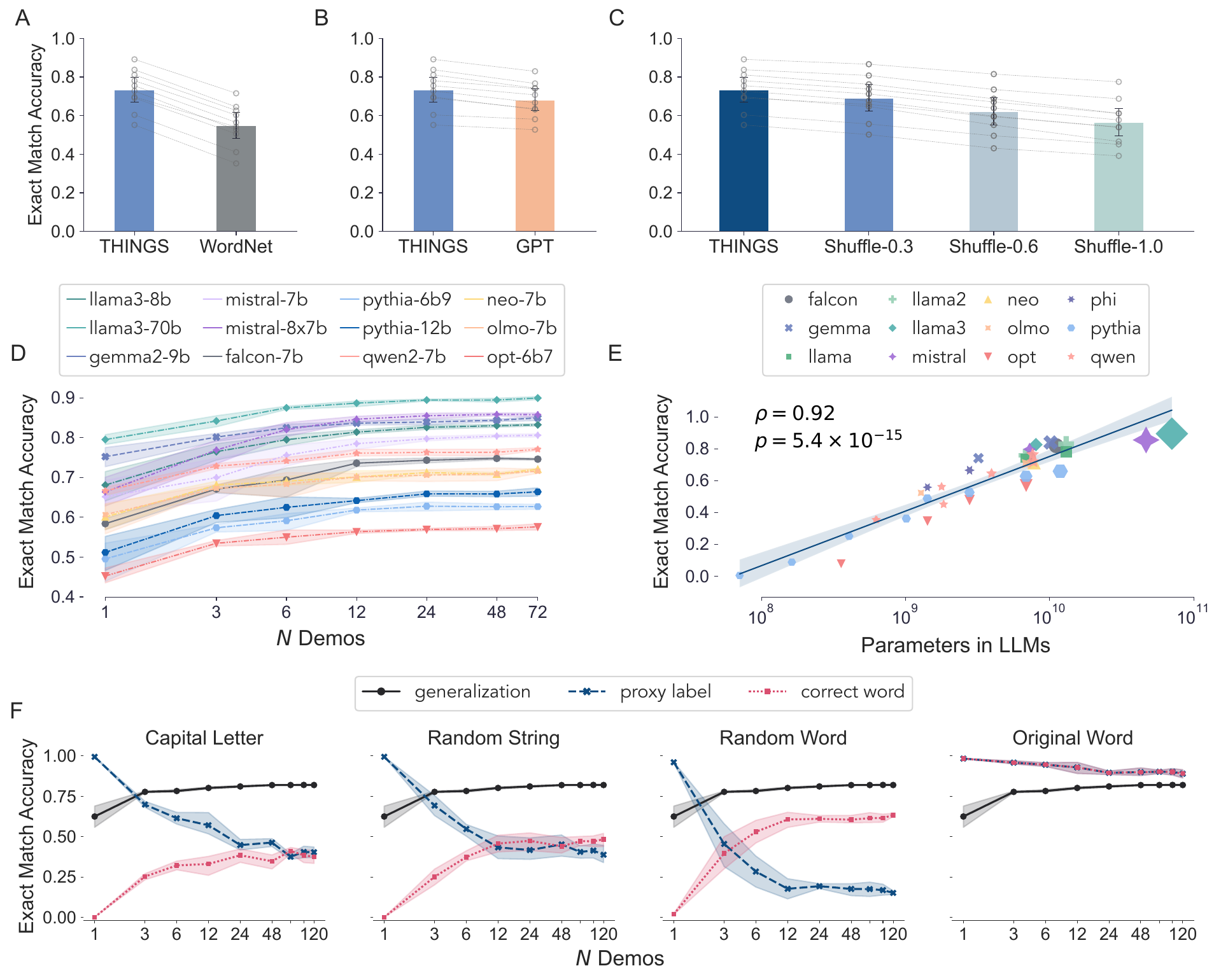}
\caption{LLMs can generalize to a broader range of concepts and descriptions. (\textit{A}--\textit{C}) Performance of LLMs on different sets of descriptions and concepts, compared to their performance on THINGS given 24 demonstrations. LLMs that excel on THINGS also adapt more effectively to diverse descriptions and concepts. (\textit{A}) Performance on concepts spanning different word classes, degrees of concreteness, and age-of-acquisition, sourced from WordNet. (\textit{B}) Performance on descriptions of the same concepts in THINGS, generated by GPT-3.5. (\textit{C}) Performance on descriptions of concepts in THINGS under increasing degree of word order permutations. Error bars represent $95\%$ confidence intervals, calculated from the average performance across five independent runs of different models. (\textit{D}) Performance of various LLMs on the reverse dictionary task, evaluated on the THINGS database and measured through exact match accuracy. The models were presented with $N$ demonstrations sampled from the training set and evaluated on an independent test set. Shaded areas denote $95\%$ confidence intervals, calculated from 10,000 resamples across five independent runs. (\textit{E}) Larger LLMs tend to perform better on concept inference. Each point represents an LLM, plotted in proportion to its scale and color-coded by model series. The x-axis denotes the number of parameters, and the y-axis shows the model's performance on the reverse dictionary task, given 24 demonstrations and averaged across five runs. (\textit{F}) LLaMA3-8B struggles more with misleading demonstrations compared to LLaMA3-70B, though it remains sensitive to relationships among concepts. Black lines denote LLaMA3-8B's performance in the standard setting with $N$ correct demonstrations. Blue and red lines show performance when one misleading demonstration---a description paired with a proxy label (capital letter, random string, or random word, as indicated in the figure titles)---is added to $N - 1$ correct demonstrations of other concepts. The model is queried with the description identical to the misleading one. The blue line shows the frequency with which the model reproduces the proxy label, while the red line indicates how often it generates the correct word for the query concept based on contextual information from other concepts. Shaded areas represent $95\%$ bootstrapped confidence intervals, calculated from 10,000 resamples over five independent runs.}\label{fig:probe-si}
\end{figure}

To evaluate the generality of our results, we extended our experiment to a broader range of concepts and descriptions. As shown in Fig.~\ref{fig:probe-si} \textit{A}--\textit{C}, the LLMs exhibited strong adaptability. The best-performing model, LLaMA3-70B, achieved an exact match accuracy of $83.00\%$ ($\pm 0.67\%$) on GPT-generated English descriptions, compared to $89.45\%$ ($\pm 0.30\%$) on the original descriptions in THINGS~\citep{hebart_things_2019}, consistently producing proper terms for the same concepts. Performance was slightly lower on the 21,402 concepts in WordNet, with LLaMA3-70B achieving $71.59\%$ ($\pm 0.30\%$) after 24 demonstrations. These findings suggest that concept inference might be more challenging for abstract or complex concepts beyond the concrete object concepts in THINGS~\citep{hebart_things_2019}, such as ``have,'' ``make,'' and ``take,'' which can be harder to describe precisely. Additionally, we observed a modest decline in performance when varying degrees of word order permutations were applied to the query descriptions. Specifically, the exact match accuracy of LLaMA3-70B dropped to $77.52\%$ ($\pm 0.22\%$) under full permutation. This indicates that while LLMs maintain some robustness to input noise, they are at least sensitive to linguistic structures when combining words into coherent conceptual representations.

Results from additional open-source LLMs further validated the generalizability of our findings. For the THINGS database, the exact match accuracy of LLMs we tested improved progressively as the number of demonstrations increased from 1 to 24 (Fig.~\ref{fig:probe-si}\textit{D}). This trend aligns with the observations in \textit{Results} presented in the main paper, though there was notable variability among models. In general, LLMs with a larger number of parameters demonstrated better performance ($\rho = 0.915$, $P < 0.0001$, $95\%$ CI: $0.806$--$0.963$, Fig.~\ref{fig:probe-si}\textit{E}).

We extended the counterfactual analysis described in the main paper to another LLM, LLaMA3-8B, to probe the interrelatedness among concepts within this model. Similar to LLaMA3-70B, this model gradually shifted from replicating the proxy symbol in context to generating the correct word for the query concept (Fig.~\ref{fig:probe-si}\textit{F}). However, it struggled with misleading demonstrations, exhibiting varying performance across different types of proxy symbols and failing to recover its original performance in the standard generalization setting. Specifically, it performed particularly poorly with symbols that lacked semantic content, such as random capital letters or strings of random characters. This suggests that the capacity to leverage contextual cues from other concepts for inference is emergent, and highlights a critical difference among LLMs in capturing interrelationships among concepts.

\subsection{Uncovering a context-independent structure through alternative metrics}
\label{si:results-converge}

\begin{figure}[htp]
\centering
\includegraphics[width=\textwidth]{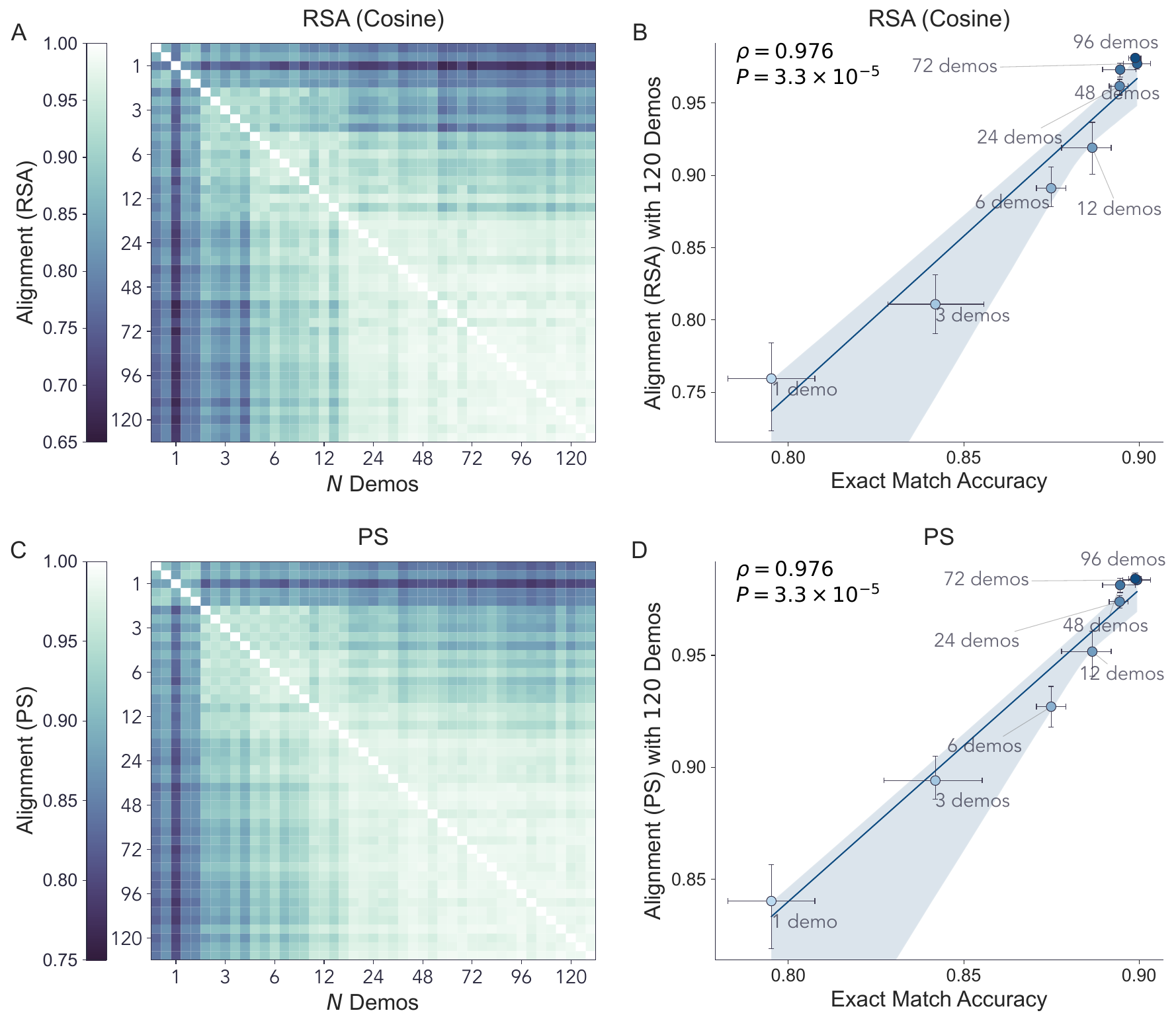}
\caption{LLMs converge toward a context-independent representational structure of concepts. (\textit{A}, \textit{C}) Pairwise alignment of LLaMA3-70B conceptual representations across different contextual demonstrations, measured through cosine-similarity-based RSA (\textit{A}) and parallelism score (PS, \textit{C}), respectively. Axes indicate the number of demonstrations, and each cell shows the alignment from a single run. (\textit{B}, \textit{D}) LLM performance on the reverse dictionary task reflects alignment with the representations formed from 120 demonstrations, measured through cosine-similarity-based RSA (\textit{B}) and PS (\textit{D}), respectively. Each point corresponds to the representations formed by the LLM based on $N$ demonstrations, with the x-axis showing performance and the y-axis indicating alignment with the 120-demonstration space. Error bars denote $95\%$ confidence intervals, calculated from 10,000 bootstrap resamples across five independent runs.}\label{fig:converge-si}
\end{figure}

Our findings in the main paper suggest that LLMs' conceptual representations are converging toward a shared, context-independent conceptual structure. This was demonstrated using representational similarity analysis (RSA)~\citep{kriegeskorte_representational_2008}, with Spearman's rank correlation coefficient as the similarity measure. We further validated this result using two additional metrics. The first, also based on RSA, used cosine similarity as the similarity measure, while the second employed the average parallelism score (PS)~\citep{bernardi_geometry_2020}, which quantifies whether the direction of vector offsets between concepts is maintained across contexts (\textit{SI Appendix}, section~\ref{si:methods-ps-analysis}). Both metrics revealed a consistent trend, where the alignment between representation spaces gradually increased as the number of contextual demonstrations rose from 1 to 24, with marginal gains beyond this threshold (Fig.~\ref{fig:converge-si}\textit{A},\textit{C}). The alignment with the space formed based on 120 demonstrations increased from $0.759$ ($\pm 0.036$) for cosine-similarity-based RSA and $0.840$ ($\pm 0.022$) for PS with one demonstration to $0.962$ ($\pm 0.006$) and $0.974$ ($\pm 0.003$) after 24 demonstrations. The alignment was strongly correlated with the LLM's exact match accuracy on concept inference ($\rho = 0.976$, $P < 0.0001$ $95\%$ CI: $0.778$--$1.000$ for cosine-similarity-based RSA and $\rho = 0.976$, $P < 0.0001$ $95\%$ CI: $0.795$--$1.000$ for PS; Fig.~\ref{fig:converge-si}\textit{B},\textit{D}). These results add to the evidence that LLMs are able to construct a context-independent relational structure, which is reflected in their concept inference capacity. They also highlight that various relationships among conceptual representations are preserved across contexts, readily supporting knowledge generalization.

\subsection{Determining appropriate similarity functions for representation spaces}
\label{si:results-similarity}

\begin{figure}[htp]
\centering
\includegraphics[width=\textwidth]{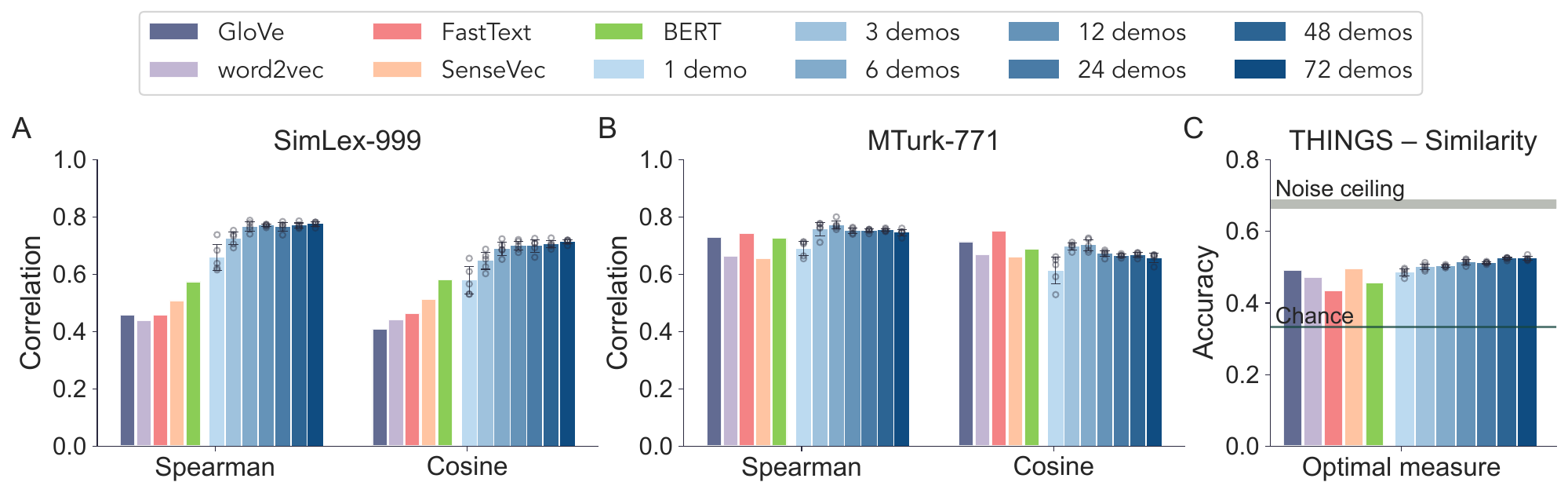}
\caption{Performance of LLaMA3-70B conceptual representations in predicting human similarity judgments, compared with baseline embeddings including static word embeddings (GloVe, word2vec, FastText), sense vectors (DeConf) and BERT embeddings obtained by averaging across all occurrences of each word in the corpus. (\textit{A}--\textit{B}) Alignment with human similarity ratings using two similarity measures: Spearman's rank correlation and cosine similarity. (\textit{A}) Spearman's correlation with human similarity ratings from SimLex-999, which targets genuine semantic similarity. (\textit{B}) Spearman's correlation with human similarity ratings from MTurk-771, which measures relatedness. LLM-derived conceptual representations significantly outperform baseline embeddings on semantic similarity, with baseline embeddings primarily reflecting relatedness. (\textit{C}) Accuracy in predicting human triplet odd-one-out judgments from the THINGS dataset, using the optimal similarity measure for each embedding. The noise ceiling indicates the upper bound of performance based on inter-subject consistency. Error bars represent $95\%$ confidence intervals computed from five independent runs.}\label{fig:similarity-si}
\end{figure}

To characterize the relational structure of LLM-derived conceptual representations, a proper similarity function is needed~\citep{roads_modeling_2024}. We employed two similarity measures, including cosine similarity and Spearman's rank correlation, and assessed whether each reflected meaningful relationships between concepts. The resulting similarity scores were compared to with human similarity ratings from two datasets: SimLex-999~\citep{hill_simlex_2015} and MTurk-771~\citep{halawi_large_2012}. While both datasets provide ratings for concept pairs, SimLex-999 explicitly distinguishing semantic similarity from association or relatedness, whereas MTurk-771 focuses primarily on relatedness. Our results (Fig.~\ref{fig:similarity-si} \textit{A}--\textit{B}) show that cosine similarity captures genuine semantic similarity instead of relatedness, yielding a correlation of $\rho = 0.705$ ($\pm 0.013$) on SimLex-999 for representations derived from 48 demonstrations, while the performance on MTurk-771 was $\rho = 0.668$ ($\pm 0.008$). In contrast, Spearman's rank correlation aligns well with both semantic similarity and relatedness, achieving $\rho = 0.771$ ($\pm 0.009$) on SimLex-999 and $\rho = 0.755$ ($\pm 0.005$) on MTurk-771. We thus adopted Spearman's rank correlation as our primary similarity function.

We conducted the same set of experiments on all baseline embeddings---FastText~\citep{grave-etal-2018-learning}, GloVe~\citep{pennington_glove_2014}, word2vec~\citep{mikolov_distributed_2013}, DeConf sense vectors~\citep{pilehvar_de_2016}, and BERT~\citep{devlin_bert_2019} (\textit{SI Appendix}, section~\ref{si:methods-baseline})---to identify the appropriate similarity function for each. Overall, the baseline embeddings yielded high correlations on MTurk-771, with FastText performing the best ($\rho = 0.753$ for cosine similarity and $\rho = 0.745$ for Spearman's rank correlation). However, these embeddings struggled on SimLex-999, where BERT achieved the highest correlation ($\rho = 0.581$ for cosine similarity and $\rho = 0.575$ for Spearman's rank correlation) (Fig.~\ref{fig:similarity-si} \textit{A}--\textit{B}). These results indicate that the baseline embeddings primarily reflect relatedness but fail to capture genuine similarity, in line with prior findings~\citep{hill_simlex_2015}. This limitation also underscores a key distinction between the contextually formed conceptual representations of LLMs and embeddings corresponding to individual word forms. Notably, unlike LLMs, the two similarity measures produced comparable results across both datasets. For subsequent analyses, we adopted the similarity function that yielded the best performance for each embedding.

\subsection{Characterizing prediction errors in human behavioral judgments}
\label{si:results-behavior-error}

Given that the performance of LLM-derived conceptual representations in predicting human similarity judgments for THINGS concepts remained below the noise ceiling estimated from inter-subject agreement ($52.41\%$ ($\pm 0.31\%$) after 48 demonstrations, compared to $67.67\% \pm 1.08\%$; Fig.~\ref{fig:similarity}\textit{C}), we conducted an error analysis to examine when these representations diverge from human judgments. To characterize the error profile, we leveraged a human similarity embedding~\citep{hebart_things-data_2023} that comprises interpretable dimensions and has been shown to account for human similarity judgments of the THINGS concepts. We quantified the relevance of each embedding dimension in determining human judgments~\citep{mahner_dimensions_2025}, and then computed the failure rate of the LLM-derived representations for trials dominated by each dimension (\textit{SI Appendix}, section~\ref{si:methods-dim-relevance}). The analysis (Fig.~\ref{fig:rep-ana-limit}\textit{A}) revealed that failures were most frequent when human judgments relied primarily on visual and perceptual features---specifically color (with ``yellow,'' ``green,'' ``black,'' ``red,'' ``orange,'' ``sand-colored,'' and ``white'' corresponding to the seven dimensions with the highest failure rates), followed by texture (e.g., ``coarse pattern/many things'' and ``oriented/many things'') and shape (e.g., ``circular/round'' and ``tubular''). These results suggest a divergence between LLM-derived representations and human similarity judgments in these visually grounded dimensions.

\subsection{Evaluating prototype- and exemplar-based approaches to categorization}
\label{si:results-categorization}

\begin{figure}[htp]
\centering
\includegraphics[width=\textwidth]{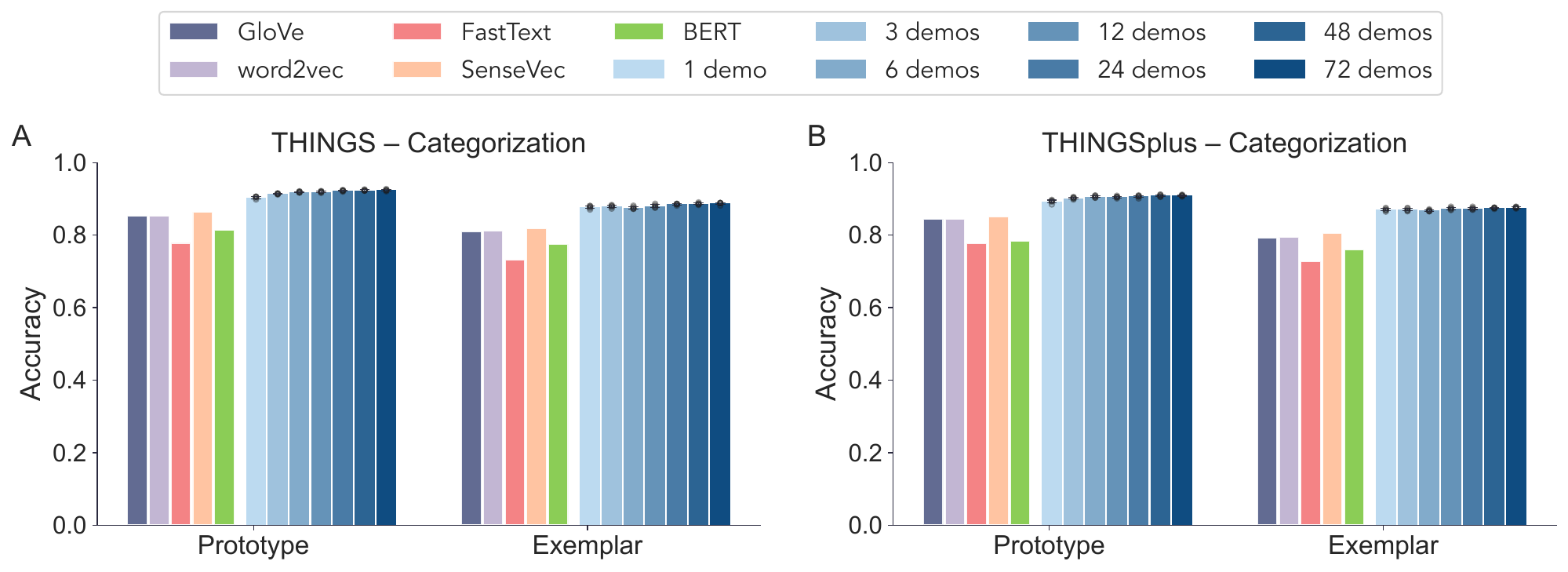}
\caption{LLM-derived conceptual representations support similarity-based categorization under both prototype- and exemplar-based strategies. (\textit{A}--\textit{B}) Performance of LLaMA3-70B conceptual representations compared with baseline embeddings, including static word embeddings (GloVe, word2vec, FastText), sense vectors (DeConf) and BERT embeddings obtained by averaging across all occurrences of each word in the corpus.  (\textit{A}) Accuracy on the original THINGS category set. (\textit{B}) Accuracy on the extended THINGSplus category set. Across both datasets and categorization strategies, LLM-derived representations consistently outperformed all baseline embeddings. Error bars represent $95\%$ confidence intervals computed from five independent runs.}\label{fig:category}
\end{figure}

To investigate whether LLM-derived conceptual representations support similarity-based categorization, we implemented two strategies grounded in the prototype and exemplar views of concepts. As shown in Fig.~\ref{fig:category}, these representations generally enabled accurate categorization. The prototype-based approach achieved the highest performance, reaching $90.87\%$ ($\pm 0.25\%$) with 48 demonstrations on the THINGSplus category set~\citep{stoinski_thingsplus_2024} and $92.37\%$ ($\pm 0.13\%$) on the original THINGS set~\citep{hebart_things_2019}. The exemplar-based strategy yielded slightly lower, but still robust, performance ($87.46\% \pm 0.14\%$ on THINGSplus; $88.63\% \pm 0.23\%$ on THINGS). In both cases, LLM-derived conceptual representations significantly outperformed all baseline embeddings, which underscores their capacity to form coherent structures that align closely with human knowledge.

\subsection{Recovering context-dependent knowledge from LLM-derived conceptual representations without extreme feature values}
\label{si:results-sem-proj}

\begin{figure}[htp]
\centering
\includegraphics[width=\textwidth]{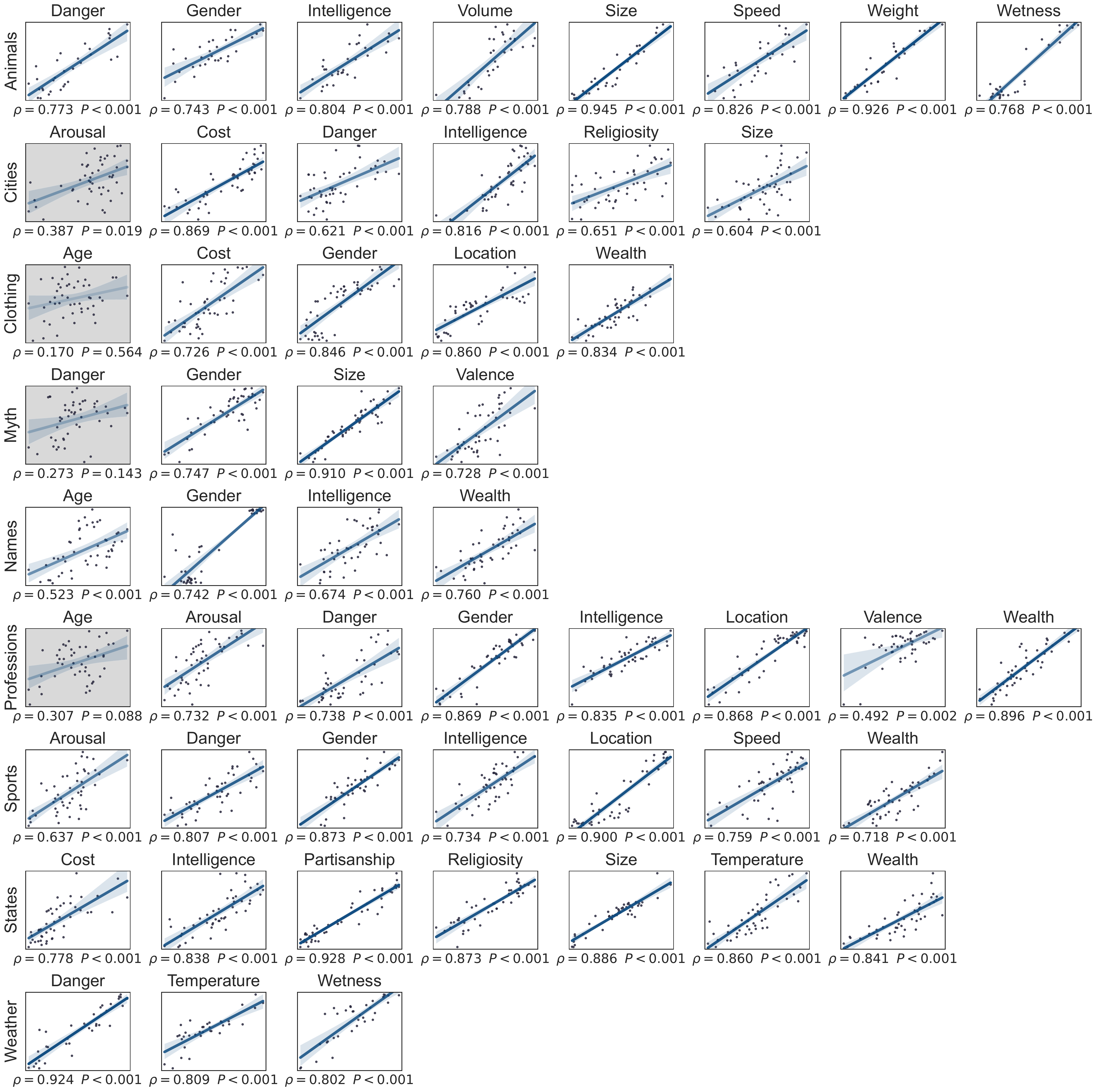}
\caption{Performance of conceptual representations derived from LLaMA3-70B in predicting context-dependent human judgments across 52 category--feature pairs, with the most extreme values excluded. Scatter plots illustrate the relationship between predicted ratings from conceptual representations (x-axis) and the average human ratings (y-axis). Linear fits are shown as straight lines, with shaded regions representing $95\%$ confidence intervals derived from 10,000 bootstrap resamples. Category--feature pairs with statistically significant correlations (Spearman's rank correlation, FDR $P < 0.01$) are displayed against a white background.}\label{fig:sem-proj-rep-rme}
\end{figure}

\begin{figure}[htp]
\centering
\includegraphics[width=\textwidth]{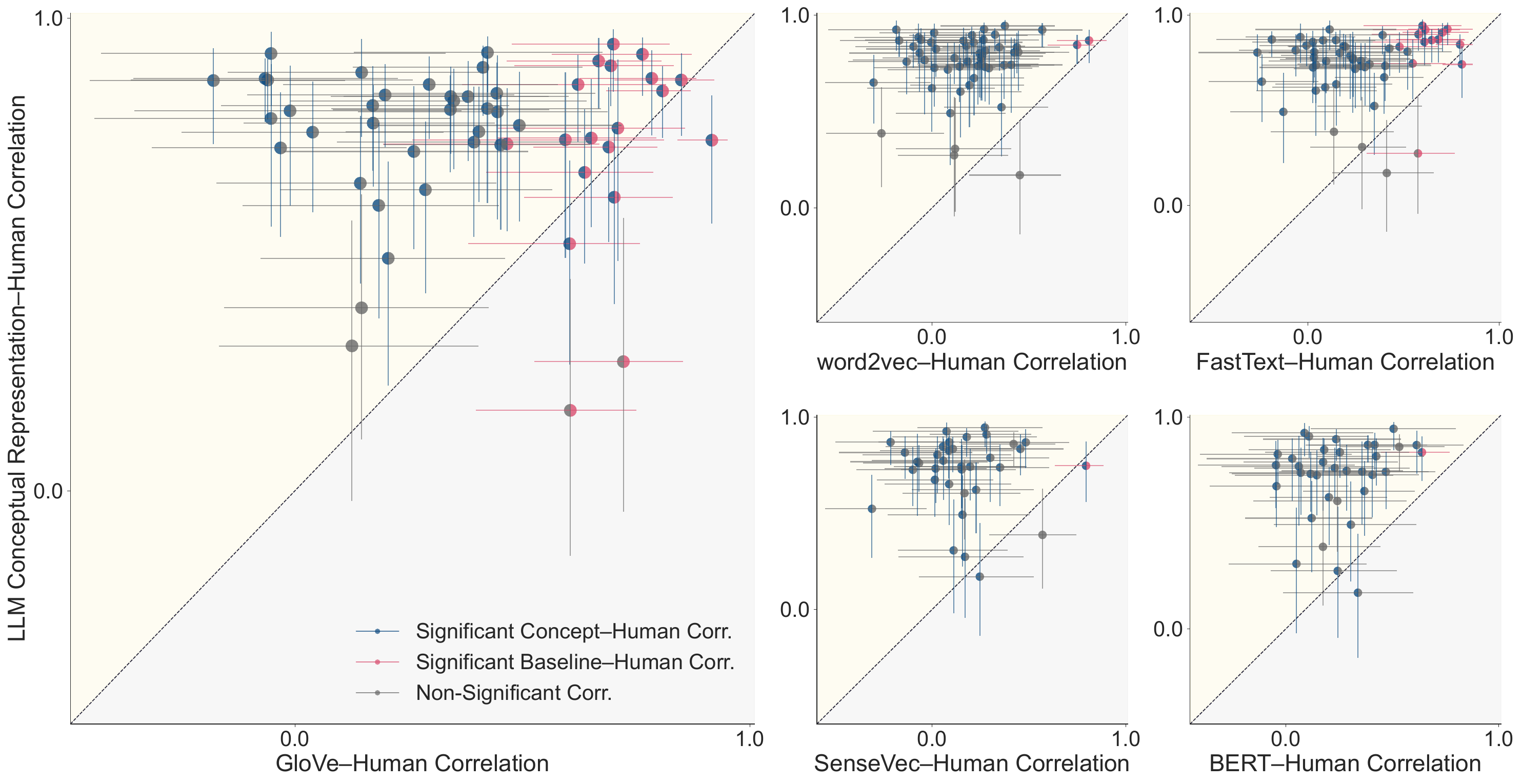}
\caption{Comparison of LLM-derived conceptual representations and baseline embeddings in predicting context-dependent human ratings, with the most extreme values excluded. The x-axis shows correlations with human ratings based on baseline embeddings, and the y-axis shows correlations based on conceptual representations derived from LLaMA3-70B. Colored points indicate significant correlations (Spearman's rank correlation, FDR $P < 0.01$). Error bars represent $95\%$ confidence intervals estimated from 10,000 bootstrap resamples. GloVe embeddings are those used in previous work~\citep{grand_semantic_2022}.}\label{fig:sem-proj-concept-vs-word-rme}
\end{figure}

Our results in the main paper suggest that LLM-derived conceptual representations can effectively recover context-dependent human ratings across categories and features. To simulate appropriate context, we presented LLMs with two demonstrations for each category--feature pair, which showcases the extreme values of the target feature within that category. This could raise concerns that the alignment with human ratings might be driven by these outliers. Although Spearman's rank correlation is comparatively robust to the influence of extreme values, we further validated our findings by reanalyzing all category--feature pairs after removing the two items with the most extreme values used in the demonstrations. The results are presented in Fig.~\ref{fig:sem-proj-rep-rme}, demonstrating high correlations with human ratings for the majority of category--feature pairs (47 out of 52; $\rho > 0.5$, $P < 0.001$, FDR corrected). The median correlation slightly decreased from $0.817$ to $0.795$, while the split-half reliability of human ratings also reduced marginally from $0.954$ to $0.946$. Moreover, LLM-derived conceptual representations consistently outperformed the strongest baseline embedding~\citep{grand_semantic_2022} across most category--feature pairs (Fig.~\ref{fig:sem-proj-concept-vs-word-rme}). These findings reaffirm the effectiveness of LLM-derived conceptual representations in handling context-dependent computations of human knowledge.

\subsection{Characterizing the variance explained by LLM conceptual representations in brain activity}
\label{si:results-brain}

\begin{figure}[htp]
\centering
\includegraphics[width=0.72\textwidth]{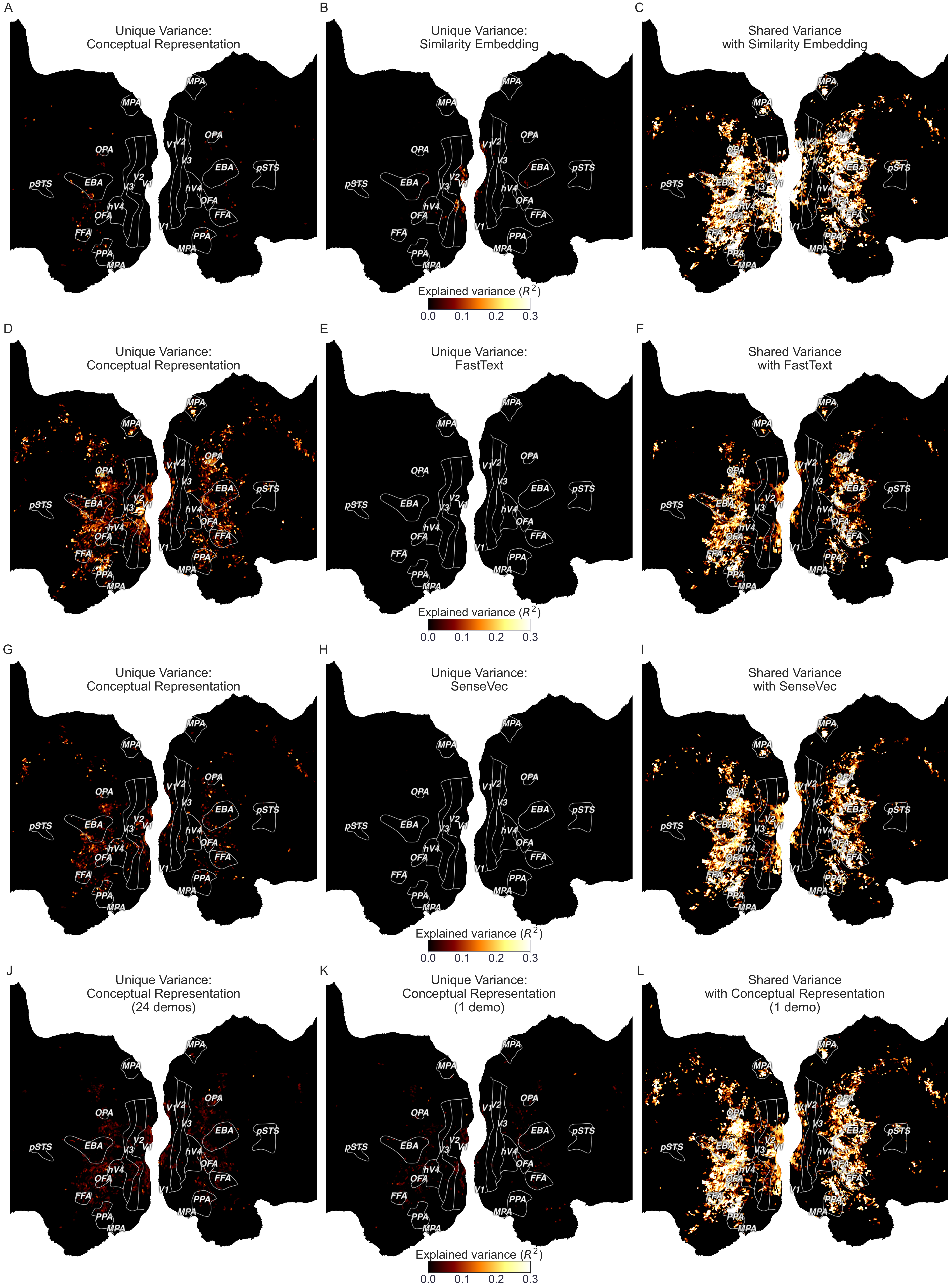}
\caption{Comparison between LLM-derived conceptual representations and baseline embeddings via variance partitioning based on full--reduced model comparisons (the primary method reported in the main text), with the dimensionality of LLM-derived representations reduced to match that of each baseline. (\textit{A}--\textit{C}) Comparison between LLM-derived conceptual representation and a similarity embedding learned from human similarity judgments. (\textit{A}) Variance uniquely explained by the LLM-derived representation. (\textit{B}) Variance uniquely explained by the similarity embedding. (\textit{C}) Shared variance explained by both models. (\textit{D}--\textit{F}) Comparison between LLM-derived conceptual representation and a static word embedding (FastText). (\textit{D}) Variance uniquely explained by the LLM-derived representation. (\textit{E}) Variance uniquely explained by FastText. (\textit{F}) Shared variance explained by both models. (\textit{G}--\textit{I}) Comparison between LLM-derived conceptual representation and DeConf sense vector. (\textit{G}) Variance uniquely explained by the LLM-derived representation. (\textit{H}) Variance uniquely explained by the sense vector. (\textit{I}) Shared variance explained by both models. (\textit{J}--\textit{L}) Comparison between LLM conceptual representation derived from 24 demonstrations and those from a single demonstration. (\textit{J}) Variance uniquely explained by the 24-demonstration representation. (\textit{K}) Variance uniquely explained by the 1-demonstration representation. (\textit{L}) Shared variance explained by both models. The colors represent the proportion of explained variance, normalized relative to the noise ceiling and averaged over five runs. Only voxels with statistically significant prediction performance are shown (FDR $P < 0.01$).}\label{fig:mri-encoding-varpart-all}
\end{figure}

\begin{figure}[htp]
\centering
\includegraphics[width=\textwidth]{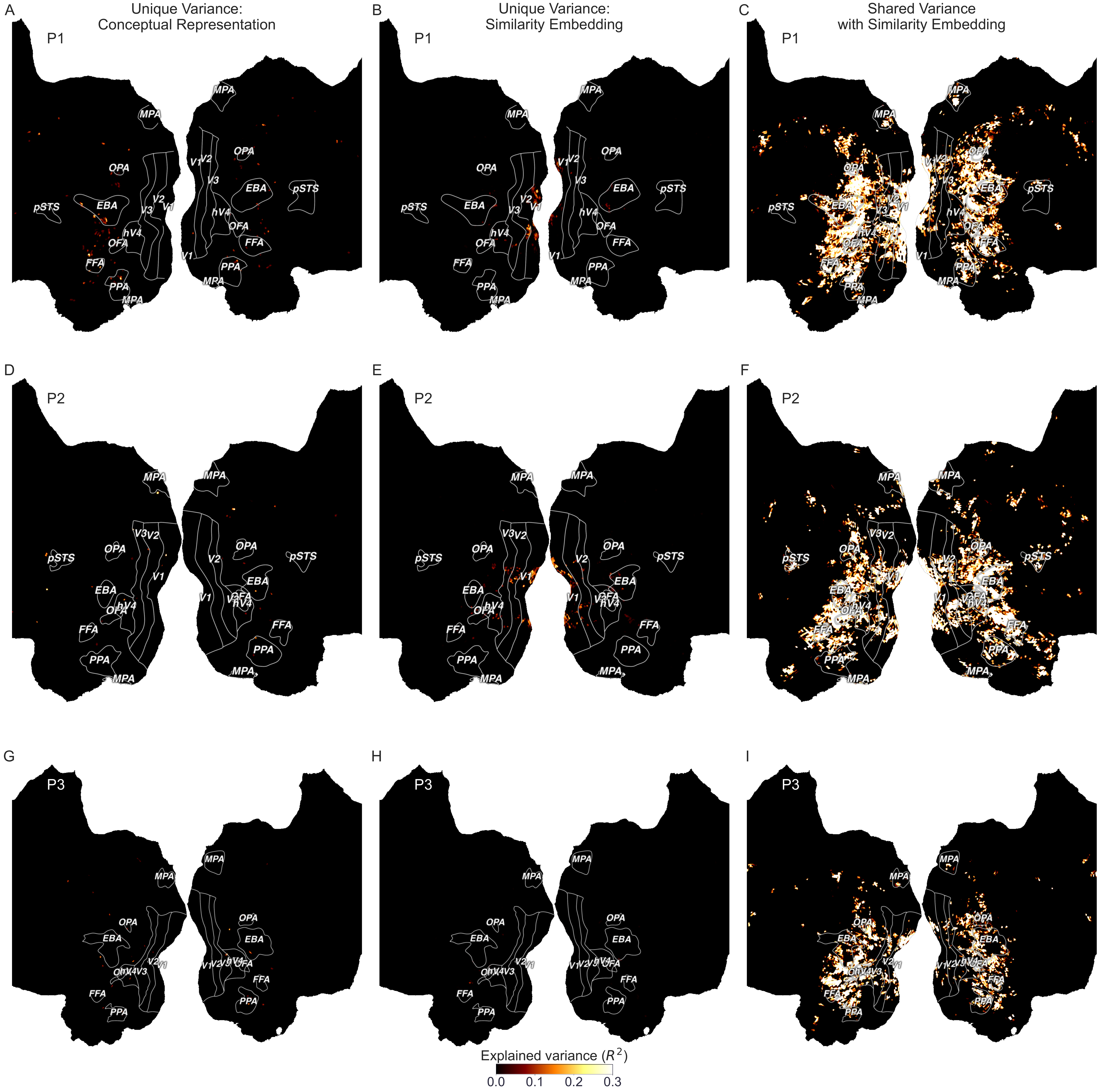}
\caption{Comparison between LLM-derived conceptual representation and a similarity embedding learned from human similarity judgments via variance partitioning based on full--reduced model comparisons (the primary method reported in the main text), with LLM-derived conceptual representations reduced to 66 dimensions to match the dimensionality of the similarity embedding. (\textit{A}) Variance uniquely explained by the LLM-derived conceptual representation. (\textit{B}) Variance uniquely explained by the similarity embedding. (\textit{C}) Shared variance explained by both models. \textit{A}--\textit{C}, \textit{D}--\textit{F}, and \textit{G}--\textit{I} show results for each of the three participants, respectively. The colors represent the proportion of explained variance, normalized relative to the noise ceiling and averaged over five runs. Only voxels with statistically significant prediction performance are shown (FDR $P < 0.01$).}\label{fig:mri-encoding-varpart-similarity}
\end{figure}

\begin{figure}[htp]
\centering
\includegraphics[width=\textwidth]{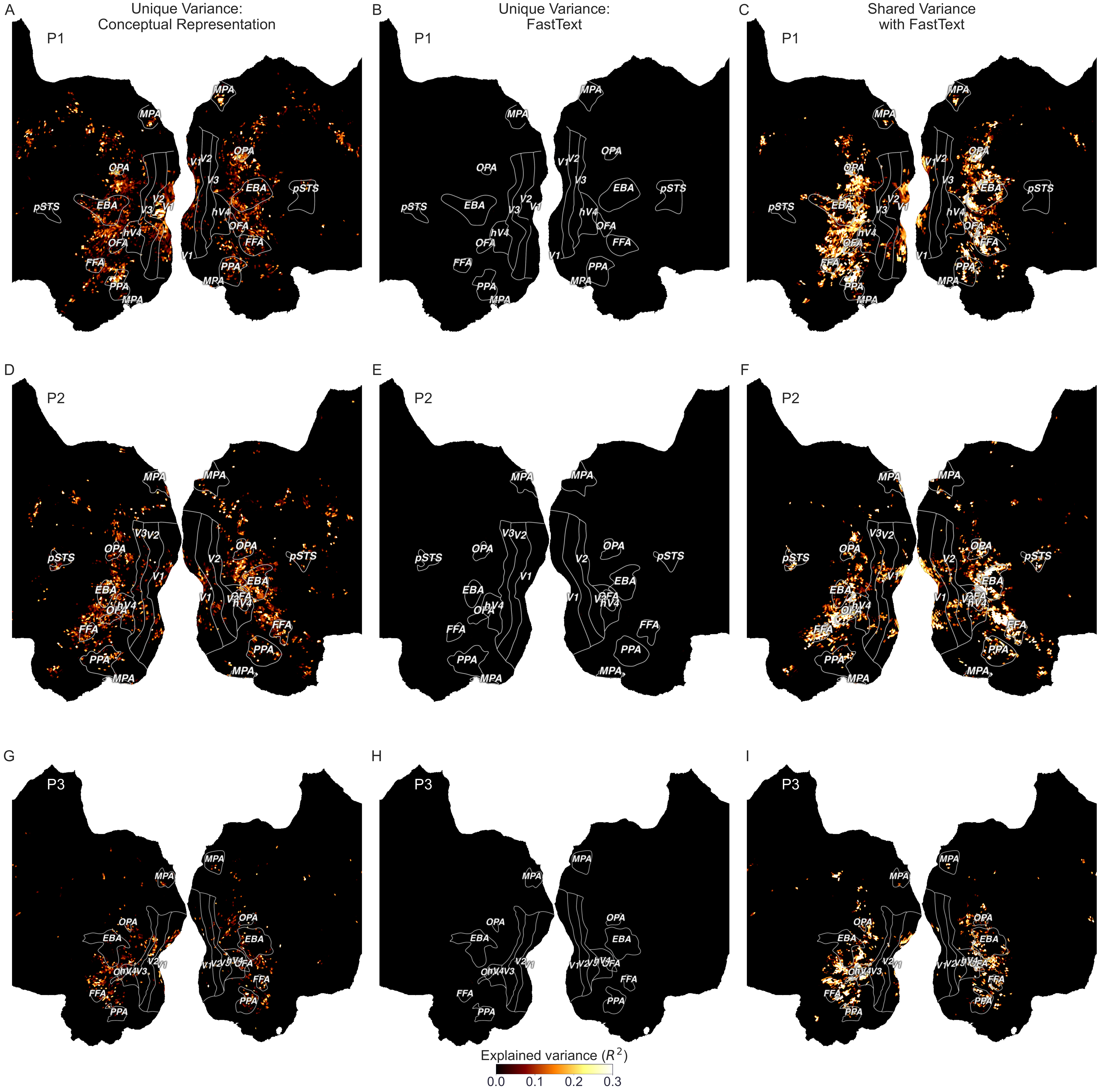}
\caption{Comparison between LLM-derived conceptual representation and a static word embedding (FastText) via variance partitioning based on full--reduced model comparisons (the primary method reported in the main text), with LLM-derived conceptual representations reduced to 300 dimensions to match the FastText dimensionality. (\textit{A}) Variance uniquely explained by the LLM-derived conceptual representation. (\textit{B}) Variance uniquely explained by FastText. (\textit{C}) Shared variance explained by both models. \textit{A}--\textit{C}, \textit{D}--\textit{F}, and \textit{G}--\textit{I} show results for each of the three participants, respectively. The colors represent the proportion of explained variance, normalized relative to the noise ceiling and averaged over five runs. Only voxels with statistically significant prediction performance are shown (FDR $P < 0.01$).}\label{fig:mri-encoding-varpart-fasttext}
\end{figure}

\begin{figure}[htp]
\centering
\includegraphics[width=\textwidth]{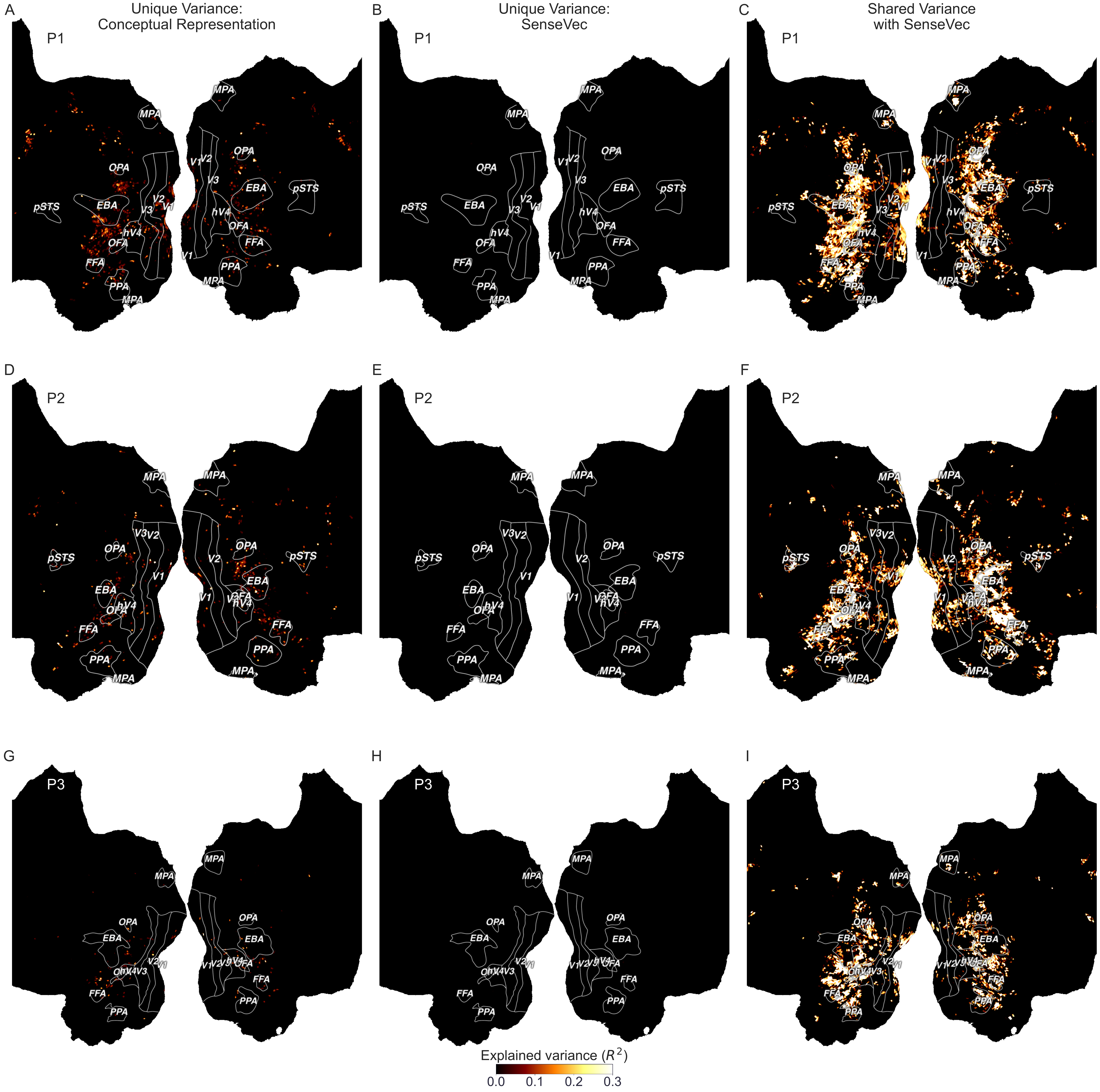}
\caption{Comparison between LLM-derived conceptual representation and DeConf sense vector via variance partitioning based on full--reduced model comparisons (the primary method reported in the main text), with LLM-derived conceptual representations reduced to 300 dimensions to match the dimensionality of the sense vector. (\textit{A}) Variance uniquely explained by the LLM-derived conceptual representation. (\textit{B}) Variance uniquely explained by the sense vector. (\textit{C}) Shared variance explained by both models. \textit{A}--\textit{C}, \textit{D}--\textit{F}, and \textit{G}--\textit{I} show results for each of the three participants, respectively. The colors represent the proportion of explained variance, normalized relative to the noise ceiling and averaged over five runs. Only voxels with statistically significant prediction performance are shown (FDR $P < 0.01$).}\label{fig:mri-encoding-varpart-sensevec}
\end{figure}

\begin{figure}[htp]
\centering
\includegraphics[width=\textwidth]{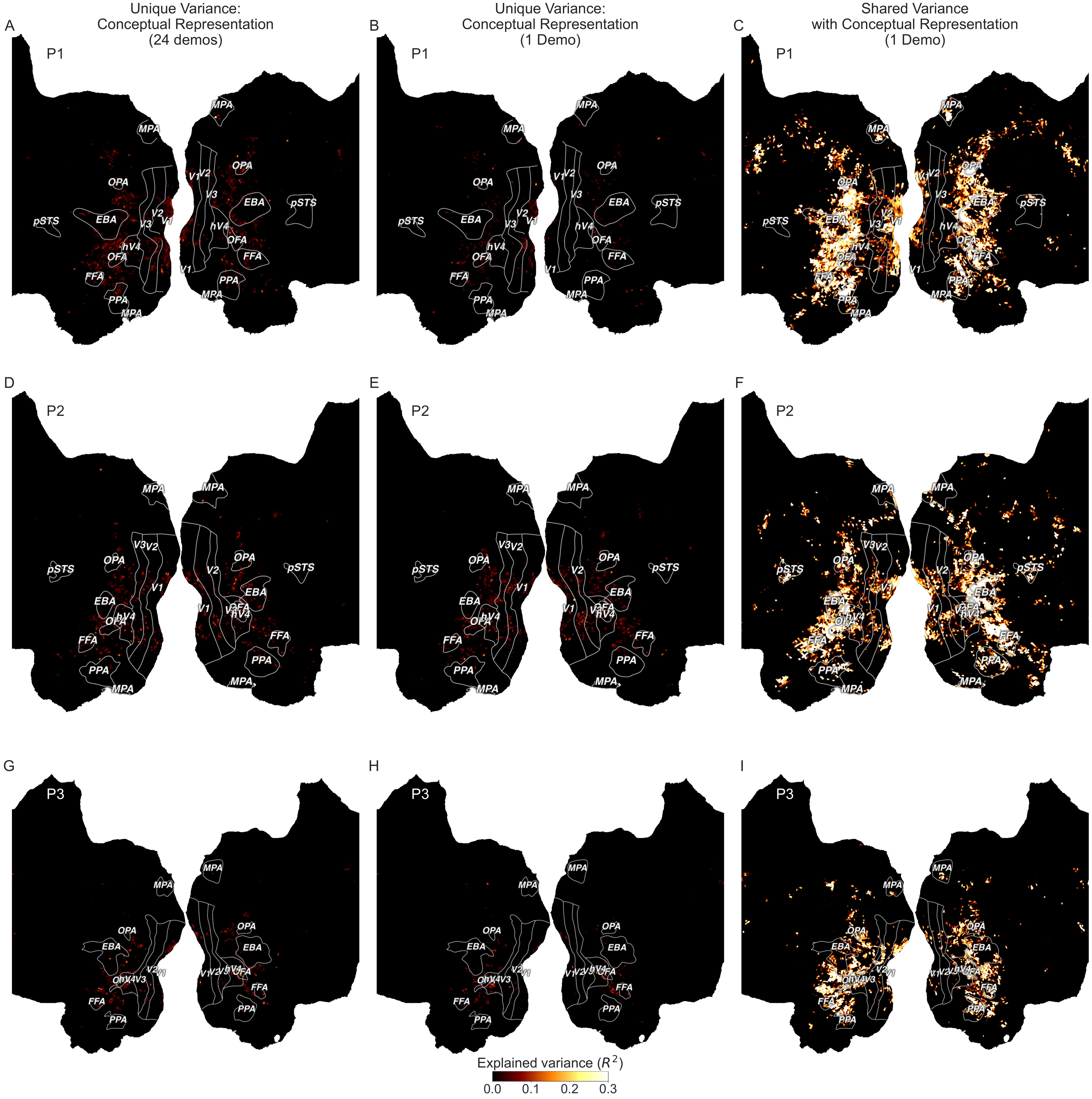}
\caption{Comparison between LLM conceptual representation derived from 24 demonstrations and those from a single demonstration via variance partitioning based on full--reduced model comparisons (the primary method reported in the main text). (\textit{A}) Variance uniquely explained by the 24-demonstration representation. (\textit{B}) Variance uniquely explained by the 1-demonstration representation. (\textit{C}) Shared variance explained by both models. \textit{A}--\textit{C}, \textit{D}--\textit{F}, and \textit{G}--\textit{I} show results for each of the three participants, respectively. The colors represent the proportion of explained variance, normalized relative to the noise ceiling and averaged over five runs. Only voxels with statistically significant prediction performance are shown (FDR $P < 0.01$).}\label{fig:mri-encoding-varpart-llm_1demo}
\end{figure}

\begin{figure}[htp]
\centering
\includegraphics[width=0.72\textwidth]{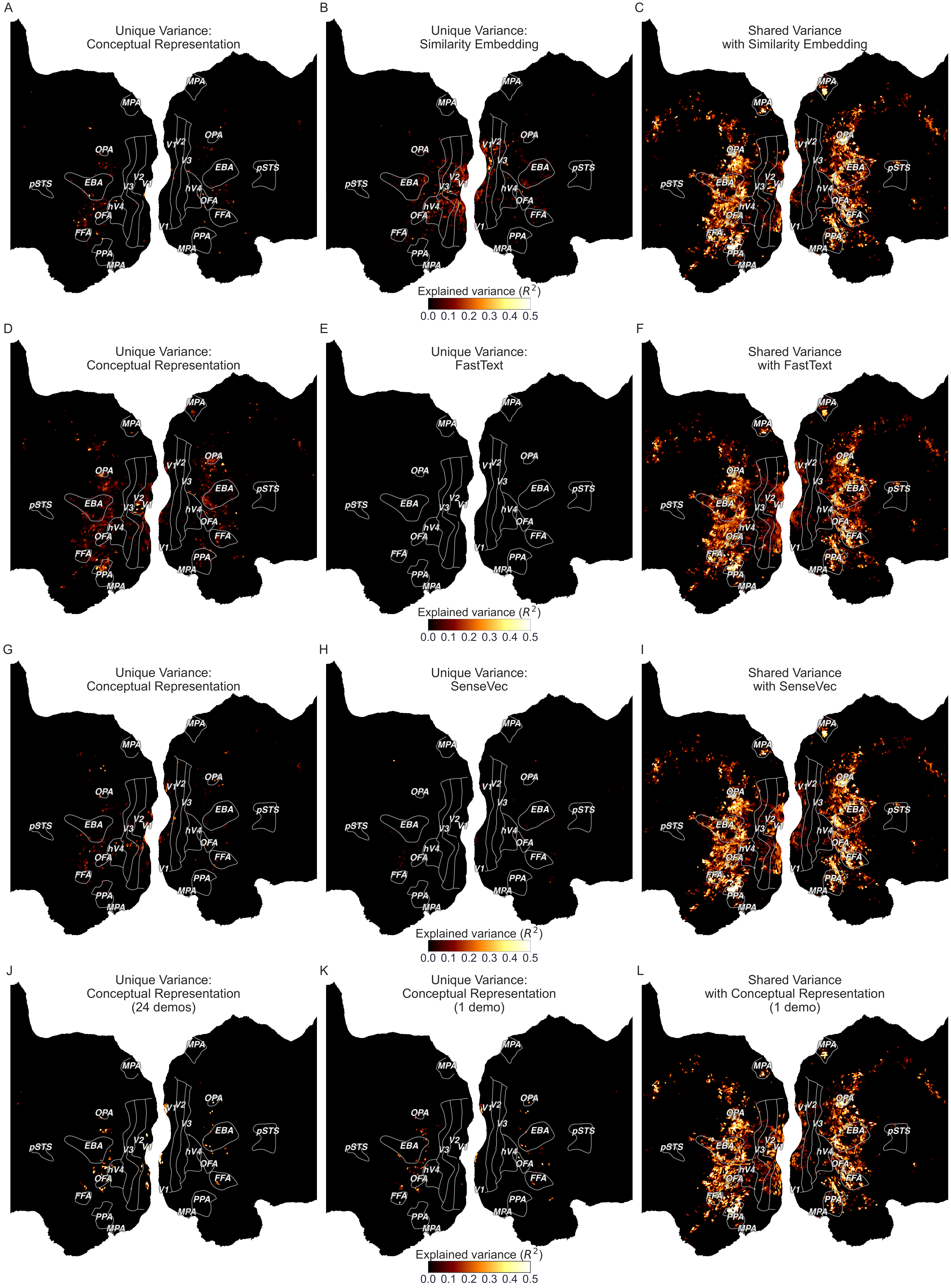}
\caption{Comparison between LLM-derived conceptual representations and baseline embeddings via orthogonalization-based variance partitioning. (\textit{A}--\textit{C}) Comparison between LLM-derived conceptual representation and a similarity embedding learned from human similarity judgments. (\textit{A}) Variance uniquely explained by the LLM-derived representation. (\textit{B}) Variance uniquely explained by the similarity embedding. (\textit{C}) Shared variance explained by both models. (\textit{D}--\textit{F}) Comparison between LLM-derived conceptual representation and a static word embedding (FastText). (\textit{D}) Variance uniquely explained by the LLM-derived representation. (\textit{E}) Variance uniquely explained by FastText. (\textit{F}) Shared variance explained by both models. (\textit{G}--\textit{I}) Comparison between LLM-derived conceptual representation and DeConf sense vector. (\textit{G}) Variance uniquely explained by the LLM-derived representation. (\textit{H}) Variance uniquely explained by the sense vector.(\textit{I}) Shared variance explained by both models. (\textit{J}--\textit{L}) Comparison between LLM conceptual representation derived from 24 demonstrations and those from a single demonstration. (\textit{J}) Variance uniquely explained by the 24-demonstration representation. (\textit{K}) Variance uniquely explained by the 1-demonstration representation. (\textit{L}) Shared variance explained by both models. The colors represent the proportion of explained variance, normalized relative to the noise ceiling and averaged over five runs. Only voxels with statistically significant prediction performance are shown (FDR $P < 0.01$).}\label{fig:mri-encoding-varpart-ortho-all}
\end{figure}

\begin{figure}[htp]
\centering
\includegraphics[width=\textwidth]{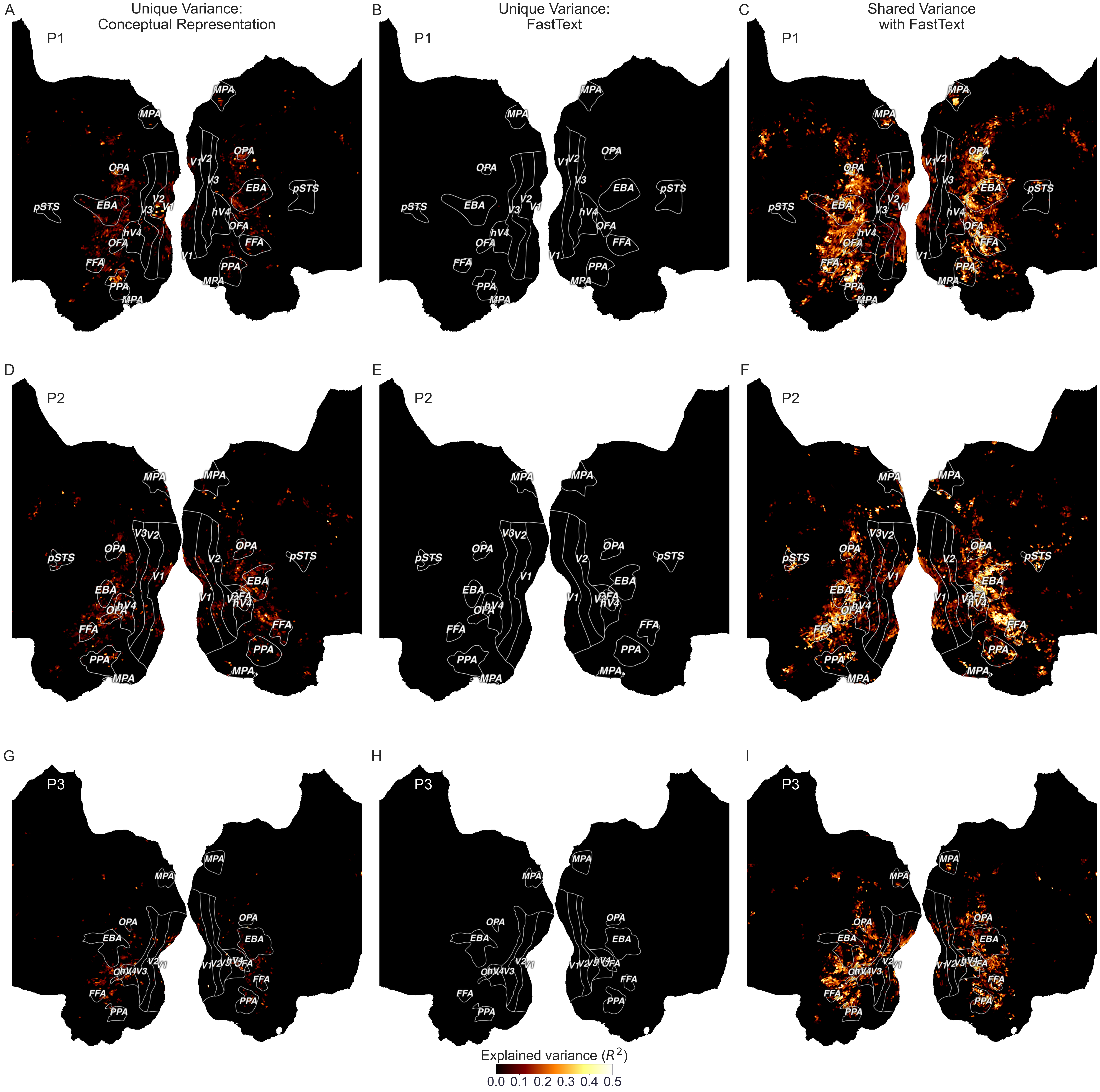}
\caption{Comparison between LLM-derived conceptual representation and a static word embedding (FastText) via orthogonalization-based variance partitioning. (\textit{A}) Variance uniquely explained by the LLM-derived conceptual representation. (\textit{B}) Variance uniquely explained by FastText. (\textit{C}) Shared variance explained by both models. \textit{A}--\textit{C}, \textit{D}--\textit{F}, and \textit{G}--\textit{I} show results for each of the three participants, respectively. The colors represent the proportion of explained variance, normalized relative to the noise ceiling and averaged over five runs. Only voxels with statistically significant prediction performance are shown (FDR $P < 0.01$).}\label{fig:mri-encoding-varpart-ortho-fasttext}
\end{figure}

\begin{figure}[htp]
\centering
\includegraphics[width=\textwidth]{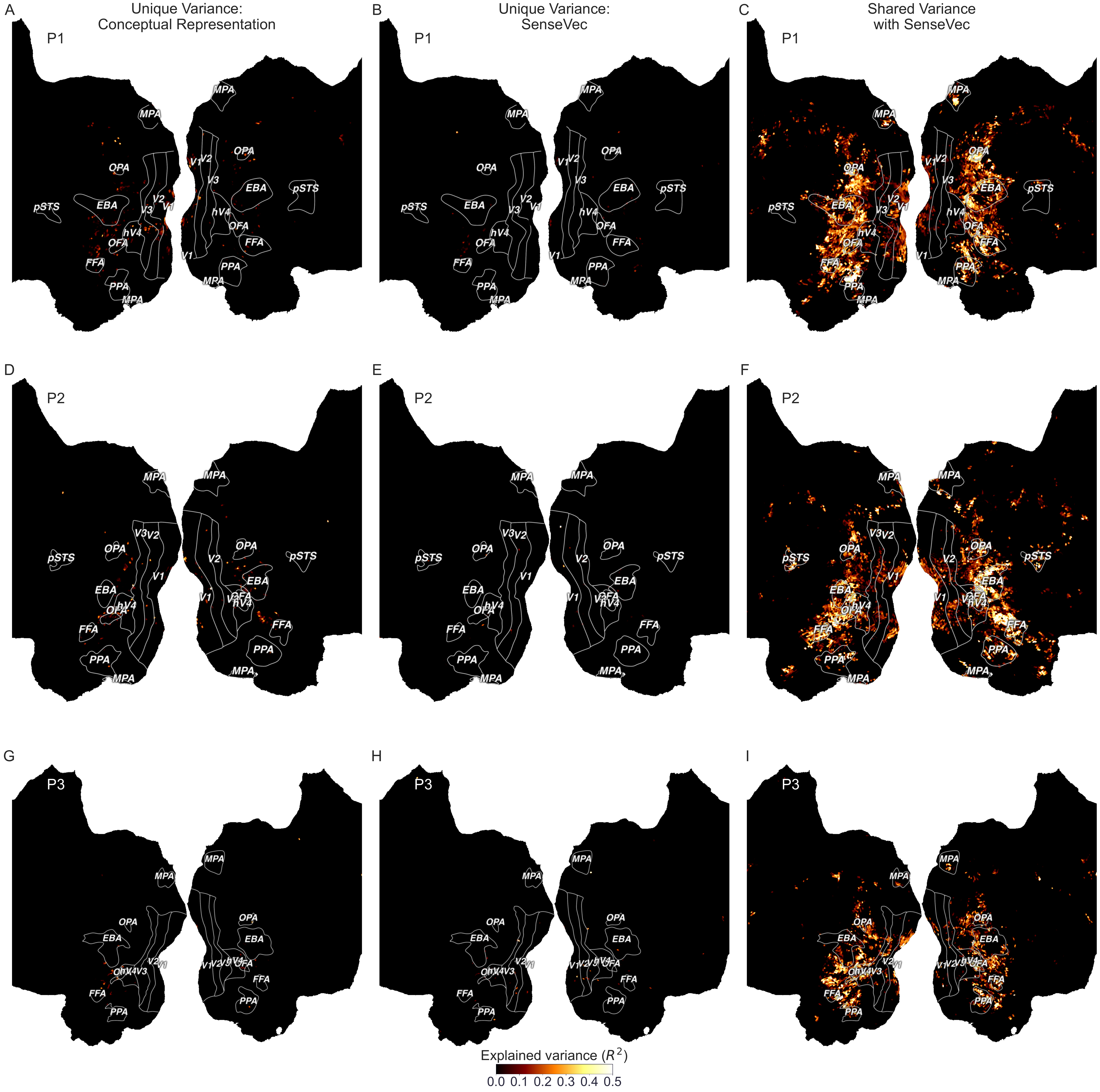}
\caption{Comparison between LLM-derived conceptual representation and DeConf sense vector via orthogonalization-based variance partitioning. (\textit{A}) Variance uniquely explained by the LLM-derived conceptual representation. (\textit{B}) Variance uniquely explained by the sense vector. (\textit{C}) Shared variance explained by both models. \textit{A}--\textit{C}, \textit{D}--\textit{F}, and \textit{G}--\textit{I} show results for each of the three participants, respectively. The colors represent the proportion of explained variance, normalized relative to the noise ceiling and averaged over five runs. Only voxels with statistically significant prediction performance are shown (FDR $P < 0.01$).}\label{fig:mri-encoding-varpart-ortho-sensevec}
\end{figure}

\begin{figure}[htp]
\centering
\includegraphics[width=\textwidth]{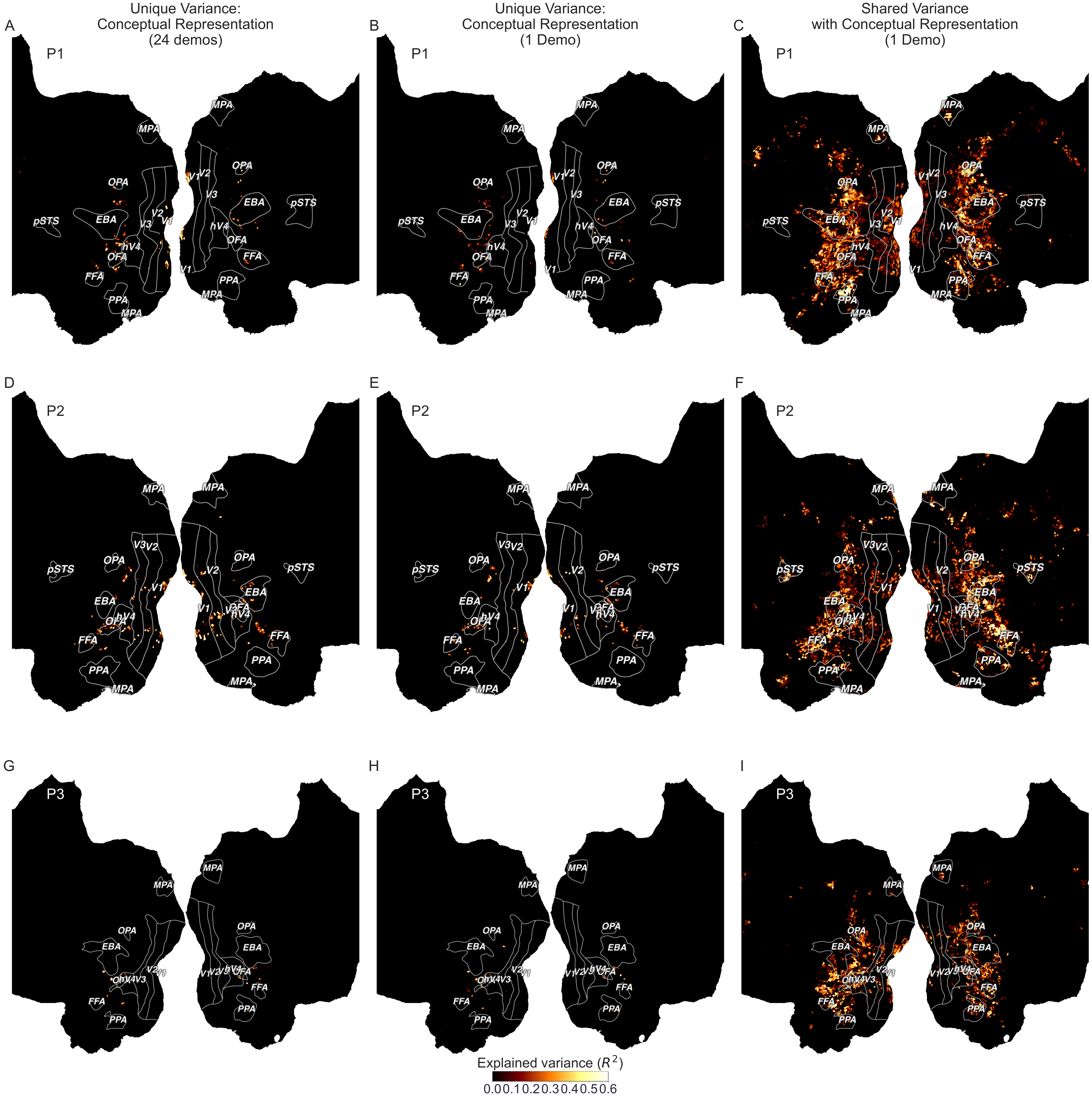}
\caption{Comparison between LLM conceptual representation derived from 24 demonstrations and those from a single demonstration via orthogonalization-based variance partitioning. (\textit{A}) Variance uniquely explained by the 24-demonstration representation. (\textit{B}) Variance uniquely explained by the 1-demonstration representation. (\textit{C}) Shared variance explained by both models. \textit{A}--\textit{C}, \textit{D}--\textit{F}, and \textit{G}--\textit{I} show results for each of the three participants, respectively. The colors represent the proportion of explained variance, normalized relative to the noise ceiling and averaged over five runs. Only voxels with statistically significant prediction performance are shown (FDR $P < 0.01$).}\label{fig:mri-encoding-varpart-ortho-llm_1demo}
\end{figure}

To test the effectiveness of LLM-derived conceptual representations in elucidating the neural coding of concepts, we compared them against three baseline representations including (1) a 66-dimensional embedding~\citep{hebart_things-data_2023} trained on and validated to successfully account for human similarity judgments for the THINGS concepts, (2) the 300-dimensional FastText word embeddings~\citep{grave-etal-2018-learning}, and (3) DeConf sense vectors~\citep{pilehvar_de_2016}. We also compared LLM-derived representations obtained with 24 demonstrations to those derived with a single demonstration, to explore potential differences in their mapping to neural activity patterns. Variance partitioning was applied to disentangle the unique and shared contributions of each representation, with the dimensionality of LLM-derived representations reduced to match each baseline for a conservative estimate of unique variance. As shown in Fig.~\ref{fig:mri-encoding-varpart-all}, DeConf sense vectors yielded qualitatively similar results to FastText but with stronger explanatory power, suggesting that integrating word sense information from WordNet enhances word embeddings. At the same time, LLM-derived representations alone accounted for greater variance across the visual system, highlighting their effectiveness in capturing neural responses. Comparisons across different numbers of demonstrations revealed that representations derived with more demonstrations explained a modestly larger portion of unique variance, though without qualitative differences in the brain regions involved. This suggests that consistently structured representations derived from LLMs (with 24 or more demonstrations) more faithfully map onto brain activity patterns than degraded ones, in line with our previous findings that such representations also align more closely with human behavioral data. A complementary variance partitioning approach based on orthogonalization (\textit{SI Appendix}, section~\ref{si:methods-varpart}) revealed similar patterns (Fig.~\ref{fig:mri-encoding-varpart-ortho-all}), further corroborating these conclusions. Consistent results across all three participants---observed in Figs.~\ref{fig:mri-encoding-varpart-similarity}--\ref{fig:mri-encoding-varpart-llm_1demo} (non-orthogonalized approach, main analysis) and Figs.~\ref{fig:mri-encoding-varpart-ortho-similarity}--\ref{fig:mri-encoding-varpart-ortho-llm_1demo} (orthogonalized approach)---underscore the robustness and generality of these findings.

\subsection{Exploring the limitations of LLM-derived conceptual representations}
\label{si:results-brain-limitations}

\begin{figure}[htp]
\centering
\includegraphics[width=0.95\textwidth]{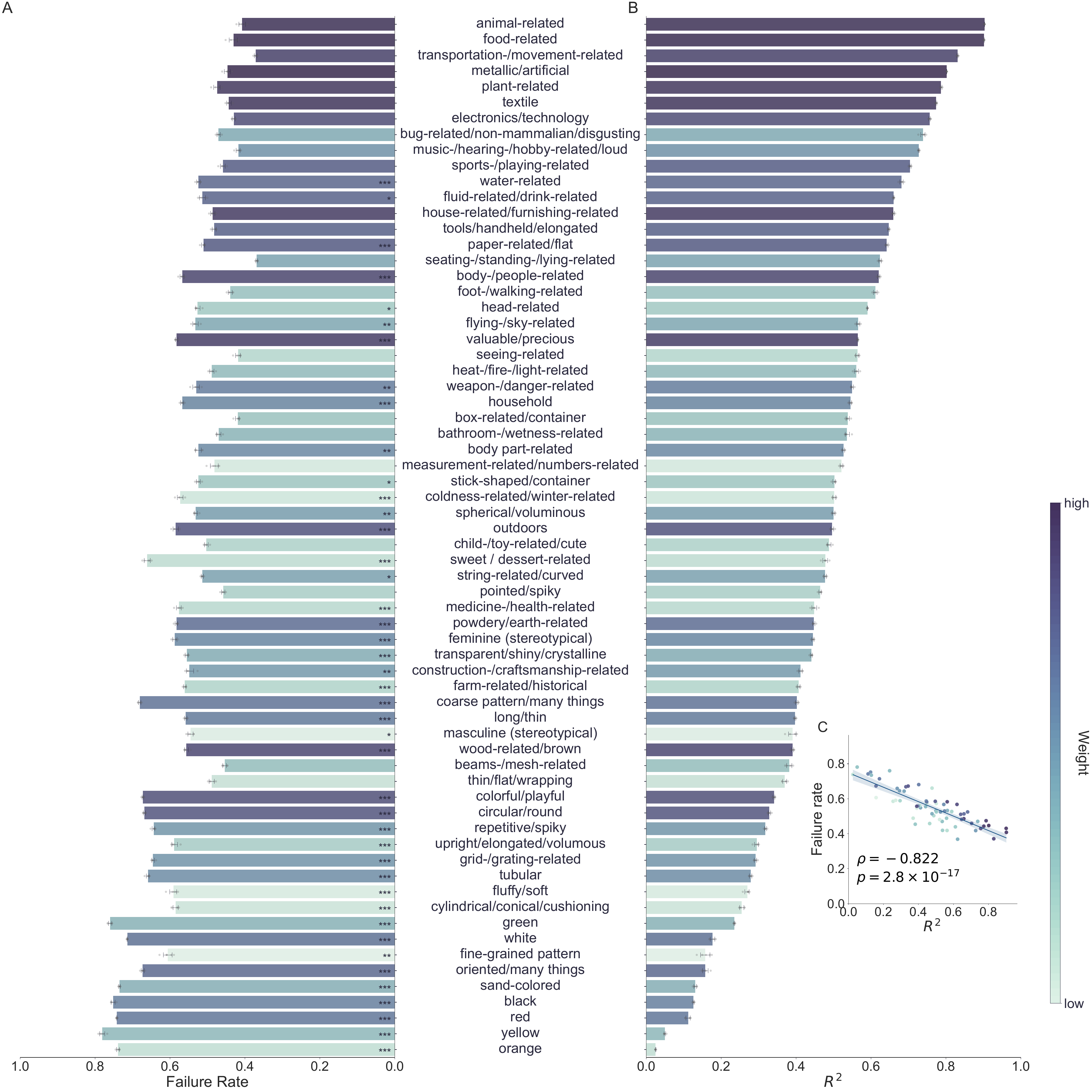}
\caption{Dimensions underlying human similarity judgments as captured by LLM-derived conceptual representations. (\textit{A}) Failure rates of LLM-derived conceptual representations in predicting human similarity judgments, computed for each dimension when it dominated odd-one-out judgments among THINGS concepts. Asterisks indicate dimensions for which the failure rate significantly exceeded the model's global error rate (one-sided exact binomial test; $P$-values corrected for multiple comparisons using the Benjamini–Hochberg false discovery rate procedure). Significance levels: ***$P < 0.001$, **$P<0.01$, *$P<0.05$. (\textit{B}) Variance explained ($R^{2}$) for each dimension of the human similarity embedding when predicted from LLM-derived conceptual representations. Each bar represents a single dimension of the similarity embedding, interpreted by human annotators and color-coded by its relative weight. Error bars denote $95\%$ confidence intervals, calculated from five independent runs. Significance markers are omitted, as all dimensions reached statistical significance after FDR correction ($P<0.0001$). (\textit{C}) Relationship between mean $R^{2}$ (x-axis) and mean failure rate (y-axis) across dimensions. Each point denotes a single dimension, color-coded by its relative weight. The solid line indicates the best linear fit, with shaded regions representing $95\%$ confidence intervals estimated from 10,000 bootstrap resamples.}\label{fig:rep-ana-limit}
\end{figure}

\begin{figure}[htp]
\centering
\includegraphics[width=\textwidth]{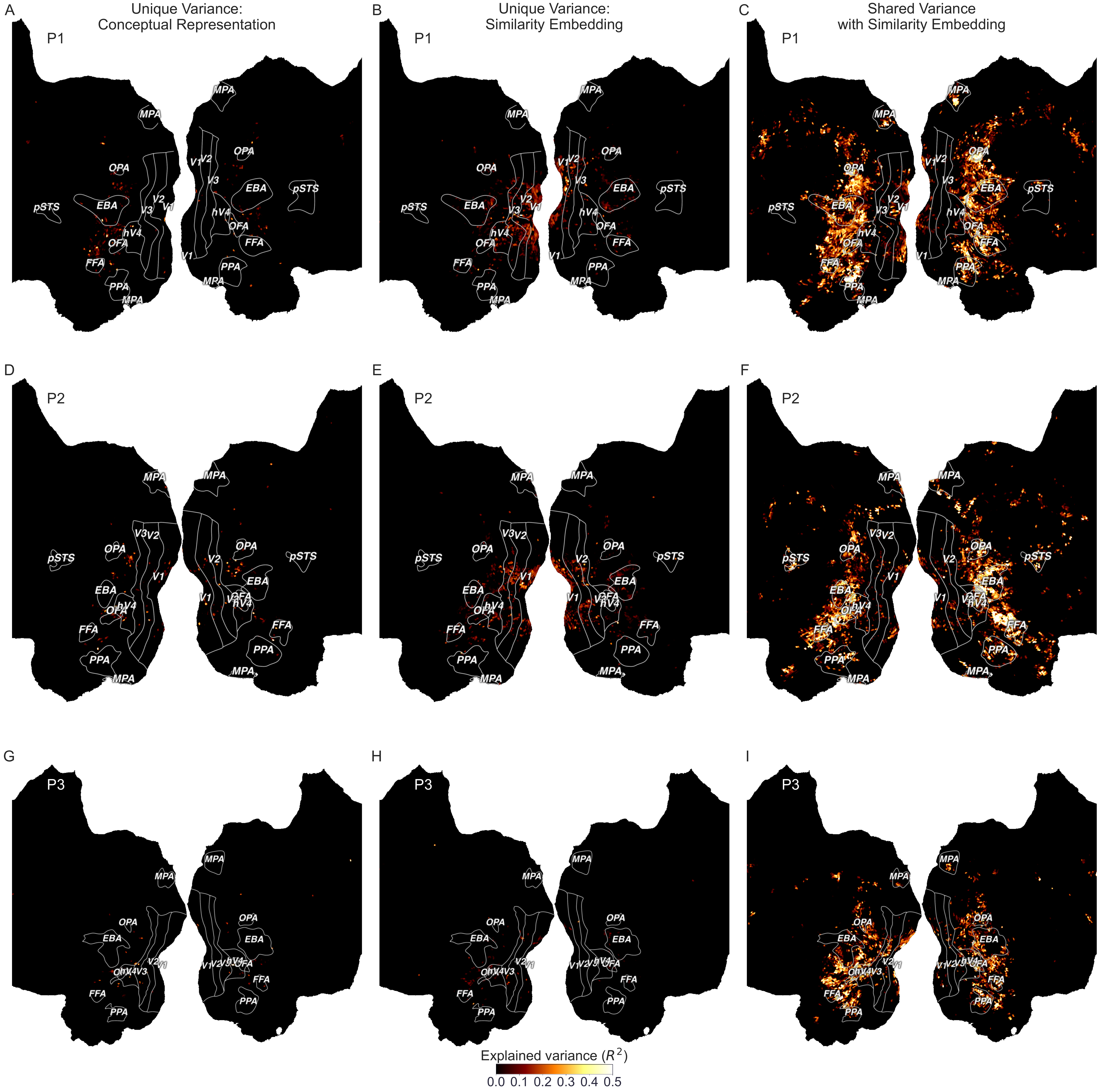}
\caption{Comparison between LLM-derived conceptual representation and a similarity embedding learned from human similarity judgments via orthogonalization-based variance partitioning. (\textit{A}) Variance uniquely explained by the LLM-derived conceptual representation. (\textit{B}) Variance uniquely explained by the similarity embedding. (\textit{C}) Shared variance explained by both models. \textit{A}--\textit{C}, \textit{D}--\textit{F}, and \textit{G}--\textit{I} show results for each of the three participants, respectively. The colors represent the proportion of explained variance, normalized relative to the noise ceiling and averaged over five runs. Only voxels with statistically significant prediction performance are shown (FDR $P < 0.01$).}\label{fig:mri-encoding-varpart-ortho-similarity}
\end{figure}

To explore the components of visual information in brain activity patterns not captured by LLM-derived conceptual representations, we regressed the human-derived similarity embedding onto these representations (\textit{SI Appendix}, section~\ref{si:methods-regress-spose}). The dimensions of the similarity embedding were sparse, non-negative, and ordered by their weights in representing object concepts. As shown in Fig.~\ref{fig:rep-ana-limit}\textit{B}, dimensions with higher weights were generally better captured by LLM-derived representations, particularly those reflecting higher-level properties such as taxonomic membership (e.g., ``animal-related'' and ``food-related'') and function (e.g., ``transportation-/movement-related''). In contrast, perceptual properties such as color (e.g., ``orange,'' ``yellow,'' ``red,'' and ``black''), texture (e.g. ``fine-grained pattern'') and shape (e.g., ``cylindrical/conical/cushioning'') were poorly accounted for, with color being the most inadequately represented. This gap mirrored the LLMs' error profile in predicting human behavior ($\rho = -0.822$, $P < 0.0001$; Fig.~\ref{fig:rep-ana-limit}\textit{C}) and mapped onto brain activity patterns in low-level visual areas, as revealed by variance partitioning based on orthogonalization (\textit{SI Appendix}, section~\ref{si:methods-varpart}; Fig.~\ref{fig:mri-encoding-varpart-ortho-similarity}). Together, these findings highlight a key limitation of LLM-derived conceptual representations, learned solely from language, in capturing visually grounded perceptual information. Incorporating grounded information offers a promising direction for enhancing alignment with human concepts~\citep{vong_grounded_2024, rane_concept_2024}.

\section{Supplementary Materials and Methods}

\subsection{Description generation using ChatGPT}
\label{si:methods-gpt-desc}

We used the following template to prompt GPT-3.5 (\texttt{gpt-3.5-turbo-0125}) to generate descriptions for the concepts in THINGS: ``Provide three distinct definitions of the word `[Word]' (referring to `[Description]') that vary in linguistic forms, without explicitly including the word itself. Try to be concise.'' 

\subsection{Parallelism score}
\label{si:methods-ps-analysis}

The parallelism score (PS)~\citep{bernardi_geometry_2020} was used to evaluate whether differences between conceptual representations are preserved across conditions, as such differences are thought to reflect relations in the representation space~\citep{mikolov_distributed_2013, grand_semantic_2022}. The PS for each concept pair, among the total $m$ concepts, is computed as the cosine similarity between the vector offsets from one concept to the other. We reported the average PS across all concept pairs for two representation spaces.

\subsection{Evaluation of model downstream performance}
\label{si:methods-eval-tasks}

We evaluated LLMs on seven widely used benchmark tasks targeting language understanding and reasoning, all framed in a multiple-choice format. We used the official test sets for evaluation when publicly available; otherwise we resorted to the development sets. Models were assessed in a zero-shot manner using natural language prompt templates. Tasks and prompt templates are provided in Table~\ref{tab:general-templates}. Let $w_{1:n}$ be the prompt composed of $n$ tokens, and $w_{n+1:c_{i}}$ denote the $i$-th candidate answer with $c_{i} - n$ tokens. For each candidate $c_{i} \in \mathcal{C}$, we calculated the score as the sum of log-likelihoods assigned by the model $\mathcal{M}$ to it: $$\sum_{t=n+1}^{c_{i}}\log p_{\mathcal{M}}\left(w_{t} \mid w_{<t}\right).$$ Model performance was evaluated by their accuracy in ranking the gold answer with the highest score.

\subsection{Estimation of dimension relevance for human similarity judgments}
\label{si:methods-dim-relevance}

We estimated the relevance of each embedding dimension for human odd-one-out choices using a dimension-wise jackknife resampling procedure, following previous work~\citep{mahner_dimensions_2025}. For each triplet, we first computed the predicted probability of selecting the odd item based on the human similarity embedding~\citep{hebart_things-data_2023}. We then iteratively omitted each dimension and quantified the resulting change in predicted probability. The relevance score of a given dimension was defined as the difference in predicted probability of the odd item prior to and following its omission. The failure rate of LLM-derived conceptual representations for each dimension was calculated as the proportion of trials in which the model failed to predict the human choice when that dimension had the highest relevance score across all dimensions.

\subsection{Exemplar-based categorization}
\label{si:methods-exemplar}

We implemented exemplar-based categorization using a nearest-neighbor decision rule~\citep{sorscher_neural_2022}, in which a test example $\mathbf{x}_{t}$ was compared to all other examples $\mathbf{x}_{e}$ and categorized based on their relative similarities. Formally, the distance between the test example and each exemplar was defined as
\begin{equation}
    d\left(\mathbf{x}_{t}, \mathbf{x}_{e}\right) = 1 - \textrm{similarity}\left(\mathbf{x}_{t}, \mathbf{x}_{e}\right).
\end{equation}
The support for each category $K$ was then calculated as 
\begin{equation}
    \textrm{Support}\left(K\right) = \frac{1}{\left|K\right|} \sum_{\mathbf{x}_{e} \in K} \exp \left(-\beta \cdot d\left(\mathbf{x}_{t}, \mathbf{x}_{e}\right) \right),
\end{equation}
where $\left|K\right|$ denotes the number of examples in class $K$, and $\beta$ is a hyperparameter that controls the weighting of distances. The test example was assigned to the category with the highest support. We explored $\beta$ values ranging from $10^{-6}$ to $10^{4}$, and selected the value that yielded the best performance for each model. For LLM-derived representations, $\beta = 100$ consistently produced the best results.

\subsection{Baseline embeddings for comparison}
\label{si:methods-baseline}

To evaluate the effectiveness of LLM-derived conceptual representations in capturing human behavioral judgments---and to explore their distinct advantages over widely-used word embeddings---we compared them against three types of baseline embeddings:

\begin{enumerate}
    \item Static word embeddings, including GloVe~\citep{pennington_glove_2014} (\url{https://nlp.stanford.edu/projects/glove/}), word2vec~\citep{mikolov_distributed_2013} (\url{https://code.google.com/archive/p/word2vec/}), and FastText~\citep{grave-etal-2018-learning} (\url{https://fasttext.cc/docs/en/crawl-vectors.html});
    \item Aggregated BERT embeddings~\citep{devlin_bert_2019}, obtained by averaging token-level representations of individual words across all their occurrences in large corpora~\citep{cassani2023meaning}, resulting in a single embedding per word type (\url{https://osf.io/94ena/?view_only=4e29d179b7664f27b6b7b7ba6b75edf7});
    \item Sense-based embeddings from DeConf~\citep{pilehvar_de_2016}, which combine word2vec embeddings with WordNet to produce disambiguated semantic representations (\url{https://pilehvar.github.io/deconf/}).
\end{enumerate}
For BERT, we reported results from the best-performing layer for each task. Among these baselines, GloVe and word2vec are inherently limited to a fixed vocabulary due to their training procedures, while the average BERT embeddings used here are similarly constrained by the coverage of the underlying corpus. FastText leverages subword modeling and thus provides broader coverage across words and phrases. DeConf representations correspond to individual WordNet senses. Items were excluded from comparisons for a given baseline if the target word or phrase did not have a corresponding embedding.

\subsection{Variance partitioning with orthogonalization}
\label{si:methods-varpart}

We employed a complementary variance partitioning approach based on orthogonalization to disentangle the unique and shared variance explained by each representation~\citep{contier_distributed_2024}. To isolate unique variance, both the target representation and fMRI responses were orthogonalized with respect to the alternate representation, thereby removing the shared variance. We then used ridge regression with 20-fold cross-validation to predict the fMRI residuals from the residuals of the target representation. The resulting $R^{2}$ reflected the variance uniquely explained by the target representation. For shared variance, we concatenated both representations to predict neural responses using the same cross-validation procedure. The shared variance was subsequently estimated by subtracting the unique contributions of each representation from the total variance explained. As in the non-orthogonalized approach based on full--reduced model comparisons described in the main text, analyses were restricted to voxels with a noise ceiling above $5\%$, and results are reported as noise-ceiling-normalized $R^{2}$. No additional dimensionality reduction was applied, as the orthogonalization step itself removed shared variance between representations and retaining the residuals in their original dimensionality better preserves the information encoded in each representation.

\subsection{Regression model for similarity embedding}
\label{si:methods-regress-spose}

To analyze the information missing from LLM-derived conceptual representations, we regressed the embedding learned from human similarity judgments~\citep{hebart_things-data_2023} onto these representations through a linear ridge regression model. This procedure served as an intermediate step in variance partitioning based on orthogonalization (\textit{SI Appendix}, section~\ref{si:methods-varpart}), where the residuals were then mapped onto brain activity patterns to estimate the unique variance explained by the human-derived similarity embedding beyond the LLM representations. The human-derived similarity embedding consisted of 66 dimensions for each of the 1,854 concepts in THINGS. As in \textit{Materials and Methods}, we modeled the value of each dimension as
\begin{equation}
    y_{d,c} = \mathbf{h}_{c} \mathbf{w} + b + \epsilon,
\end{equation}
where $y_{d,c}$ is the value at dimension $d$ for concept $c$, $\mathbf{h}_c$ is the corresponding LLM representation, $\mathbf{w}$ is a vector of regression weights, $b$ is a constant, and $\epsilon$ denotes residual error. The parameters $\mathbf{w}$ and $b$ were estimated via ridge regression, with the regularization parameter selected via cross-validation within the training set. Model performance was assessed through twenty-fold cross-validation with non-overlapping concepts, evaluating the $R^{2}$ between predicted and observed values across folds.

\section{Supplementary Figures and Tables}

\clearpage

\clearpage

\begin{figure}[htp]
\centering
\includegraphics[width=\textwidth]{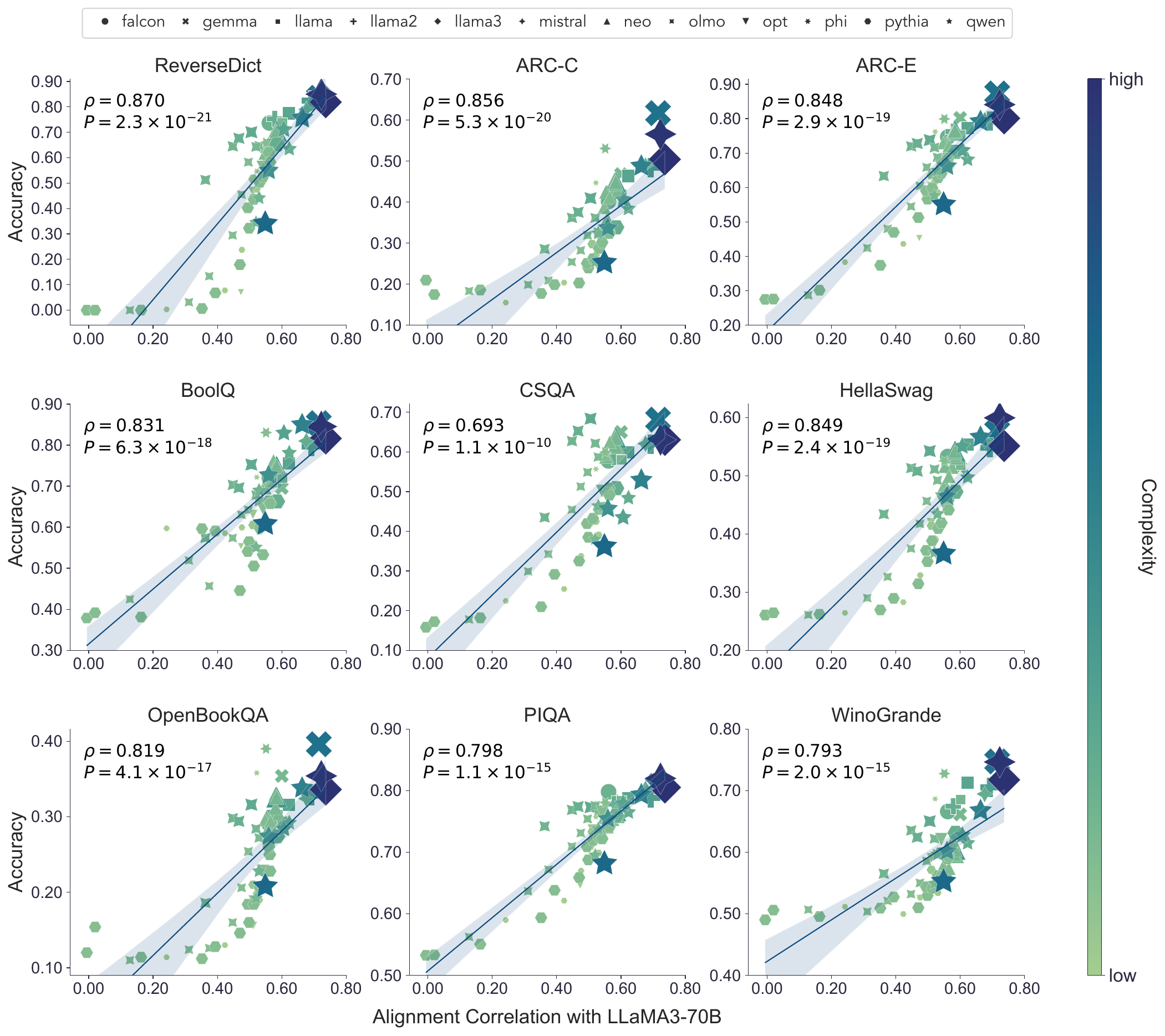}
\caption{Alignment with LLaMA3-70B conceptual representations predicts model performance across downstream tasks. In each panel, the x-axis denotes the alignment correlation (RSA) between each model and LLaMA3-70B, and the y-axis shows model accuracy on the corresponding task. For the reverse dictionary task, models were presented with 24 contextual demonstrations and evaluated using exact match accuracy, averaged over five independent runs; all other tasks were evaluated in a zero-shot manner. Each point represents a distinct model and is color-coded by model complexity. Linear fits are shown as straight lines, with shaded areas denoting $95\%$ confidence intervals estimated via 10,000 bootstrap resamples.}\label{fig:converge-tasks-si}
\end{figure}

\clearpage

\clearpage

\clearpage

\clearpage

\clearpage

\begin{figure}[htp]
\centering
\includegraphics[width=\textwidth]{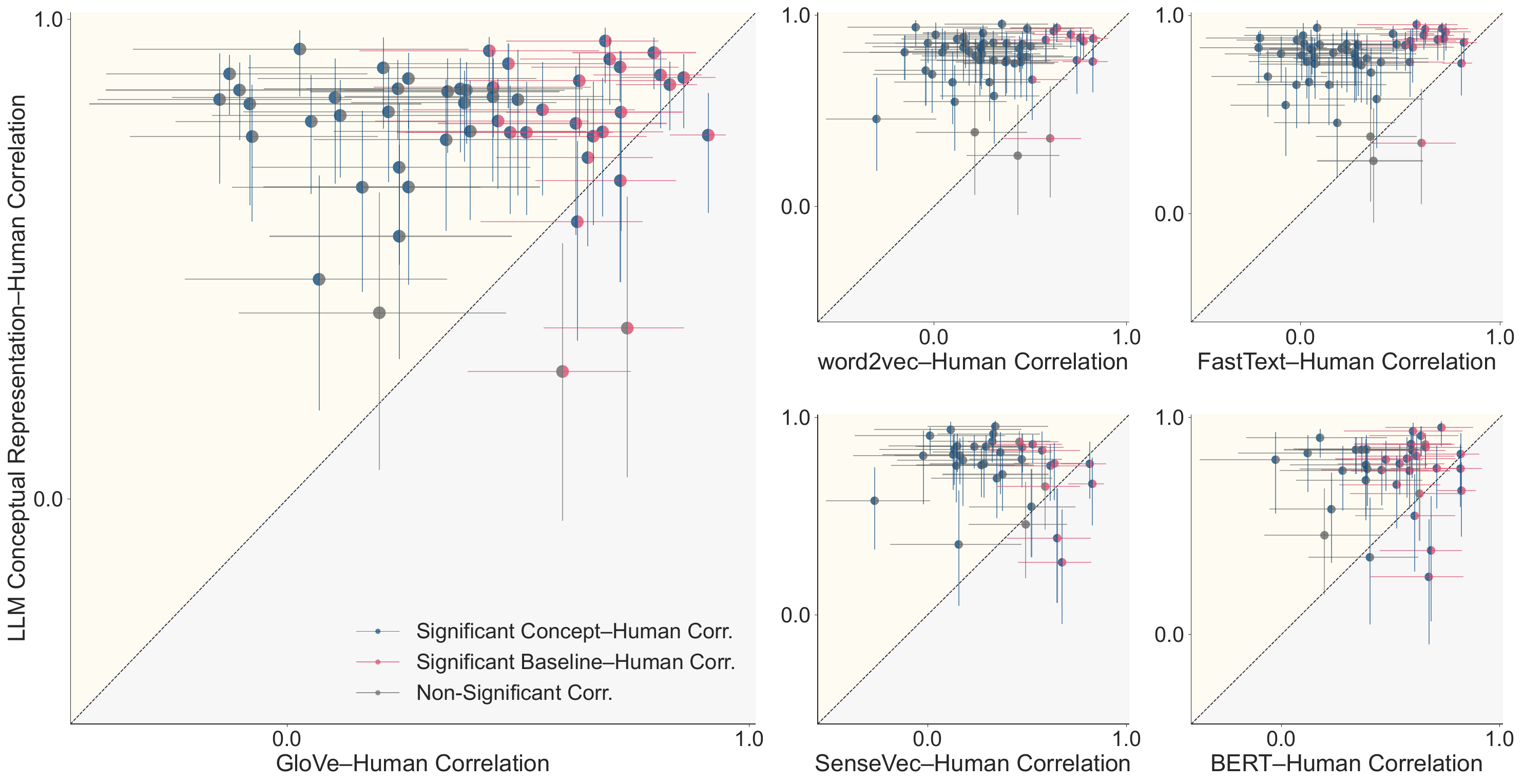}
\caption{Comparison of LLM-derived conceptual representations and baseline embeddings in predicting context-dependent human ratings along multiple feature dimensions. The x-axis shows correlations with human ratings based on baseline embeddings, and the y-axis shows correlations based on conceptual representations derived from LLaMA3-70B. Colored points indicate significant correlations (Spearman's rank correlation, FDR $P < 0.01$). Error bars represent $95\%$ confidence intervals estimated from 10,000 bootstrap resamples. GloVe embeddings are those used in previous work~\citep{grand_semantic_2022}.}\label{fig:sem-proj-concept-vs-word-si}
\end{figure}

\clearpage

\clearpage

\begin{figure}[htp]
\centering
\includegraphics[width=0.8\textwidth]{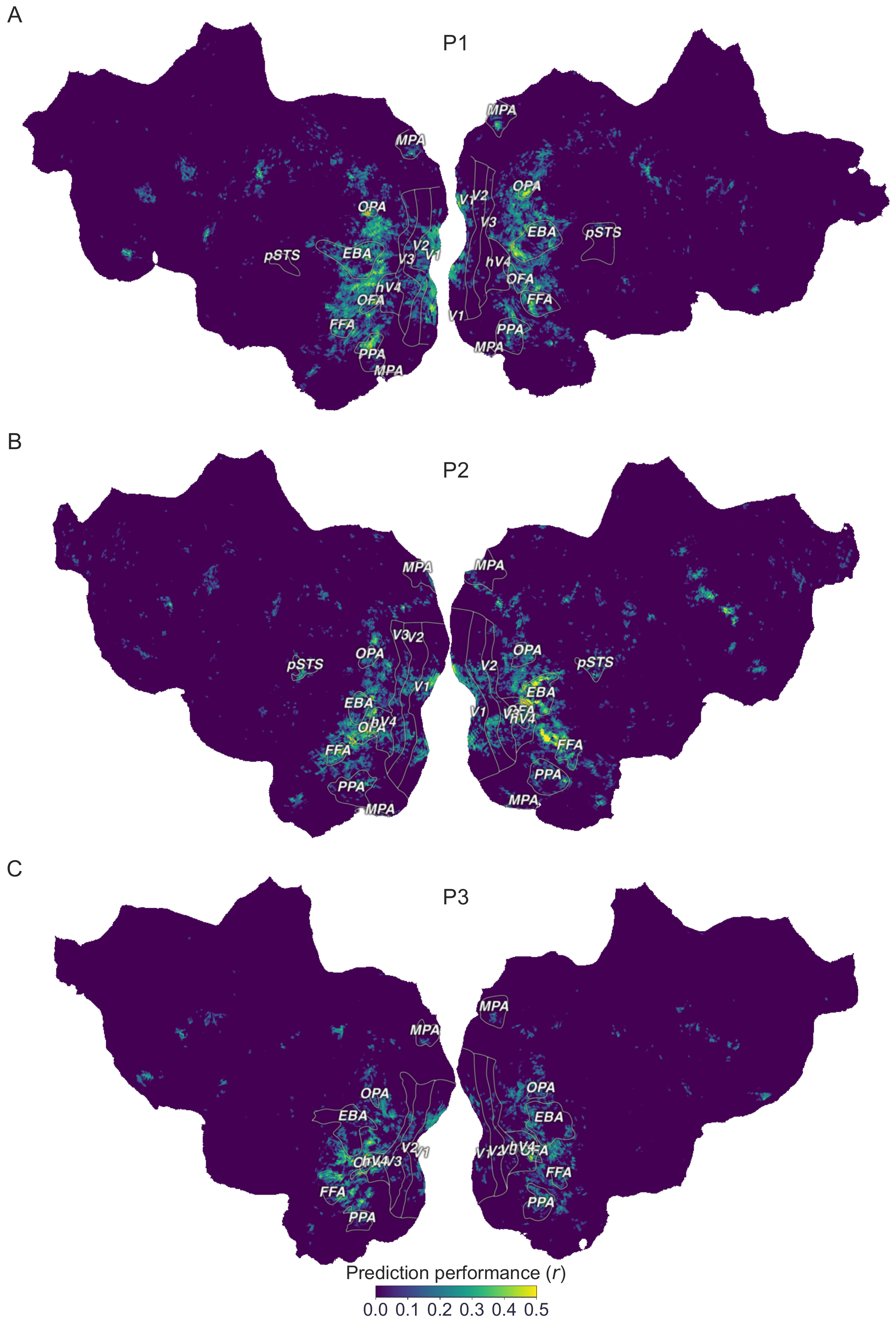}
\caption{fMRI prediction performance of LLM-derived conceptual representations (LLaMA3-70B) assessed via Pearson's correlation ($r$). \textit{A}--\textit{C} show the cortical maps of prediction performance for three individual participants. Only voxels with statistically significant correlations between predicted and observed brain activations are shown (FDR $P < 0.01$).}\label{fig:mri-encoding-corr}
\end{figure}

\clearpage

\clearpage

\clearpage

\clearpage

\clearpage

\clearpage
 
\clearpage

\clearpage

\clearpage

\clearpage

\clearpage

\begin{figure}[htp]
\centering
\includegraphics[width=.8\textwidth]{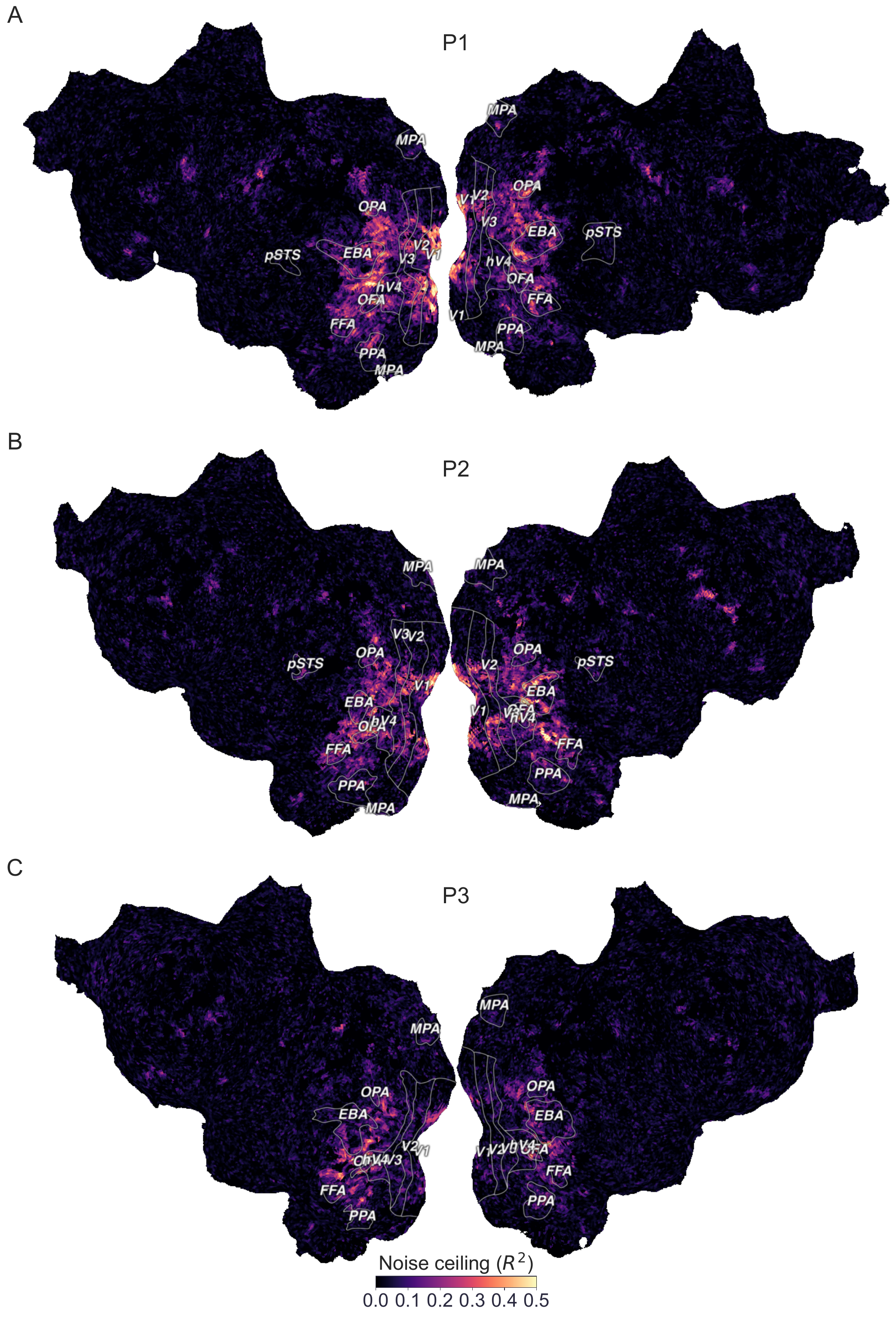}
\caption{Noise ceilings estimated for brain responses of three participants. Colors represent the explainable variance ($R^{2}$) in neural responses for different concepts. These noise ceilings were used to normalize the prediction performance across concepts.}\label{fig:mri-nc}
\end{figure}

\clearpage

\begin{table}[htp]
\centering
\caption{Major LLMs used in our experiments. The columns ``\#Parameters'' and ``\#Tokens'' represent each model's scale (number of parameters) and training data volume (number of tokens), respectively. For Mistral models, as the training data volume is not publicly available, we estimated it based on the training data of LLaMA 2 models, which were used for comparison in the Mistral paper~\citep{jiang_mistral7b_2023, jiang_mixtral_2024}. All LLMs are available at \url{https://huggingface.co}.} \label{tab:llms-details}
\begin{tabular*}{\textwidth}{@{\extracolsep\fill}llrr}
\toprule
\textbf{Series} & \textbf{Model} & \textbf{\#Parameters} & \textbf{\#Tokens} \\
\midrule
Falcon~\citep{almazrouei_falcon_2023} & \makecell[l]{\href{https://huggingface.co/tiiuae/falcon-7b}{\texttt{tiiuae/falcon-7b}}} & \makecell[r]{7.2B} & \makecell[r]{1.5T}  \\
\midrule
Gemma~\citep{gemmateam_gemma2_2024} & \makecell[l]{\href{https://huggingface.co/google/gemma-2-2b}{\texttt{google/gemma-2-2b}} \\ \href{https://huggingface.co/google/gemma-2-9b}{\texttt{google/gemma-2-9b}}} & \makecell[r]{ 3.2B \\ 10.2B} & \makecell[r]{ 2T \\ 8T} \\
\midrule
LLaMA 1~\citep{touvron_llama_2023} & \makecell[l]{\href{https://huggingface.co/huggyllama/llama-7b}{\texttt{huggyllama/llama-7b}} \\ \href{https://huggingface.co/huggyllama/llama-13b}{\texttt{huggyllama/llama-13b}}} & \makecell[r]{ 6.7B \\ 13.0B} & 1T \\
\midrule
LLaMA 2~\citep{touvron_llama2_2023} & \makecell[l]{\href{https://huggingface.co/meta-llama/Llama-2-7b}{\texttt{meta-llama/Llama-2-7b}} \\ \href{https://huggingface.co/meta-llama/Llama-2-13b}{\texttt{meta-llama/Llama-2-13b}}} & \makecell[r]{ 6.7B \\ 13.0B} & 2T \\
\midrule
LLaMA 3~\citep{dubey_llama_2024} & \makecell[l]{\href{https://huggingface.co/meta-llama/Meta-Llama-3-8B}{\texttt{meta-llama/Meta-Llama-3-8B}} \\ \href{https://huggingface.co/meta-llama/Meta-Llama-3-70B}{\texttt{meta-llama/Meta-Llama-3-70B}}} & \makecell[r]{ 8.0B \\ 70.6B} & 15T+ \\
\midrule
Mistral~\citep{jiang_mistral7b_2023, jiang_mixtral_2024} & \makecell[l]{\href{https://huggingface.co/mistralai/Mistral-7B-v0.3}{\texttt{mistralai/Mistral-7B-v0.3}} \\ \href{https://huggingface.co/mistralai/Mixtral-8x7B-v0.1}{\texttt{mistralai/Mixtral-8x7B-v0.1}}} & \makecell[r]{ 7.2B \\ 46.7B} & $\textrm{2T}^{\ast}$ \\
\midrule
MAP-Neo~\citep{zhang_mapneo_2024} & \makecell[l]{\href{https://huggingface.co/m-a-p/neo_7b}{\texttt{m-a-p/neo\_7b}}} & 7.8B & 4.5T \\
\midrule
OLMo~\citep{groeneveld-etal-2024-olmo} & \makecell[l]{\href{https://huggingface.co/allenai/OLMo-1B-hf}{\texttt{allenai/OLMo-1B-hf}} \\ \href{https://huggingface.co/allenai/OLMo-1.7-7B-hf}{\texttt{allenai/OLMo-1.7-7B-hf}}} & \makecell[r]{1.3B \\ 6.9B} & \makecell[r]{3T \\ 2.5T} \\
\midrule
OPT~\citep{zhang_opt_2022} & \makecell[l]{\href{https://huggingface.co/facebook/opt-350m}{\texttt{facebook/opt-350m}} \\ \href{https://huggingface.co/facebook/opt-1.3b}{\texttt{facebook/opt-1.3b}} \\ \href{https://huggingface.co/facebook/opt-2.7b}{\texttt{facebook/opt-2.7b}} \\ \href{https://huggingface.co/facebook/opt-6.7b}{\texttt{facebook/opt-6.7b}}} & \makecell[r]{356.9M \\ 1.4B \\ 2.8B \\ 6.9B} & 180B \\
\midrule
Phi~\citep{li_textbooks_2023} & \makecell[l]{\href{https://huggingface.co/microsoft/phi-1_5}{\texttt{microsoft/phi-1\_5}} \\ \href{https://huggingface.co/microsoft/phi-2}{\texttt{microsoft/phi-2}}} & \makecell[r]{1.4B \\ 2.8B} & \makecell[r]{30B \\ 1.4T} \\
\midrule
Pythia~\citep{biderman_pythia_2023} & \makecell[l]{\href{https://huggingface.co/EleutherAI/pythia-70m-deduped}{\texttt{EleutherAI/pythia-70m-deduped}} \\ \href{https://huggingface.co/EleutherAI/pythia-160m-deduped}{\texttt{EleutherAI/pythia-160m-deduped}} \\ \href{https://huggingface.co/EleutherAI/pythia-410m-deduped}{\texttt{EleutherAI/pythia-410m-deduped}} \\ \href{https://huggingface.co/EleutherAI/pythia-1b-deduped}{\texttt{EleutherAI/pythia-1b-deduped}} \\ \href{https://huggingface.co/EleutherAI/pythia-1.4b-deduped}{\texttt{EleutherAI/pythia-1.4b-deduped}} \\ \href{https://huggingface.co/EleutherAI/pythia-2.8b-deduped}{\texttt{EleutherAI/pythia-2.8b-deduped}} \\ \href{https://huggingface.co/EleutherAI/pythia-6.9b-deduped}{\texttt{EleutherAI/pythia-6.9b-deduped}} \\ \href{https://huggingface.co/EleutherAI/pythia-12b-deduped}{\texttt{EleutherAI/pythia-12b-deduped}}} & \makecell[r]{70.4M \\ 162.3M \\ 405.3M \\ 1.0B \\ 1.4B \\ 2.8B \\ 6.9B \\ 11.8B} & 300B \\
\midrule
Qwen~\citep{bai_qwen_2023, yang_qwen2_2024} & \makecell[l]{\href{https://huggingface.co/Qwen/Qwen1.5-0.5B}{\texttt{Qwen/Qwen1.5-0.5B}} \\ \href{https://huggingface.co/Qwen/Qwen1.5-1.8B}{\texttt{Qwen/Qwen1.5-1.8B}} \\ \href{https://huggingface.co/Qwen/Qwen1.5-4B}{\texttt{Qwen/Qwen1.5-4B}} \\ \href{https://huggingface.co/Qwen/Qwen1.5-7B}{\texttt{Qwen/Qwen1.5-7B}} \\ \href{https://huggingface.co/Qwen/Qwen2-0.5B}{\texttt{Qwen/Qwen2-0.5B}} \\ \href{https://huggingface.co/Qwen/Qwen2-1.5B}{\texttt{Qwen/Qwen2-1.5B}} \\ \href{https://huggingface.co/Qwen/Qwen2-7B}{\texttt{Qwen/Qwen2-7B}}} & \makecell[r]{619.6M \\ 1.8B \\ 4.0B \\ 7.7B \\ 630.2M \\ 1.8B \\ 7.6B} & \makecell[r]{3T \\ 3T \\ 3T \\ 3T \\ 12T \\ 7T \\ 7T} \\
\noalign{\hrule height 1.2pt}
\end{tabular*}
\end{table}

\clearpage

\begin{table}[htp]
\centering
\caption{Additional checkpoints of LLMs utilized in our experiments.} \label{tab:llms-details-checkpoints}
\begin{tabular*}{\textwidth}{@{\extracolsep\fill}lcrr}
\toprule
\textbf{Series} & \textbf{Model} & \textbf{\#Parameters} & \textbf{\#Tokens} \\
\midrule
MAP-Neo & \makecell[l]{\texttt{m-a-p/neo\_7b/20.97B} \\ \texttt{m-a-p/neo\_7b/41.94B} \\ \texttt{m-a-p/neo\_7b/83.89B} \\ \texttt{m-a-p/neo\_7b/157.29B} \\ \texttt{m-a-p/neo\_7b/251.66B} \\ \texttt{m-a-p/neo\_7b/387.97B} \\ \texttt{m-a-p/neo\_7b/524.29B} \\ \texttt{m-a-p/neo\_7b/681.57B} \\ \texttt{m-a-p/neo\_7b/723.52B}} & 7.8B & \makecell[r]{20.97B \\ 41.94B \\ 83.89B \\ 157.29B \\ 251.66B \\ 387.97B \\ 524.29B \\ 681.57B \\ 723.52B} \\
\midrule
OLMo & \makecell[l]{\texttt{allenai/OLMo-1.7-7B-hf-step1000} \\ \texttt{allenai/OLMo-1.7-7B-hf-step5000} \\ \texttt{allenai/OLMo-1.7-7B-hf-step10000} \\ \texttt{allenai/OLMo-1.7-7B-hf-step15000} \\ \texttt{allenai/OLMo-1.7-7B-hf-step20000} \\ \texttt{allenai/OLMo-1.7-7B-hf-step36000} \\ \texttt{allenai/OLMo-1.7-7B-hf-step72000} \\ \texttt{allenai/OLMo-1.7-7B-hf-step126000} \\ \texttt{allenai/OLMo-1.7-7B-hf-step197000} \\ \texttt{allenai/OLMo-1.7-7B-hf-step240000} \\ \texttt{allenai/OLMo-1.7-7B-hf-step287000} \\ \texttt{allenai/OLMo-1.7-7B-hf-step334000} \\ \texttt{allenai/OLMo-1.7-7B-hf-step382000} \\ \texttt{allenai/OLMo-1.7-7B-hf-step430000}} & 6.9B & \makecell[r]{4B \\ 20B \\ 41B \\ 62B \\ 83B \\ 150B \\ 301B \\ 528B \\ 825B \\ 1006B \\ 1203B \\ 1400B \\ 1601B \\ 1802B} \\
\midrule
Pythia & \makecell[l]{\texttt{EleutherAI/pythia-6.9b-deduped-step256} \\ \texttt{EleutherAI/pythia-6.9b-deduped-step512} \\ \texttt{EleutherAI/pythia-6.9b-deduped-step1000} \\ \texttt{EleutherAI/pythia-6.9b-deduped-step2000} \\ \texttt{EleutherAI/pythia-6.9b-deduped-step4000} \\ \texttt{EleutherAI/pythia-6.9b-deduped-step8000} \\ \texttt{EleutherAI/pythia-6.9b-deduped-step16000} \\ \texttt{EleutherAI/pythia-6.9b-deduped-step24000} \\ \texttt{EleutherAI/pythia-6.9b-deduped-step32000} \\ \texttt{EleutherAI/pythia-6.9b-deduped-step48000} \\ \texttt{EleutherAI/pythia-6.9b-deduped-step64000} \\ \texttt{EleutherAI/pythia-6.9b-deduped-step96000} \\ \texttt{EleutherAI/pythia-6.9b-deduped-step128000}} & 6.9B & \makecell[r]{0.5B \\ 1.1B \\ 2.1B \\ 4.2B \\ 8.4B \\ 16.8B \\ 33.6B \\ 50.3B \\ 67.1B \\ 100.7B \\ 134.2B \\ 201.3B \\ 268.4B}  \\
\noalign{\hrule height 1.2pt}
\end{tabular*}
\end{table}

\clearpage

\begin{table}[htp]
\centering
\caption{Overview of tasks and corresponding prompt templates used to evaluate LLM performance on language understanding and reasoning. Spearman's $\rho$ and $95\%$ confidence intervals indicate the correlation between models' alignment with LLaMA3-70B-derived conceptual representations and their performance on the respective task.} 
\begin{tabular*}{\textwidth}{@{\extracolsep\fill}llrr}
\toprule
\textbf{Task} & \textbf{Template} & \textbf{Spearman's $\mathbf{\rho}$} & \textbf{$\textbf{95\%}$ CI} \\
\midrule
ARC easy~\citep[ARC-E;][]{clark2018think} & \makecell[l]{Question: [Question] \\ Answer: [Answer]} & 0.848 & 0.727--0.923 \\
\hline
ARC challenge~\citep[ARC-C;][]{clark2018think} & \makecell[l]{Question: [Question] \\ Answer: [Answer]} & 0.856 & 0.754--0.919 \\
\hline
BoolQ~\citep{clark-etal-2019-boolq} & \makecell[l]{[Context] \\ Question: [Question] \\ Answer: [Answer]} & 0.831 & 0.715--0.903 \\
\hline
CommonsenseQA~\citep[CSQA;][]{talmor-etal-2019-commonsenseqa} & \makecell[l]{Question: [Question] \\ Answer: [Answer]} & 0.693 & 0.493--0.839 \\
\hline
HellaSwag~\citep{zellers-etal-2019-hellaswag} & \makecell[l]{Question: [Question] \\ Answer: [Answer]} & 0.849 & 0.725--0.924 \\
\hline
OpenbookQA~\citep{mihaylov-etal-2018-suit} & \makecell[l]{Question: [Question] \\ Answer: [Answer]} & 0.819 & 0.690--0.907 \\
\hline
PIQA~\citep{bisk_piqa_2020} & \makecell[l]{Goal: [Question] \\ Answer: [Answer]} & 0.798 & 0.640--0.899 \\
\hline
WinoGrande~\citep{sakaguchi2021winogrande} & \makecell[l]{[Context] [Completion]} & 0.793 & 0.643--0.894 \\
\bottomrule
\end{tabular*}
\label{tab:general-templates}
\end{table}

\clearpage

\begin{table}[htp]
\centering
\caption{Retained categories after filtering for categorization. The column ``\#Categories'' shows the number of categories retained out of the original set, and ``\#Concepts'' gives the total number of concepts within the retained categories. THINGSplus~\citep{stoinski_thingsplus_2024} provides an expanded categorization of the same 1,854 concepts in the original THINGS database~\citep{hebart_things_2019}.}
\begin{tabular*}{\textwidth}{@{\extracolsep\fill}llrr}
\toprule
\textbf{Dataset} & \textbf{Categories} & \textbf{\#Categories} & \textbf{\#Concepts} \\
\midrule
THINGSplus & \makecell[l]{animal; arts and crafts supply; body part; \\clothing; container; drink; electronic device; \\food; furniture; game; hardware; home decor; \\medical equipment; musical instrument; \\part of car; personal hygiene item; plant; \\scientific equipment; sports equipment; \\tool; toy; vehicle; weapon} & 23/53 & 962 \\
\midrule
THINGS & \makecell[l]{animal; body part; clothing; container; \\electronic device; food; furniture; home decor; \\medical equipment; musical instrument; \\office supply; part of car; plant; \\sports equipment; tool; toy; vehicle; weapon} & 18/27 & 1,112 \\
\bottomrule
\end{tabular*}
\label{tab:categories}
\end{table}

\end{appendices}

\end{document}